# The MIT Voice Name System

Main Authors: Brian Subirana and Harry Levinson
Collaborators: Ferran Hueto, Prithvi Rajasekaran, Alexander Gaidis, Esteve Tarragó, Peter Oliveira-Soens

Massachusetts Institute of Technology and Harvard University

**Abstract**
This RFC white Paper summarizes our progress on the MIT Voice Name System (VNS) and Huey. The VNS, similar in name and function to the DNS, is a system to reserve and use "wake words" to activate Artificial Intelligence (AI) devices. Just like you can say "Hey Siri" to activate Apple's personal assistant, we propose using the VNS in smart speakers and other devices to route wake requests based on commands such as "turn off", "open <*this brand's*> shopping list" or "271, start flash card review of my computer vision class". We have tested parts of a suggested VNS architecture to converse with a light switch, to create shopping lists for different grocers and to review flash cards. We also introduce Huey, an unambiguous Natural Language to interact with Artificial Intelligence devices. We aim to standardize voice interactions to a universal reach similar to that of other systems such as phone numbering, with an agreed world-wide approach to assign and use numbers, or the Internet's DNS, with a standard naming system, that has helped flourish popular services including the World-Wide-Web, FTP, and email. Just like these standards are "neutral", we also aim to endow the VNS with "wake neutrality" so that each participant can develop its own digital voice. This means that designers can securely customize the voice, personality and "wake word" of their digital voice experiences, and that no one unintentionally eavesdrops on the resulting communications. We focus on voice as a starting point given the recent surge of conversational commerce devices, and explain briefly how the VNS may be expanded to other AI technologies enabling person-to-machine conversations (really machine-to-machine), including computer vision or neural interfaces. We also describe briefly considerations for a broader set of standards, MIT Open AI (MOA), including a reference architecture to serve as a starting point for the development of a general conversational commerce infrastructure that has standard "Wake Words", NLP commands such as "Shopping Lists" or "Flash Card Reviews", and personalities such as Pi or 271. Privacy and security are key elements considered because of speech-to-text errors and the amount of personal information contained in a voice sample.

---

Harry and Brian are affiliated with Harvard University, where they met while Harry was a graduate student and Brian on the Faculty of Arts and Sciences. Brian is also affiliated with the Massachusetts Institute of Technology as Director of the MIT Auto-ID Laboratory, where most of the work was done over the last five years (2017-2022), and as Director of the MIT and Accenture Convergence Initiative for Industry and Technology. For any inquiries, please email Brian Subirana at subirana@mit.edu and Harry Levinson at harry@akubot.com.

## Contents

## 1. Introduction

### An example of how we are trying to enable your digital voice

Following the ideas and rationale developpped over the last five years in the MIT Auto-ID Laboratory, some of which are presented in [24, 25, 49, 50, 52–54], we propose and summarize here a shared infrastructure humans, animals and things can use to hold natural language conversations with each other using digital services. Our aims share many things in common with the Open Voice Network, which started from research in our lab sponsored by Target Corporation, Intel and Cap Gemini (see Figure 2).

The following natural language commands illustrate the infrastructure we aim to support:

```
Hi Sigma, please add to my Alpha Beta shopping list apples and
bananas. Merge my Alpha Beta list with the home shopping list.
Remove cumin and send the home list to our
home email. Please, switch to Sigmund.

Hi Sigmund, please create a new spreadsheet using the travel
expenses template: add an expense for lodging, set the description
to hotel and the date to June first; the amount for this expense
was two hundred dollars and fifty seven cents. Ask google for
my forecast. Switch to Alexander.

Hi Alexander, please tell Alexa tomorrow morning at 8am to play
Bruce Springsteen by Spotify and tell Google Home to turn on
the kitchen lights.
```

In this example, Sigma, Sigmund and Alexander are meant to be digital personalities operated by different organizations. The conversation "switches" from one to the other through a novel open transfer mechanism we call the MIT Voice Name System (VNS). Personalities may share devices, work on their own web pages or have their own. Sigma may respond from your browser, while Alexander and Sigmund may share a raspberry Pi. Our system is designed to be joined by existing platforms such as Apple's Siri, Amazon Echo and Google Home.

Different personalities may share the same grammar or define their own, using the conventions of Huey, a new language we have designed. To this date, Huey aims to be a super set of existing languages such as those introduced by Apple (Siri and Voice Control), Google and Amazon.

### Ownership of your Digital Voice

Infrastructures listening to audio conversations have the potential to train very powerful artificial intelligence models that may do things such as longitudinally predict the onset of diseases or assess your emotional state, political intentions and purchasing priorities. We aim to architect a system that enables multiple options on how the data ownership and rights may be handled, including the existence of third party organizations that manage such rights for users based on preferences. Our view is that such organizations would not only govern how model training may access your data but also which models you are exposed to. This may include basic preferences such as what voice you want to be addressed by, as well as more sophisticated ones that may not be yet possible such as selecting second language learning targets.

| BRAIN UNIT | LAYER NAME | MIT OPEN VOICE | APPLE | AMAZON | GOOGLE |
|---|---|---|---|---|---|
| Symbolic Comp. Mdls. | Conversation | ELIZA | - | - | - |
| | | MIT OV Regular Expressions | On Shot | One Shote | One Shot |
| | Virtual Selves | Sigma/Sigmoid/Sigmund/271/... | Siri | Alexa | OK Google |
| Cognitive Models | Cognitive Builder | Forgettable Ontologies | - | - | - |
| | | MIT OV Use Cases | Apps, VC | Skills | Smart Home Action |
| | Language | Huey, JAK | SiriKit | AWS Lambda | Assistant SDK |
| Brain OS | Sketch | Story Sketch Memory Architecture | | | |
| | Routine Instantiation | VNS: CWC, CPF, SVC | Hey Siri | Alexa | OK Google |
| Sensory Stream | Symbol | Sigma Open Source | OCR | OCR | OCR |
| | Sensor - Actuator | IoT Open Source Robotics | Smart Home | Smart Home | Smart Home |
| | Communication | Link with OSI Layers | | | |

**Figure 1.** MIT Open Voice Architecture Components

## Overview of the rest of the White Paper

We first lay the ground for the different “Wake Constructs” (WK) that may be used to wake a device under a possible standard, including voice, IoT activators, brain or gaze interfaces and computer vision. We move on to first present a simple use case: replacing the functionality of a basic light switch with an artificially intelligent system that allows robust user interaction and customization. This use case is a proxy for the process of waking any Artificial Intelligence device. We then expand to more complex basic use cases, including a coffee pot and a voice based smart assistant. We then evaluate specific use cases relating to the issues that arise from handling voice through Natural Language Processing. Finally, we expand and conclude this section by evaluating use cases in several industries.

One of the key aims of this white paper is to propose a routing standard for IoT and voice based requests from Wake Constructs: the VNS (Voice Name System). The VNS would be part of a larger architecture that would support the example use cases we’ll present throughout the paper. Figure 1 illustrates some of the possible architectural components that we envision such system would have.

We present progress towards a reference architecture where one or more organizations can have copies of the root server. In particular we present an example where GS1 would host the VNS Root Server. We created a prototype that can create shopping lists from an iPhone/Android App, which we downloaded for test purposes as MIT Voice, or through accessing a service on a website, which we tested as Sigma at https://opensigma.mit.edu/. The architecture we introduce is designed to also route smart speaker voice requests to 3rd party services in a “neutral” way. Our architecture behaves, as suggested above, similarly to the DNS on the web, the NANP on the phone, and to GS1’s barcode standard. Key issues our architecture addresses include: the collection of voice samples to design wake engines; the creation of an emotional firewall to prevent leaking PII; the extension to other interaction modes such as EEGs or Vision; the support of AI services based on shared health data; and provisions for the prevention of phishing attacks.

In the conclusion we present a few possible call-to-action options hoping this will inspire the leader to pursue additional ideas stemming from our proposal.

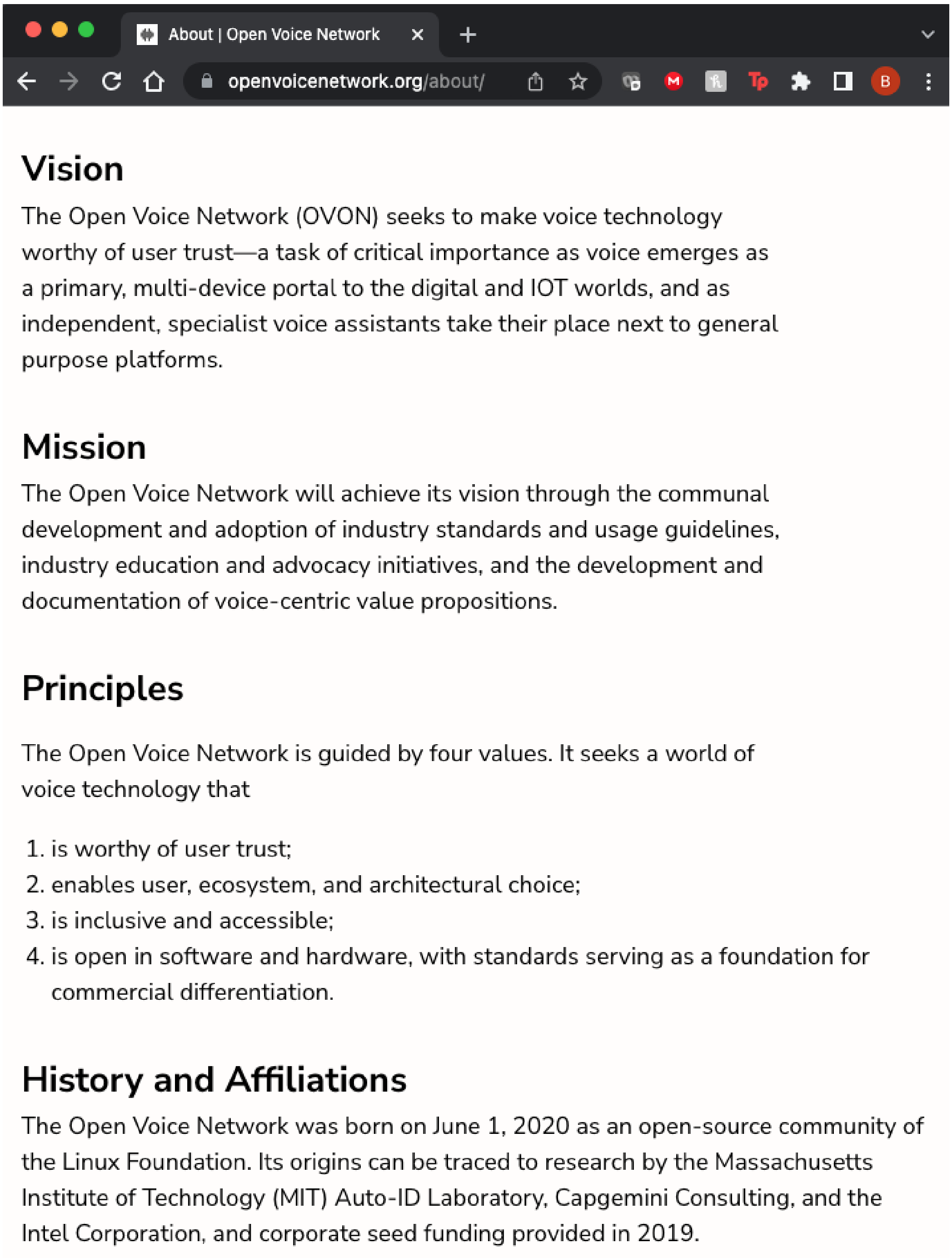

## Vision

The Open Voice Network (OVON) seeks to make voice technology worthy of user trust—a task of critical importance as voice emerges as a primary, multi-device portal to the digital and IOT worlds, and as independent, specialist voice assistants take their place next to general purpose platforms.

## Mission

The Open Voice Network will achieve its vision through the communal development and adoption of industry standards and usage guidelines, industry education and advocacy initiatives, and the development and documentation of voice-centric value propositions.

## Principles

The Open Voice Network is guided by four values. It seeks a world of voice technology that

1. is worthy of user trust;
2. enables user, ecosystem, and architectural choice;
3. is inclusive and accessible;
4. is open in software and hardware, with standards serving as a foundation for commercial differentiation.

## History and Affiliations

The Open Voice Network was born on June 1, 2020 as an open-source community of the Linux Foundation. Its origins can be traced to research by the Massachusetts Institute of Technology (MIT) Auto-ID Laboratory, Capgemini Consulting, and the Intel Corporation, and corporate seed funding provided in 2019.

**Figure 2.** About page of The Open Voice Network (OVON), a neutral, non-profit industry association operating as a directed fund of The Linux Foundation, whose origins can be traced to our research group in the MIT Auto-ID Laboratory. The aims of this white paper are very much in line with those of OVON as shown in this image of their official about page. The image was taken directly as a screenshot of the About page of http://www.openvoicenetwork.org (Accessed March 2022).

## 2. Huey Grammars and Conversational Wake Constructs (CWC)

One of the key ingredients of the VNS-Huey system is to separate the smart speaker from the handler of the commands. For us, smart speakers include Echo and OK Google but also any device that has a web browser with microphone and speaker capabilities. Today not all browser/device combinations support voice and sound but many do. The VNS-Huey system is defined in such a way that interactions may occur also via typed commands, sign language, IoT devices or even EEG input. In this context, the vision we are proposing is that a VNS speaker may route commands to a number of voice handlers, including Alexa, Google or any specific web page that has a VNS voice handler such as opensigma.mit.edu. One of the key architectural components of our proposed system is a mechanism to switch from one handler to another, very much like when you are browsing a page like www.amazon.com and decide to switch to <*Grocery-Store-Name*> and you simply type www.<*Grocery-Store-Name*>.com on the browser. From that point on, the browser experience is controlled by <*Grocery-Store-Name*> until you decide to type a different domain or you click on a place that routes you out of <*Grocery-Store-Name*> into another web server.

In Huey, switching from one device to another is done by using some reserved words and constructs. We call these constructs “Conversational Wake Constructs” (CWC) and we define them explicitly in Huey. The design and implementation of Huey is described more in detail in section 5. Here we present briefly the Switching Grammar, which determines how Conversational Wake Constructs Work, and then show a couple of examples of specific grammars to illustrate how the language can be used to standardize other conversational commerce interactions such as generating shopping lists or building a simple spreadsheet.

### 2.1. *VNS Switching Grammar in Huey*

We start by illustrating how users can switch between one voice handler and another one. The aim is to create a natural language interface that is simple for users to master and that it’s predictable. This means that VNS approved voice handlers will have to incorporate this grammar in order to follow the standard. A second thing that will be required is that the way commands are defined follows the Huey grammar style. As we will demonstrate in section so-and-so, Huey can define arbitrarily complex grammars and can be deterministic if used wisely. Let’s start with an example of how a user may switch between two handlers, say from Sigma to Alexander.

Via voice, or simply typing, to change from one assistant to another, the user can input a command such as:

```
Hi Huey, please switch to Alexander.
```

Switch is not the only word reserved to do so. Huey v1.0, as reflected in the switch grammar (see Appendix A), also supports four other “synonyms” of switch: connect, tell, ask, go to. This means that the following sentence would achieve the exact same transition:

```
Hi Huey, please connect to Alexander.
```

which would be parsed by Huey V1.0 and converted to an S-expression, which is a widely used data format in NLP systems.In Huey V1.0, implemented using ANTLR,

```
$ huey
Huey Shell
Version 1.0
Copyright 2020 MIT.
MIT License.
Type :h to get help, :q to quit
huey> login harry
Hello harry!
huey>
huey> :t
Verbosity set to:  TRACE
huey> Hi Huey, please connect to Alexander.
OK
(input (stmt (tell_assistant (meta (attn hi) (wake huey) (ignore ,)) (ignore please)
(connect connect (to to)) (assistant (wake alexander)))))
huey> :f
FRAME
Awake:  true
Interpreter:  VNSInterpreter
Last command:  Hi Huey, please switch to Alexander.
Last verb:  connect

huey> add bananas to shopping list
[OK]
(input (stmt (stmt_shop_top (stmt_shop (add_item (add_qty_item (add add) (item bananas))
 more_items (to_shop_list (to to) (shoppingList shopping list))))))
huey> :f
FRAME
Awake:  true
Interpreter:  ShoppingInterpreter
Last command:  add bananas to shopping list
Last verb:  add
Last object:  bananas
```

**Figure 3.** This figure illustrates the prototype Huey system taking as inputs the phrases "Hi Huey" and "add bananas"

the above user statement would be described by this S-expression:

```
(input (stmt
    (tell_assistant (meta (attn hi) (wake huey) (ignore ,)) (ignore please)
    (connect connect (to to)) (assistant (wake alexander)))))
```

The chain of grammar statements supporting this is

```
input : stmt ...
stmt : tell_assistant ...
tell_assistant : meta ignore* connect assistant
connect : 'ask' | 'tell'
assistant :  wake | item
wake : 'alexander'
```

where `item` is a phrase of one to six words.

Refer to Appendix A for detailed grammars.

We distinguish three goals in designing Huey. The first, and most important one, is that there is predictability in how smart speakers will respond to human produced voice commands. The second and third goals aim, respectively, are privacy and security. To achieve these two goals we introduce next a new intermediate language, Jak.

## 2.2. *The Need for Jak and Specialized Huey Grammars*

Huey is a language that facilitates humans speaking to digital assistants with natural language. As with existing human languages, Huey requires grammars to provide some structure to these conversations. Huey runs on collections of grammar files, which is why we use the plural form *grammars* here.

In the previous section we described how users may switch from one assistant or digital persona to another. Now we provide concrete examples of designs for several domains. We present selected grammar statements to illustrate how Huey can understand natural language, by providing templates to match phrases provided as input from users.

Appendices A, B, and C contain the grammars in the ANTLR format (for Java and Python developers), and Appendix D presents the same grammars in BNF format.

We also introduce the Jak language, which represents users' requests in computer code. Jak is an invented computer language which we propose to use as an intermediate layer of the MIT Open Voice Architecture.

Given the current state of sentiment analysis, unless we highly restrict human expressions, it will be very hard to avoid the recipient from learning about the emotions of the users. By converting user input into intermediate forms, for example using Jak, we have an opportunity to filter sentiment to some degree.

While not yet fully developed, Jak will provide the means to translate input text to actions for a digital assistant on the VNS to carry out.

### *2.2.1. Shopping List Grammar*

In the examples explained so far, we see that Huey can convert text (from speech or chat) into an s-expression (which we may also call an *s-expr*).

An advantage of the s-expr is that and can be parsed in almost computer language, and then the application perform a data operation (CRUD) or an action based on the information. In particular the s-expr format should be very familiar to Scheme or LISP programmers. In those languages, lists can represent code, data, or both.

Taking this concept a step further, we can use Jak to represent the information from an s-expr. Then the Jak source code can be executed by a Jak interpreter or compiler written in any common computer language. This frees us from the constraint of using Scheme or LISP, which for some time have been much less popular than Java or Python.

Here is a slightly simpler version of our shopping list example:

```
add bananas

(input (stmt (stmt_shop_top
  (stmt_shop (add_item
    (add_qty_item (add add) (item bananas))
        more_items to_shop_list)))))
```

Huey interpreted `add bananas` and output the s-expr for us.

An s-expr compiler could convert the s-expr to this short Jak program:

```
set(shoppingList, "shopping list")
set(add_item, "bananas")
call(shoppingHandler, add_item, shoppingList)
```

Now we have a Jak program. How do we execute it, and what is the result?
First the Jak interpreter sees

```
set(shoppingList, "shopping list")
```

which is an assignment statement. In Java, this would cause the following statement to be executed:

```
String shoppingList = "shopping list";
```

Likewise the next Jak statement would result in

```
String add_item = "bananas";
```

and finally

```
call(shoppingHandler, add_item, shoppingList)
```

would perform the action of updating the database.

In this case the user did not name a specific shopping list, so their general shopping list would be updated. Also, the quantity will default to `1` because it was not specified explicitly by the user.

### *2.2.2. Spreadsheet Grammar*

Earlier in this paper, our long example contained the following request:

```
Hi Sigmund, please create a new spreadsheet using the travel
expenses template: add an expense for lodging,
set the description to hotel and the date to June first;
the amount for this expense was two hundred dollars and
fifty seven cents.
```

Huey supports a request such as this via two grammar files, included in the appendix:

- Sheet – general spreadsheet statements
- ExpenseSheet – expense report statements

For example, the first part

```
Hi Sigmund, please create a new spreadsheet using
the travel expenses template
```

matches a clause in the `stmt_sheet_top` statement in the Sheet grammar:

```
<stmt_sheet_top> ::= <meta> <ignore>* <meta_sheet> <from>
  <det>? <topic_expense>? <topic_expense>? <template>
```

and in turn `topic_expense` matches this statement:

```
<topic_expense> ::= "travel" | ("travel")? "expense" | "expenses"
```

in the ExpenseSheet grammar file.

Likewise, the remainder of the expense sheet request uses other statements from Sheet in combination with the shared grammar files CommonParser and CommonLexer.

## 2.3. *Use a Secure Voice Channel, an IoT Plug or other service*

### *2.3.1. Designing a Secure Voice Channel*

Commodity smart speakers typically respond to spoken commands from anyone they can hear. In addition to not necessarily distinguishing between the device owner and other people nearby, there is not much security.

This design choice provides some ease of use for asking familiar questions about the weather, sports scores, encyclopedia topics, and stock ticker prices.

But what if the user wants to buy something? A smart speaker can be configured to allow ordering from particular approved vendors. In the case of Alexa, the device owner typically configures the device with their smartphone over a local network, so the device can use the owner's Amazon store account. Then anyone speaking to that Alexa device can order or re-order common items.

In late 2020, Alexa also has some awareness of device ownership. You can ask, "Who am I?" and Alexa will reply "I'm not sure, but this is Dave's Echo Dot".

If a user would like to make a request that involves an unusually large expense, or send a message to contact who has not been communicated with recently, how would a device know the current user is the device owner or has authorization to make such requests? Also, if the digital assistant does not have a screen or keyboard, how does the user provide credentials?

Web and mobile applications commonly use multi-factor authentication (MFA) to securely log into a user account. The initial login information is typically user name and password. The additional authentication step may be providing a short numeric code that is emailed or texted to a previously registered account or device, respectively. Prior to the massive proliferation of general purpose mobile devices, the use of dedicated security key fobs (simple radio receivers) was common for receiving such codes.

Once a user is authorized to enter a web site or application, that application will determine the users privileges to access features within the application. This step is known as *authorization.*

To prevent eavesdropping on communications between the user and applications on the internet, secure protocols such as TLS (formerly known as SSL) provide encrypted messaging using encryption keys that are exceedingly difficult to discover by brute-force computing power. Users can see in their web browsers that their communications are secured using the `https` protocol and will be warned if a web site's certificate is expired or invalid.

A new protocol, Secure Voice Channels (SVC), will provide similar functionality for users speaking to digital assistants. The user can log into a device by speaking a command. When SVC is enabled, the device will initiate the authentication sequence. All data communications to authenticate, that occur outside speech, will occur with encryption.

Here is a sample interaction demonstrating the normal case of authentication for a voice user:

```
Alice:  Hi Huey
Huey:   Hello Alice. Please provide the 6 digit code I
```

```
        just sent to your mobile phone.
Alice:  293749
Huey:   Thank you Alice, you are now logged in.
```

If Alice did not have her phone with her, the conversation might go like this instead:

```
Alice:  I don't have my phone with me right now
Huey:   OK, here is a challenge question.
        Who was the last person you sent a voice message to
        and what day was it?
Alice:  Bob on Tuesday
Huey:   Thank you Alice, you are now logged in.
```

Suppose the user is logged in, and requests a highly destructive operation by voice. Authorization for this would require additional proof of identity as well as other information that was provided when the user's account was configured. This is analogous to the Linux `sudo` command, which temporarily gives a non-root user more privileges to carry out an action from a pre-set list.

```
Alice:  Huey, please delete all my messages from 2019
Huey:   The requested action requires additional privileges.
        Would you like to elevate now?
Alice:  Yes
Huey:   Please follow the instructions I sent to your
        secure message account just now.
```

This would also guard against nefarious activity by black-hat third parties, who might be using radio signals, ultrasonic signals, or similar cracking techniques to make the digital assistant perform privileged or destructive operations.

As with the highly privileged Linux `root` account, SVC will not allow initial entry into the system (login) at the strongest access level. The user must log in as a regular user, then provide additional information to request elevation of privileges. The signaling between user and back-end systems must always be encrypted in transit, to lessen or eliminate the ability of third parties to intercept credentials and impersonate legitimate users.

Note that as with other computer systems, SVC as a protocol cannot prevent arbitrary attacks. Specifically, if a device has a back door that was inserted at the factory or while the device was unattended before purchase, an attacker may be able to enter through the back door. Likewise, if a home or office user has digital assistants that are not physically out of reach, they may be compromised. Physical security is always essential to guard against intrusion. This is why data centers, research labs, and government facilities have very high levels of security and monitoring.

#### *2.3.2. Conversational Privacy Firewalls*

Sometimes we would like to protect the user from inadvertently transmitting "too much information". We can accomplish this using a Conversational Privacy Firewall (CPF).

Imagine the user now would like to buy tickets for a sporting event, and tells Huey:

```
"I desperately need two tickets for Sunday's Red Sox
baseball game and would pay any amount"
```

In addition to hearing the words spoken, a digital assistant is capable of capturing (or at least guessing) the emotion or *sentiment* of the spoken sounds and words.

CPF *Speech Incognito Mode* is the architectural component that provides protection against the emotion represented in the audio signal being forwarded along with the words. This building block has analog audio waveforms as input, but outputs only text resulting from speech recognition.

Now a person *reading* the text above might understand the stated urgency, but they will not hear the sound of the person's voice, identity, mood, emphasis, accent, or other nuances that could deliver sensitive, confidential, or tactical information to a recipient. We have effectively filtered out information present in the original payload to protect the user.

A truly helpful digital assistant should be smart enough to go even further to help its user to avoid problems. In this case, we would like the assistant to intelligently filter out the emotional and unlimited financial parts of the request represented by the text.

We can write Huey grammars to produce the following, non-neutral s-expr based on the user's literal request:

```
(input (stmt (stmt_shop_top
  (stmt_shop_search
    (subject I)
    (search_item (search need) (qty 2)
      (bundled_item (unit tickets) (for_of for) (det the)
      (when Sunday) (item red sox game))
    (offer_price (amount any))
    (sentiment_literal desperately need)
    )
  )
)
```

Notice how some sentiment is still clearly preserved in the phrase "desperately need".

If we engage the CPF *Strong Incognito Mode* our application will ignore the grammatical term

```
sentiment_literal
```

This would filter out both "desperately" and "need" from the request. Likewise, heuristics would detect the edge case of

```
offer_price = any
```

and filter this out. (Sentiment filtering may be implemented with standard NLP libraries or via manually coded word filtering and matching.)

The transformed and filtered s-expr could then be stated as:

```
(action_shop
  (shop_search
    (search_item (qty 2) (unit tickets) (when Sunday) (item red sox game))
```

```
    )
  )
```

We can see that all emotion and urgency, as well as budget information, have been removed from this request.

A possible implementation in Jak would be:

```
set(search_item, qty=2, unit="tickets", when="Sunday",
    item="red sox game")
set(shop_search, search_item)
call(action_shop, shop_search)
```

(See the Appendix for a sample Python program that generates the first Jak `set` statement from the s-expr.)

Notice we omitted the subject “I” from the code, because we implicitly set the subject to the current user. In this scenario, we would expect the application to notify the user with search results as they are available. The user could then issue a “buy” request by referring to a search result identifier.

Here we have successfully protected the user from unnecessarily transmitting both

- desperation level
- price/budget criteria

This use case can be extended from this search/purchase example to more general auction scenarios. Also, it can be extended further to cover non-financial use cases, potentially those including personal relationships.

Conversely, an agent should be capable of knowing when personal safety is being mentioned. In these cases we might not want to filter out urgency and emotions, because we would like to include them when forwarding the information to the proper authorities.

### *2.3.3. Decompiling and Reverse Engineering*

A computer firewall by definition controls communication traffic so that we isolate protected systems from users, systems, and networks on the outside of the firewall.

Here we have described a combination of hardware (voice assistants with microphones or keyboards for chat) and software (architectural components such as SVC and CPF) to control interactions between users and the intelligent agents handling requests.

The software translates textual commands from one form to another, to allow further processing and optionally to screen out certain information to protect the user.

If someone sees an intermediate representation, such as an s-expr or a Jak code snippet, can they get back to the original spoken or typed natural language request? If they could reconstruct the request, an interloper could gain useful information for staging attacks.

The short answer is no. With each transformation, we are essentially erasing information to get to the next representation. Mathematically we could state this as:

```
p -> q -> r -> s  etc.
```

Generally we are filtering data such that `q` contains lower information content than p and so on.

The situation is analogous to how Java compilers treat some generic code, employing type erasure when translating source code. Static analysis of the compiled code will not reveal the exact type of objects that will be used by application at runtime, because it is dynamically determined.

So while a programmer can use a decompiler, they will not necessarily have a full representation of how the code will operate at runtime. Also, by design most compilers remove or ignore comments. If the programmer included some emotional warning about the source code, decompiling the object code will not restore this lost information.

Another analogy is baking or frying food. You begin with a set of ingredients, and then push them through a series of operations. The chemical and mechanical processes transform the original ingredients into a form that contains some aspects of the original, but you cannot unbake the cake or unfry the egg.

This system of transformations provides higher security for users, by limiting the ability of interlopers who would like to get back to the original request which might contain sensitive information the legitimate user would like to protect.

#### *2.3.4. Connecting to Another Service*

If a users want to switch to a different assistant, it will be necessary to reload the grammars and recompile the system. This is very much like when you change sites within a web browser. Etc etc

## 3. Basic VNS Use Cases

In this section we present several use cases illustrating the required VNS functionality. We begin with voice routing and a simple light switch use case, where a user wants to act on the switch using voice. This use case is a proxy of turning on any appliance and requires three issues to be addressed by a VNS implementation. The first one is ensuring the conversational commerce interface awakes in response to a users request, which is also necessary in voice routing. Second, is capturing the command a user would like to execute including on what light or switch it would do so. Third, is to complete the desired specified action and provide feedback if necessary. The more commands a conversational commerce device supports, the more complicated the situation may become, such as specifying how long the light must be turned on. A further complication arises because speech, the current dominant way to engage a conversational commerce device (though not the only way our standard supports), is unlike a traditional switch that is turned on or off by toggling an actuator and without any interpretation uncertainty. A person's voice, instead, may not be recognized 100% of the time because of many aspects that can lower the quality of the speech translation. These include the direction the voice originates from, various pitches and accents it might carry, and the language used. If there are multiple voice recognition devices in a room, the VNS may have to resolve which device the user would like to interact with it.

### 3.1. *A Continuous Wake Architecture: VNS Wake Word Detection*

Perhaps the most simple use case for VNS, yet the most powerful one, is to implement a simple interaction to turn *on* or *off* a light, or, more broadly, for waking a given behavior: The VNS device continuously listens for a specific set of "wake words" and wakes a given command corresponding to the wake word. For example, we implemented a switch in our laboratory that turned the lights on or off depending on whether it heard the words "on" or "off". We implemented this use case with a neural network that was constantly listening for one of the two "wake words" using a Raspberry Pi 3 (RPi3), *PyAudio*[1] and a websocket that connected to a server processing overlapping fragments of 5 seconds every 2.5 seconds using a variation of the *DeepSpeech*[2] network to conduct speech-to-text processing. We developed various improvements aiming at running the network directly on a low cost computer [56].

This simple architecture may be applicable to the whole umbrella of use-cases we delve into next, since they can all incorporate the *continuous wake architecture* as part of the wake mechanism for each use case. Note that following the same concepts outlined for light switches, the standard may be able to handle doors and other devices analogously. A door's basic functionality is similar to that of a light switch in that it can be closed (similar to a light's off setting), open (a light's on setting), or somewhere between the two states (varying light intensities). We find the light switch to be a fairly useful proxy for most use cases we have been able to imagine.

## 3.2. *VNS Voice Routing*

### *3.2.1. Use Case: Wake Word Voice Routing*

The VNS voice routing use case is one in which voice recorded by a device is routed to a given handler based on the detection of a specific "Wake Word" through speech recognition. It's a continuous wake architecture where the command is routing voice to the handler as specified by the VNS root server. Our current focus is on internet and phone voice routing but other options may be added in the future, most notably one way and two-way radio. The specific "Wake Word" and its associated handler preferences need to be specified in the VNS Root Server. The Root Server is currently being developed at MIT. We are exploring a few proposals to host it including that GS1 takes care of it as we discuss later in the paper. The VNS Root Server stores a descriptor specifying various ways and options to manage voice requests by the handler services. This includes VNS sound conversions, broadcast options, privacy settings, security provisions and file formats. The VNS will treat speech and text indistinctly, so that text interactions be supported too.

The current broadcast options being considered include multiple parties, sending speech to a third party speech recognition engine (for privacy protection) and just sending the audio file directly to the handler. The handler can then react to the speech, text file or alternative file format received in whatever form it wants. Voice routing may either be two-way (e.g a phone call), one-way (e.g. a basic music request) or sanitized (e.g. a speech to text, or speech to speech conversion done by a 3rd party as specified by the handler). A key question that needs to be addressed is the file format that will be used by a VNS system. We have implemented a first version of the VNS using WAV audio file format because we found it's the most supported. This file format allows storing audio information without compression, making it very suitable for deep learning, as well as very versatile for reformatting to other file formats with minimum information loss.

[1] https://people.csail.mit.edu/hubert/pyaudio/
[2] https://github.com/mozilla/DeepSpeech

To demonstrate the viability of our suggested VNS standard, we have successfully implemented Wake Word Voice Routing in several web browsers (Chrome, Safari, Firefox, Opera, all of these both for desktop and mobile), in an iPhone App (available as “MIT Voice”) and in an Android App of the same name. Our code demonstrates audio can be successfully sent from all these options illustrating the potential for a voice infrastructure. We used React for our website front end, leveraging the Media Recorder API with a Polyfill to support recording for all the browsers mentioned above. We used React Native for both iOS and Android Apps, with consistent naming conventions and overall design between all platforms. These were all connected to a Python Flask back end, which processes requests from each platform individually to account for different file recording formats and converts them to a single standard WAV format.

### *3.2.2. Use Case: IoT Wake Voice Routing*

Wake routing can be implemented as well from any kind of actuator. If one were to interact with a specific shelf in a store for example, wake routing through voice might be convoluted or inconvenient (“271 <*Grocery-Store-Name*> Massachusetts Ave Store Shelf 23”). Instead, that same wake routine could be presented as a macro in the shape of a button. Similarly, the shelf could start “listening” when the user is standing in front of it, detecting it either with a distance sensor or via computer vision. IoT wake routing could be a feasible solution to privacy issues rising from “ever-listening” devices. Furthermore, deep learning speech recognition models are prone to error, and because of a lack of explainability, these errors can be sometimes hard to predict. Use cases that require more control or where errors in routing can result in severe consequences (such as in factory environments, medical, automated driving and so on), would benefit from more consistent wake routines.

### *3.2.3. Use Case: EEG/Gaze Wake Voice Routing*

One use case we have been designing in the lab is based on the gaze of the user. Through computer vision or simply by connecting an EEG device, we are able to asses where a user is looking at and activate a wake routine assigned with that item. Specifically, different light frequencies from a light source can be captured through EEG in the human brain when the person’s gaze is centered around it.[23] From simple use cases such as looking at a light and turning it off, to more complex ones such as activating different processes in a factory environment, EEG/Gaze Wake Routing could provide advantages over the voice routing discussed above in both noisy and silent environments.

### *3.2.4. Use Case: Sign Language Wake Voice Routing*

This use case addresses sign language. Wake routing would behave very similarly to voice routing but elicited through gestures.

## 3.3. *Wake Word Basic Commands Use Cases*

### *3.3.1. Use Case: Light Switch*

In this section, we will consider a switch, like in the first case above, but elaborate on its transformation with artificial intelligence to create a robust standards-related sophisticated switch. Beyond "on" and "off" here are many potential commands we may relate to such an enhanced light switch such as: send usage statistics to a user, authorize an application to control the switch, update software, or pay usage using cryptocurrency.

The switch use case discussion will serve to establish some of the most basic functionality we feel is necessary for elaborated wake word neutrality. In fact, subsequent use cases will build on the switch use case in terms of device activation and basic settings. These settings may eventually come pre-programmed on every device with a basic vocabulary to trigger abilities.

We leave further explanation of a standard vocabulary for later, and focus on just turning on the device by presenting different interactions to turn on a light switch in the form of seven proposed use cases:

(1) **Switch Command Use Case** - This would be the same one as the first use case described above.
(2) **Switch Wake Word Use Case** - *A user calls a given light and asks it to change state.* In this use case, a light switch is given a name that is spoken to rouse it before issuing commands. For example, a user voices, "**office switch** turn on" where "switch" is the name given to wake the light switch, and anything said after that is assumed to be a command for the light switch. In this case, "turn on" is the command to illuminate lights. Alternatively, if there are multiple switches within a range to hear speech, the switch's name is configured. In this case, it may be "kitchen lights turn on" where "kitchen lights" is the name of the light switch. To achieve this, the switch (or more generally, the voice infrastructure) has to be constantly listening and be able to discriminate between different names which can be challenging when some names share words and sounds, such as "studio lights" and "garage lights". A user could ask the light switch, "give me your information." To which, the switch might reply with its install date, manufacturer, energy consumption and more.
(3) **Switch Physical Activation Use Case** - *A user swipes an identification card, or presses a button to notify the light switch that they are about to issue a command.* This may be an essential use case when security is important. A television remove may be another typical example of this use case. The standard may support buttons or card readers in different locations than the actual device or even duo security activation with a remote, wireless, battery-operated key.
(4) **Switch Proximity Detection Use Case** - *A user walks into a room with their hands full of groceries and approaches a light switch, getting within two feet, before commanding it, "turn on". The light switch, sensing it is closest to the user compared to other light switches in the room, activates the lights that the user requests.* Similar to the other use cases, this differs from normal lights wired to a specific switch because the user could ask for any of the lights in the house to be turned on or off through any light switch. If the user walks away from the light switch, the light switch becomes no longer sensitive to speech and returns to a sleeping mode. Proximity sensors or even machine vision could be used for this use case. Proximity could present a crude use of computer vision, with gaze detection being a more sophisticated use as we explore next.

(5) **Gaze** - *A user stares at a light switch that uses computer vision and artificial intelligence to map gaze to infer where the user is looking. Upon determining the user is looking at the switch, it becomes sensitive to speech and the user utters a command to turn on the lights it is looking at.* The implement this use case, advanced computer vision algorithms need to be implemented.

(6) **Neural Activation** - *A user's gaze is determined via EEG.* A user's brain produces measurable frequencies that match the frequencies of a light-emitting source that a user focuses on. Thus, these frequencies in the brain can be mapped to specific functionality in a device such as a light switch. So called *neural activation* allows a user to select a single device out of many. This allows a user to choose between even having a light switch or just activating lights individually by looking at them. Using a light source above a doorway, a user could look at its specific "frequency" and have their neural response measured for the same frequency. Upon deducing a match between the brain and door's frequencies, the user could then command the door to close, open, or open part-way. Alternative to neural activation, this door commanding could be achieved through gaze-mapping software that would use cameras and computer vision to determine which door the user was looking at. Since gaze detection and neural activation aren't yet completely reliable, the other interaction methods mentioned above could be included in the standard among others as new technology develops. [23]

(7) **Electronic Communication** - *A user sends an electronic signal through the network to a light switch in their house with the contents, "turn off house lights". The light switch responds with a message confirming it has turned off all lights in the house.* Alternatively, the user could have the option to send a direct message, email, or voice note to the light switch to interact with it or alert it to accept a verbal command. Light switches may have email addresses, social media group accounts or even blog feeds.

(8) **Hub Wake Word Switch Use Case** - All of the use cases listed above can also be used as possibilities to awaken a "smart speaker" or hub which listens for commands and routes them to the device. This form of activation already exists in many devices on the market as of 2019. For example, the Amazon Echo begins listening to a user command after one of four keywords is uttered: Alexa, Amazon, Computer, or Echo. Once the Amazon Echo hears a command meant for a different device, it routes that command to the appropriate device. The hub could be dynamically paired to a given location to enable remote voice control.

After exploring the basic functionality of a simple, standards-related light switch, we move on to examine more elaborate systems. The next use case explores a coffee pot to expand the standard to a more complex environment with an appliance that has multiple settings and configurations.

### *3.3.2. Use Case: Coffee Pot*

This section considers the standard and how it could relate to an intelligent coffee pot. There are natural similarities between this use case and the prior light switch use case since the seven ways a user could activate or "wake" a light switch could carry over to appliances. We will assume that, unless stated otherwise, the topics discussed in one use case are carried-over to any of the subsequent use cases presented. In deed, a coffee pot may be activated via a wake word, physical activation, proximity, gaze, neural activation, electronic communication, or a hub. Additionally, the coffee pot could have basic on/off and information fetching commands similar to the light switch. In this section, we define additional actions and vocabulary specific to appliances. Before delving into this standard set of actions and vocabulary, we begin with an illustrative scenario of a VNS standards-related coffee pot:

*In a household of two, one person, user x, makes eye contact with their conversational commerce coffee pot and verbally commands it to brew a cup of coffee. The artificially intelligent coffee pot responds with a short, affirmative response and brews a cup of coffee to match user x's taste profile after using computer vision and voice recognition to determine who made the request. User y walks into the room moments later, holds a button down on the coffee pot, and asks it to brew a cup of coffee. Similar to user x's experience, user y gets a cup of coffee brewed to match their taste profile. After user y's cup of coffee, the appliance informs the two users that coffee grounds are running low and asks if they would like coffee grounds added to their grocery shopping list.*

The coffee pot use case scenario shows that a coffee pot that follows the standard could integrate into a user's life, offering them more customizability and personalization. Further, it is seen that an appliance such as the coffee pot could have the ability to communicate with another appliance or artificially intelligent agent. To demonstrate this with another example, consider an oven that could have the ability to be turned off from another appliance in a different room of a house, or even a different building altogether. If a user ever worries that they might have left their oven on, they could ask the nearest appliance or conversational commerce device to ping the oven and ask it to shut off. Each appliance could serve as an HTTPS endpoint that could receive and deliver commands and communications. To facilitate this level of connectivity and personalization, a standard set of actions could come be part of appliances supporting VNS:

- **Coffee Pot Expression Adjustment Use Case** - A user modifies how a coffee pot will communicate back to them. Deciding to switch from a response via a screen on the appliance, the user sets the coffee pot to respond verbally. Other possibilities include SMS, email, or a differing visual response based on lights. The user further customizes the appliance's response by adjusting response specific settings such as the appliance's tone of voice, accent, and more. Expression adjustment allows for personification of appliances, which can make users feel more invested and attached. It also governs communication from an appliance to a user, contrasting the next use case which reviews appliance-to-appliance or appliance-to-AI communication.
- **Coffee Pot Communication Use Case** - Communication between appliances could allow a robust user experience where a user could interact with their alarm clock in one room to start the coffee pot brewing in another room. In relation to the light switch use case, a user can interact with an appliance in room $x$ to turn off the lights in room $y$. Communication can be achieved through either a hub model where communications are packetized and routed through a central entity, or, in a distributed model similar to the Internet, where information is packetized and follows a semi-random path along a web of nodes to get to its destination. Appliances can communicate through different channels such as private local networks or public networks. Within an organization, for extra security communication between appliances could be kept behind a firewall and inside the confines of the company.
- **Coffee Pot Energy Saving Use Case** - A user asks an appliance to go into energy saving mode which limits the power consumption of the device. This works by shutting off non-critical features of the appliance, or by having the device go into a "hibernation" mode where physical interaction with the appliance is needed to wake it back up.

Once a device knows that a user would like to command it, it could begin sensing for further gestures, speech, or other forms of signals. To issue a command, a simple set of actions and vocabulary could come standard with all light switches and similar devices and fall in a few categories:

- **Basic Powering Commands** - Devices generally need commands for powering up and powering down. For a switch, this may mean toggling it from one setting to another. For a door, this may mean the door is open or closed. The phrase to achieve this could simply be "on" or "off". Establishing a basic common set of commands could greatly simplify wide adoption of the standard.
- **Information Fetch** - These may be a set of phrases that would retrieve details about a device including when it was installed, its energy consumption, WiFi settings, and more. This information could be delivered all at once or by a user specifically asking for one attribute such as "energy consumption". Part of these may include access to a verbal manual for the device.
- **Naming and Configuration** - These would be set up commands to establish basic naming and personalization configurations such as name, security codes and voice type.
- **Operational** - These may be a set of device-specific words that may be activated by a standard menu activation approach. This may include even a number pad, as is common in many help lines.

While the examples above are given via a voice interface, as we have mentioned before, the standard could be broadened to incorporate other forms of communication such as gestures or neural interfaces to command a device.

The coffee pot use case analyzes a domestic and consumer oriented scenario. The next use case considers expanding the standard in a manufacturing setting, moving away from more consumer-based use cases into commercial-based use cases.

#### *3.3.3. Use Case: Alexander*

The following use case scenario depicts a personal assistant device that is able to wake up on its own and perform specific tasks, including talking to people or other conversational commerce devices in and outside its network. For simplicity, we will build upon the use case of a home smart device that interacts with home appliances and the user. One of the key benefits of this use case is that the user's voice is not leaked to third party devices.

*User x owns a set of smart devices, which are all in his household. They are all independent and can be interacted with by using a wake word for each. One of these devices is a VNS smart speaker with the ability to wake up on its own and talk to other devices. When user x wakes up in the morning, the personal assistant detects it (using a combination of sensor options) and proceeds to prompt the smart coffee machine.*

*The personal assistant decides on a specific order matching user x's taste profile and proceeds to wake up the coffee machine and ask for the order. Later that day, user x texts the personal assistant asking it to turn off the heating. The personal assistant wakes up the thermostat using a wake word, and asks it to turn off the heating. Finally, when user x comes home, the personal assistant wakes up on its own and asks the user whether he wants to practice his knowledge on a language course he is enrolled in.*

This use case introduces a new idea that hadn't been brought forth previously. For now, most smart speakers such as Alexa or Siri are restricted to activating only under the user's command. Simple tasks such as appointment reminders are not contemplated in most devices. A standard set of use cases could be defined for devices to monitor activity in the household and autonomously respond when appropriate. Similarly, the device could relay information from the user to other smart devices in the household. This would allow for less "smart" appliances (which is, able to recognize voice commands but having limited scheduling/predictive capabilities and Internet access) to be part of a Smart Network. All these rules could be defined including which kind of monitoring (computer vision, movement sensors, temperature sensors, etc.) and what security standards (storing information locally, secure HTTPS requests to centralized or private server) should be in place for a device as such. The following use cases may be considered:

- **Alexander talks to Alexa** - A device is used as an add-on to currently available smart speakers. The device wakes up autonomously and uses a wake-word to engage the smart speaker as if the user was doing so. This device has different sensing capabilities and Artificial Intelligence routines that direct it as to when it should wake up. For example, when the user is in the room, it will prompt Alexa to test his knowledge on a course he's enrolled in.
- **Alexander talks to smart appliances** - A device is equipped with several sensing capabilities around the household that monitor temperature, movement, light intensity and so on. The device uses these combined with information on the user's routine and preferences to model an automation plan for the household. It then proceeds to use speech to wake up and activate the different appliances and smart products in the household to implement the model.
- **Alexander relays information** - A device can be interacted with speech, email, SMS and phone calls. The user texts the device and directs a set of instructions that should be carried out by the different devices in his household. The device proceeds to use speech to wake up and activate the different appliances and smart products in the household to relay this information. This same use case could be expanded on by a centralized computer system that can be accessed through a phone call, granting access to the voice cloud.

## 3.4. *Natural Language Processing Use Cases*

### *3.4.1. Use Case: Standard Lists*

This use case tackles the generalized problem of creating a list of items. It is easy to conceive many use cases in which this would be necessary. In retail, creating shopping lists is already an important feature of online shopping. Similarly, lists can be used to grant access to specific users to certain resources or store contacts in the form of phone numbers or emails. In this section we tackle the problems that arise in creating such list in the context of voice and conversational commerce, including list ownership and permissions, and how to possibly implement this in a VNS standard. We now show the versatility of lists in the following scenario:

*A user walks in his kitchen. He opens the fridge and realizes he's run out of milk. He wakes his personal assistant and creates a shopping list. He adds milk to the shopping list and makes sure to leave the list open so the rest of the family can modify and add items to the list. Indeed, later in the day, a family member decides to order batteries for the TV controller by adding the item to the list. At the end of the day, the list is delivered by the personal assistant to the retailer, which proceeds to complete the order request. As the order comes in, the warehouse manager is able to access it and add all the items to a new list for delivery the next morning. This list can only be modified by him and read by the delivery company, which uses this information to track the progress in the delivery. As the driver proceeds to deliver the purchased items, a new list is generated that can only be viewed by him and not modified, which directs him to specific addresses at specific delivery times.*

This use case scenario highlights the complexity introduced by creating lists through voice, where permissions and ownership have to be determined to avoid both privacy and security concerns. A shopping list for example could be modified by all users in a household, but should only be readable (and therefore not modifiable) by a retailer. The user should furthermore be able to choose whether this list can be referenced by machine learning algorithms to better suit his shopping experiences later on. This same scenario is completely different for a warehouse, where the manager might have more permissions to user created files than a delivery truck driver. The standard should encompass all use cases, by delimiting a unified grammar, but should not restrict the depth and extent of specificity for each product. Although the aim of this document is not to fully define this language, we now list several points of interest that should be taken into account:

- **Users and Log in** - The standard should allow for the creation of multiple users with diverging permissions and responsibilities. Furthermore, the application for each use case should determine which ways are suitable for log in. *Example: face recognition might be suitable for a shopping mall, while scanning a fingerprint might be more appropriate for an application in a hospital.*
- **Groups and Admins** - The standard should allow for the organization of users in different groups, which could grant special permissions, responsibilities and tools, or otherwise restrict them. Similarly, the standard should allow for certain users to be identified as *admins. Example: A shopping manager should be able to access all information regarding stock and purchasing, while a sales clerk might have restricted access to this kind of information.*
- **Permissions** - Different applications will have different needs when granting permissions to users and groups. The standard should allow for the definition of specialized permissions for each application. *Example: A simple to-do list might grant read and write permissions to different users, while a more complex purchasing application might restrict operations such as purchases above a threshold.*
- **States** - The standard should declare states for each application, based on variables defined by the application itself and stakeholders that influence the role and permissions of users and groups. *Example: When a user is logged in to a shopping list application, only he is able to interact with the device. When logged out, any new user can begin a purchasing session.*
- **Tools** - The standard should allow for each application to define different tools, which can be granted through permissions to different users. These could include the creation of a new user, the customization of a state and so on. *Example: A shopping list should allow for the creation of a new list, and to add or remove items; while a bookstore might allow for prompts to find out more about specific books.*

- **Communication** - The standard should allow for different applications to define different ways in which the device responds to prompts and more importantly communicates with the user. *Example: A smart speaker shows a red light when running out of battery. Instead, a smart sprinkler sends a phone notification for the same purpose.*

The list above is only but a set of building blocks for a standard language that should encompass all possible interactions between users and smart devices, through not only voice but any medium in conversational commerce.

Audio is sent to speech-to-text API, returning a transcription of the file. Several words, including "Sigma", "Sigmoid", "Sigmund", "Pi", "314", "271" can be used as a "wake word" for the VNS system. In other words wake word routing is only executed after that word is spoken. As soon as that word is identified through regular expressions, the whole text command is sent to the VNS routing system. This system goes on to identify the desired API location of the spoken command. For example, if the user were to speak "Sigmond <*Grocery-Store-Name*> add apples to my shopping cart", the VNS would identify that the command should be sent to <*Grocery-Store-Name*>'s API, which would then process the command "add apples to my shopping cart".

### *3.4.2. Use Case: One Shot Private Shopping Lists*

In the use case above, we discussed the multiple challenges brought up with creating a simple list through voice. In this use case we address some of these issues by proposing a shopping list application that can be shared while not capturing any user specific information. We now show the versatility of this shopping list in the following scenario:

*A user creates a shopping list with several items he wishes to purchase from multiple retailers. Before he shares it with them, he wishes to share it with his friends as they will add items to the shopping list as well. The user is concerned about his privacy as he doesn't want these new items to be used for targeted ad campaigning towards him or any of his friends. The shopping list is assigned several uniquely identifying IDs, each one serving a different purpose. He shares the "read and write" IDs with his friends, who use it to find the shopping list and add their desired items. The user then anonymously shares a "read" ID with the retailers, who proceed to complete the shopping list order. When the order is ready, the user can use his "admin" ID to set a delivery method and address.*

See Figure 4 for a depiction of the state machine model to support one shot requests.

This use case defines a space where all stakeholders interacting with a shopping list are anonymous to each other. The "Admin" user, or creator of the list, can give different access rights to other users or companies by sharing uniquely generated IDs. We propose below a set of possible generated IDs below:

- **Read** - The admin user grants reading rights to the shopping list by sharing this ID.
- **Read & Write** - The admin user grants reading and writing rights to the shopping list by sharing this ID.
- **Admin** - The admin user grants reading, writing and deleting rights to the

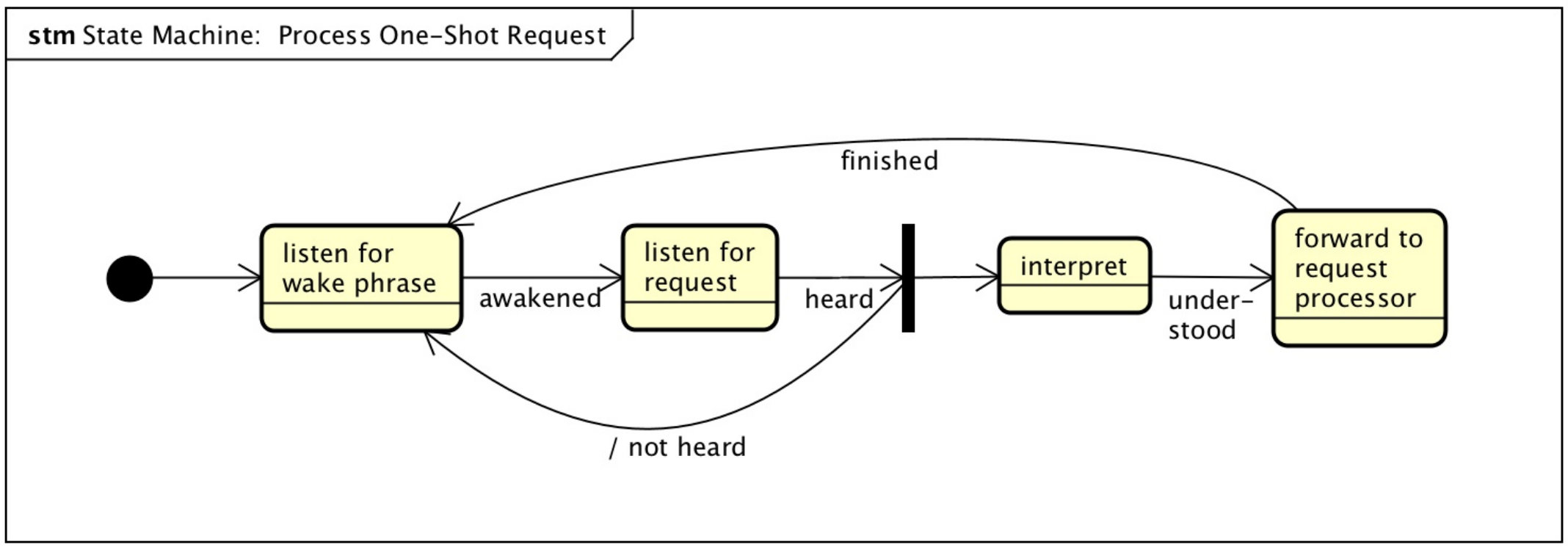

**Figure 4.** State machine for a one shot request

shopping list by sharing this ID, as well as the ability to share any of the other IDs.

- **Admin Transfer** - The admin user grants the same rights as the ones above, with the particularity that all IDs are regenerated and only shared with the newly appointed admin user. This ID can only be shared by the creator of the list or the user who was transferred to admin status with a similar code.

Sharing of these IDs could be done in any of the following formats:

(1) Direct ID number sharing.
(2) Barcode.
(3) QR Code.

### *3.4.3. Use Case: Multiple Choice Browsing*

While voice has many advantages in terms of user experience, it presents some limitations when browsing through a long list of items, as it makes it hard to properly visualize such lists in a way that is time efficient and practical for the user. For example, remembering items that were spoken at the beginning of the list might be challenging or downright impossible. In this use case, we present a possible solution to this conundrum with a multiple choice like interaction that allows the user to visualize short sets of items at once, interact with them and subsequently navigate through new sets of items until the whole list has been visualized. Here we describe a use case for this possible solution.

*A user walks into a large retail store. A smart assistant is hosted in a smart speaker near the entrance. The user queries the speaker, asking for the aisle where he/she can find soap. The smart assistant processes the query, and finds multiple matches. The assistant asks the user if any of the first four matches is what the user is looking for: "soap bars", "dish soap", "detergent", "body wash". The user is looking for "hand soap" specifically, and therefore asks the smart assistant to display other possible matches to his/her query. The assistant displays: "liquid soap", "hand soap", "novelty soap", "medicated soap". The user selects the "hand soap" option. The assistant proceeds to specify the aisle where this category is located, as well as depict on a screen four hand soap products available with price, reviews, ratings and so on. The user navigates through the list of products once again by switching between multiple four product displays. The user makes a decision on what hand soap to buy, walks to the*

*aisle specified, picks up the item and proceeds to checkout.*

This use case proves the power of list visualization through limited multiple choice options. Navigation in such environments could be standardized to provide a consistent experience throughout different environments and platforms. Below we describe several characteristics that are important when defining multiple choice browsing and should be taken into account depending on the application and platform.

- **Subset length -** In the use case described above, the costumer was presented with four options when making a query. This number might decrease or increase depending on the format of the voice assistant (an assistant with graphic interface might allow for longer lists) as well as the content of the lists (items that are longer in content might suggest browsing with shorter subsets of items).
- **Query type -** Queries could be not only voice but text, images or any other format. This would affect both the way the request is handled and how the response is displayed or conveyed.
- **Response display -** Responses could be restricted to a description from the assistant, but could also include audio recordings (i.e. song extracts), image or video (i.e. product pictures or videos) or any IoT actuators (i.e. different light intensity and hue combinations).

Up to this point, all use cases described assumed a mono-user world, where sessions didn't need to be accounted for and everyone who interacted with the device had access to the same skills and content. The poly-user world described here allows for more complex and sector specific use cases. We now move on to describe three of these in education, manufacturing and retail.

## 3.5. *Sector Specific Use Cases*

### 3.5.1. *Use Case: Education*

The following use case scenario depicts an application built by an education company to be engaged through a personal assistant. A personal assistant could take on many forms, but for this use case, we consider a personal assistant to be a smart speaker. Another variation of a personal assistant includes an artificially intelligent agent that is accessible through multiple devices and is constantly present. This personal assistant is considered in the 5th and final use case. For now, we turn our attention to a teacher teaching a virtual classroom of students through a single smart speaker:

*A teacher records a lesson via her personal assistant for students learning sign language around the world. Each student engages with their personal assistant which translates the lesson into the students' native languages. Following the lesson, the students participate in exercises to practice sign language where they sign to their personal assistant that uses computer vision to recognize and process different words and phrases. If a student has a question, they ask their personal assistant as if it was their teacher, waking the device by the teacher's name. Using a combination of artificial intelligence and web scraping, an answer is returned to the student. Both the question and the answer are recorded in a database for the teacher to review at a later time.*

This use case scenario highlights a few topics that could be covered by the standard that have not previously been mentioned. First of these topics is the potential benefit of a language agnostic system. Two users may want to communicate, but could be separated by a language barrier. Regardless of whether the language is spoken or signed, the conversational commerce device, in this case a smart speaker, could support seamless translation and communication between individuals. In order to achieve this and other functionality, a storage system is necessary.

The topic of storing information in a database was also introduced in this use case scenario. Having the capability to store relevant information about a user in a secure manner could lend itself to improved user interaction and the development of better applications across devices. The following types of interactions, among others, could be stored in an application specific database or in a public database (where a user's private information would be sufficiently masked):

- **General Expressions** - How a user acts in front of or with a device could be collected and stored to improve interaction models. For example, this could mean collecting speech to improve natural language processing models, or collecting gestures to improve computer vision models.
- **Application Specific Interactions** - A user may act a certain way with a company's application. Collecting this information and storing it could improve a user's experience with an application by offering more personalized recommendations and settings.
- **Personality Metrics** - A person's mood, disposition, and tone of voice are all personality metrics that could be collected and stored. This could lead to a personal assistant or service engaging with a user in a way the user finds more pleasing and interesting.

Storage of information and the subsequent changes to a device's interaction model is a passive process. A user of a device or application wouldn't necessarily know that the device is updating how it will interact with them in the background. The interactions the user has with the device just become more personalized over time. This passive process contrasts with the actions we have been defining in each of the use cases which are very much active and require the user to initiate them. Though, as with most topics covered by the standard, there could be a choice between making collection and interaction manipulation an action that a user would have the ability to turn on or off. Additional actions that a standard could support in pretension to this use case are as follows:

- **Education Change Speed Use Case** - A user commands the device to increase or decrease subsequent response times. For example, if a student is watching or listening to a lecture, they have the ability to manipulate the pace of media playback.
- **Education List Interactions Use Case** - A user asks a device to detail all of the prior answers and communications a user has had with that device. This also details where these phrases and words are stored, for example on a private or public server and the affiliation of said server.
- **Education Notifications Use Case** - A user may want or need to be prompted by a device. These interactions could either be set by a user or enabled in an application to remind or inform a user of something. For example, a user is notified when the time to solve a problem on a virtual quiz has expired.

In the education use case, we saw a specific model of the personal assistant, where a user's personal assistant was a smart speaker. The next use case again explores the possibilities of a standard in a commercial setting, but in this use case the personal assistant is more fluid and "exists" on multiple devices.

### *3.5.2. Use Case: Productivity*

The following use case explores the conversational commerce standard in a commercial setting – a production line in a factory. Whether the production line is fully automated or not, machines following the conversational commerce standard could bring down costs and work with more efficacy and velocity[51]. This could result from more streamlined interactions with machines where artificial intelligence chooses actions that have the highest probability of speeding up production, which could be confirmed by a human overseer.

A user could walk up to a machine, and due to their proximity to the machine, it would know that the human wants to engage it. Following that, the human could issue a command. The other options mentioned in the light switch use case are all viable options for activating or "waking" manufacturing machines. A worker could potentially be in a different country than the factory, realize something in the production-line needs to change, and send an SMS to a machine to tell it how to alter its behavior. However, feedback for machines does not need to just come from human users.

Additional devices could be used in conjunction with the machines to create a robust "cloud" of locally networked devices and machines to increase productivity. Below, we consider a non-finite list of four technologies that could be used to enhance a machine's decision making abilities. Note that using additional technology to enhance a device or machine's "senses" does not need to be limited to a commercial setting. In a home environment, a sensor network could greatly enhance the feedback a user receives from its appliances and devices. We now consider the four example technologies:

- **RFID Tags** - Using RFID, machines could identify a given item and track its movement on a product line. Specific wake words could be designed for engaging with RFID tags including specific commands such as, "send me your id" or "read me your description."
- **Moisture Sensors** - These sensors can be used to judge water content of the air to ensure a stable environment for sensitive products and may be voice activated too.
- **Temperature Sensors** - Knowing the temperature of a specific tool, or the environment in general, can verify that appropriate conditions for products, machines, and human workers are met.
- **Indoor Positioning Systems** - These systems could help detect and track humans and non-organic matter that might not be tagged via RFID. If a human gets too close to an active, dangerous machine, the machine could turn off and wait for the human to establish a safe distance.

The aforementioned technology could serve to enhance a standard set of actions that come with machines. For product-line machines, the previous standard actions mentioned in the light switch use case and coffee pot use case may be expanded for industry 4.0 settings. An energy saving mode may be warranted for buildings efficiency. Though basic functionality, such as switching a machine on and off should still be readily available, the standard actions for machines may include the following ones:

- **Productivity Track Unit Use Case** - This action determines where on the production line a certain unit is, where it has come from, and where it is going. For custom-made items, a consumer queries the product-line for details on the creation of their order.
- **Productivity Information Fetch Use Case** - Similar to the information fetch

presented in the light switch use case, a machine has an information fetch that delivers factory specific information. This includes details on the last time the machine was serviced and by whom.

- **Productivity Human Bypass Use Case** - A user intervenes to take over a machine's job for a period of time after a malfunction by uttering a bypass command. This setting is also security specific, only allowing certain individuals to engage with a machine if they have the correct user credentials. This prevents new workers from accidentally interacting with a machine crucial to a product-line and making it do something it shouldn't be.

These actions could work with an enhanced environment built from the ground-up for sensing and efficiency. If a machine's parts begins to show wear, it could be detected and the parts replaced at a convenient time. This could prevent other parts from wearing faster as well as an untimely shut-down of the machine. This is one example of how the conversational commerce standard could save both time and money in a manufacturing setting. Another could be by way of the aforementioned energy savings a connected factory could deliver through activation of "green modes" on non-critical systems.

The next use case will expand upon the commercial applicability of the standard through education. A different type of commercial setting than a factory, education will continue to reflect the monetary value that can be derived through the implementation of a standard.

### *3.5.3. Use Case: Retail*

Lists under the standard can be used for simple tasks as portrayed above. This final use case presents the conversational commerce standard for lists in a retail space. Elements of the previous six use cases, as well as a few new actions, are woven together to make a fluid, robust shopping experience for a user. This use case introduces a new potential architecture for personal assistants. In the education use case, we saw a personal assistant that could be tied to a hub such as a smart speaker and could only be accessed through that central point. In the grocery shopping use case scenario illustrated below, the personal assistant is an artificially intelligent agent that lives on a distributed network of conversational commerce devices. We now show this use case scenario:

*A user peruses the aisles at a local grocery store. Their personal assistant has made a list of items that their family consumes on a weekly basis. The grocery store's cloud of conversational commerce devices has access to this list and guides the user to specific items in the aisles using directed speakers to deliver audio uniquely to the user. Tracking their gaze, they are informed that their daughter's favorite brand of biscuit is found two shelves higher than the one they are looking at. Lights illuminate the grocery store walls in a pattern that helps them discover the package of biscuits. After grabbing the package, they engage in a conversation directed at the shelf that they pulled the biscuits from. "Give me the health information on these biscuits. Is there a healthier option?" Tactfully placed ambient microphones pick up the user's speech and the grocery store's directed speakers talk back to them, engaging them in a private discussion about the food they selected, despite other people being in the aisle with them. The user speaks again, this time asking to send a message to their daughter, asking if the recommended brand of biscuit is acceptable. In response, they hear the voice of their personal assistant confirming that the message has been sent. Eventually, the user checks out of*

*the grocery store simply by walking out. RFID tags on items are used to deduce what items the user bought, and their bill is subtracted from their checking account.*

In this grocery shopping scenario, we have included many standard actions that were presented in previous use cases to depict them in a different setting and show their versatility. The scenario also highlights additional actions could be included in a standard. We detail these actions next.

- **Grocery Shopping Information Fetch Use Case** - *A user walks up to an item in a grocery store and asks, "give me your price," to which the item verbalizes the amount it would cost to purchase itself. Other things a user can ask include the dietary information of an item or who manufactured or harvested the item.* This action is similar to the the information fetch presented in the light switch use case and productivity use case, but has more knowledge regarding retail items. Additionally, a user is not restricted to ask price and dietary information by walking from item to item and inquiring. Rather, they can engage with one item and access all other items in the store to make their shopping experience more efficient.
- **Grocery Shopping Cryptocurrency Use Case** - *A user is ready to leave the grocery store and walks out with their goods. RFID readers near the exit detect tagged items that the user is carrying and wishes to purchase. Having a balance of x Bitcoin in their cryptowallet, the grocery store deems the user has enough currency to buy the items, and the store subtracts their shopping balance from their wallet.* Cryptocurrency offers an opportunity to streamline transactions in a secure way.
- **Grocery Shopping Recommendations Use Case** - Given the history of our laboratory and our sponsors, this is a use case we have researched a lot. The following are a sample of the areas that may be covered by a standard:
  (1) A user engages with a product in a grocery store and asks it to offer recommendations for other items they might find interesting. The product responds with a list of recommended items generated using machine learning models that use historical information about the user.
  (2) A user does not find an item in the shelves and says "white beans missing".
  (3) The situation is the same as the previous one, but this time with a loyalty card enabling the digital infrastructure to alert store associates to bring the product to the user or to the check-out lines.
  (4) The user develops a shopping list throughout the week and converses with the shopping list when visiting the store.

These grocery use cases may require anonymous sharing of a user's PII with device back-ends. We feel they require broad industry discussions to insure all players are treated neutrally and that enough space for competition is given. In many ways the chosen use cases themselves may impact the shape of the competition.

### *3.5.4. Use Case: GS1 Resolver*

The following use case scenario depicts a device built to navigate through a worldwide VNS, which in turn gives access to a wide array of voice applications. These are produced, designed and maintained by individuals or companies. The VNS is used as a gateway, by storing the links corresponding to the different APIs that the user should interact with for each application. The different domain names for the VNS are stored in a centralized repository, which is in this use case maintained by GS1, the GS1 Resolver. Similarly to the way the DNS works, companies and individuals can purchase a domain name for a specific country through GS1. This domain should be tied to a Name Server, which is a URL of a server where a VNS database is stored. In those databases, information regarding specific application endpoints and other Name Servers is stored. In this use case we use the Wake Words *Sigma, Sigmond & Sigmoid* to grant a VNS access. The following three use cases showcase the wide variability of applications that could be sustained under such a system.

- **Use Case 1: VNS Certification -** *A user says "Sigma <Grocery-Store-Name> Add Shopping List Soap". The smart assistant proceeds to connect with the GS1 Resolver, which in turns redirects the assistant to the <Grocery-Store-Name> Name Server assigned for United States. The <Grocery-Store-Name> Name Server directs the whole command to its voice processing API, where the command "Add Shopping List Soap" is acted upon. The item "Soap" is added to the user's shopping list. The user now says "Sigma Carrots". The connection between <Grocery-Store-Name> Shopping List and the device is conserved: "Carrots" is added to the shopping list. Finally, the user says "Sigma Disconnect" to terminate the connection.*
- **Use Case 2: Standard Commands -** *A user interacts with a smart device with a video display. In this case, instead of navigating to a specific company, the user says "Sigmoid Information Rice Recipes". The command "Information" is defined as a standard command by the GS1 Resolver. The GS1 Resolver provides a URL to the smart device. The device displays the website from that URL, which corresponds to a list of recipes where Rice is one of the main ingredients, with reviews, comments and ingredients/directions.*
- **Use Case 3: Zone of Authority -** *A user interacts with a smart speaker by saying "Sigmund Heinz Ketchup Goodmorning Song". The GS1 Resolver connects the smart speaker with the "Heinz Ketchup" Name Server. The Name Server sends the command to the Heinz Ketchup API, which responds by delivering a song selected by Heinz Ketchup. The smart speaker plays the song. Similarly, the user then says "Sigmund <Grocery-Store-Name> Heinz Ketchup Goodmorning Song". In this case, the user now interfaces with "<Grocery-Store-Name>" (instead of "Heinz Ketchup"), which decides to deliver a song similar to the one provided by Heinz Ketchup, but combined with a <Grocery-Store-Name> theme song.*

These three use cases raise many questions in terms of how the VNS should be structured to grant secure access to all applications, appropriate zones of authority for companies or individuals who register a name domain, and standard commands designed to operate with this VNS. These concepts are explored in more detail in section 4.

The GS1 Resolver use case concludes our section on standards-related use cases. In each use-case we highlighted a potential set of actions that could come standard on every device, appliance, or machine. While these use cases may be distinct in nature, the standard actions of all five use cases defined within could be shared between the use cases presented here as well as in other scenarios.

## 4. Wake Word Standard Architecture

Continuing with the shopping use case scenario, when a user engages with an item in the grocery store, that item could have the ability to respond to brand specific commands that can wake it and query it. This could ultimately give the user more interesting and engaging interactions with the items in the store. Though, this does not need to be just retail specific. As described in the light switch use case, a light switch should be able to be turned on by saying “lights on” or “light switch, turn on.” There doesn’t always need to be a hub where users direct their commands that then get rerouted to the device the user was trying to activate. To achieve this, wake words must be neutral. Having wake word neutrality means specific key words could be reserved and recognized to attain a universal way of addressing devices. A user could be able to turn on a device regardless of manufacturer, and communicate with it.

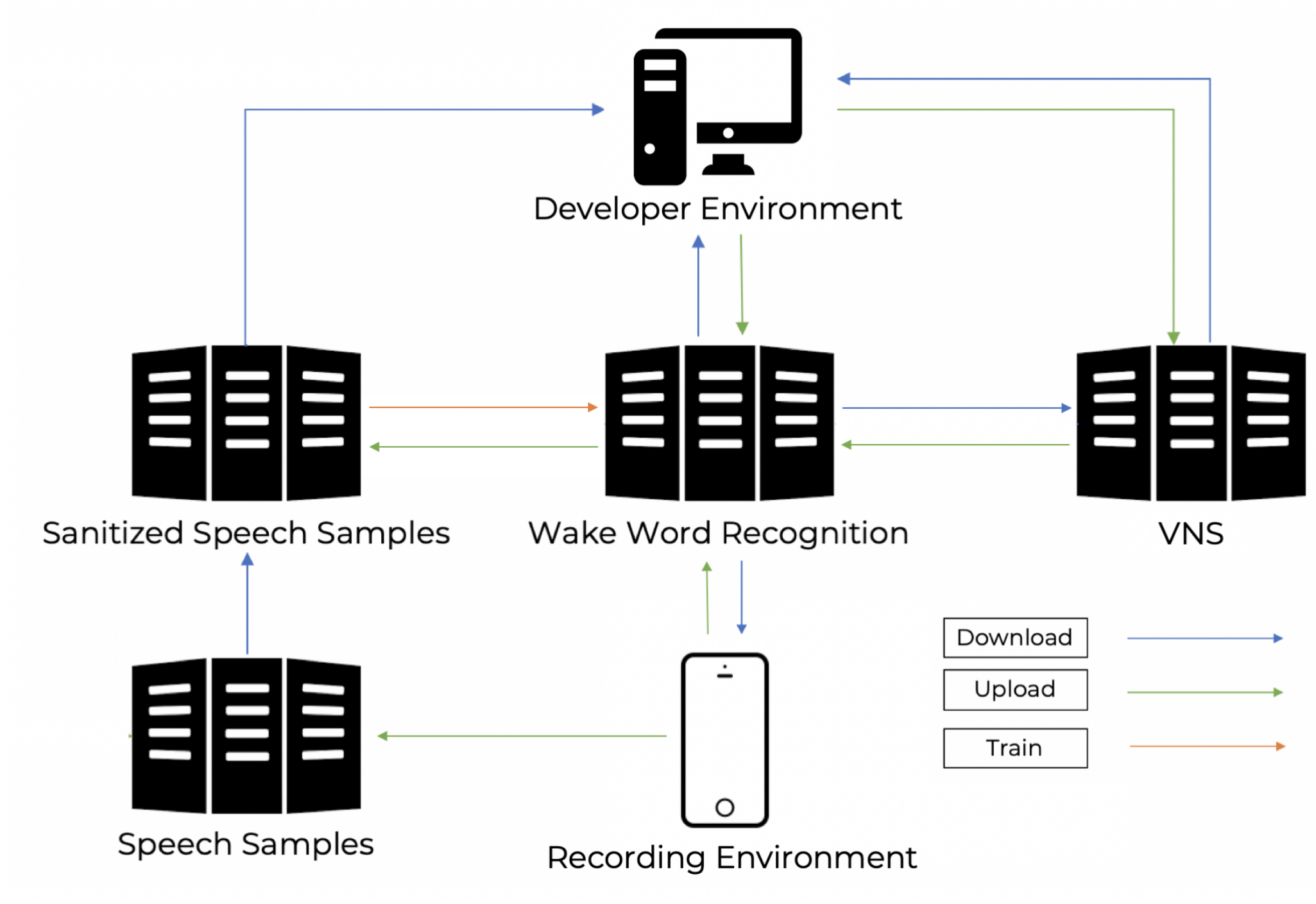

**Figure 5.** Open Voice Standard Architecture

A possible way to achieve this wake word neutrality is to develop a system similar to the decentralized DNS system of the Internet. A registry of words could be created so that a device could be activated with a given wake word that is specific to either a company or a device manufacturer. For example, a device could parse a phrase spoken to it, look up to see if any words in the phrase match one of the registered wake words, and if it does, it could pass the uttered phrase to a back-end owned by the company that registered that wake word. If the device doesn’t recognize any words in the phrase as registered wake words, it could continue listening. It should be noted that while this example is given with wake words, the standard could be built to incorporate different forms of standard communication and standard actions as noted above in our use cases.

A proposed architecture that implements the Open Voice standard could consist of 4 computing clusters combined with a front-end service. The work of aggregating and processing data from devices could be split between these servers. A Voice Name System takes the function of a DNS by registering wake words. The second server is used as an open source repository for speech recognition software, where external contributions are encouraged. A server is used to store audio samples for deep learning training, while a third server is used to sanitize them, both for the speech recognition repository and public download. Finally, website and mobile applications can be provided to serve as a gateway for the public to upload audio samples, register wake words, contribute to the code repository and download all the services explained above. Figure 6 describes the interactions between these computing clusters[3].

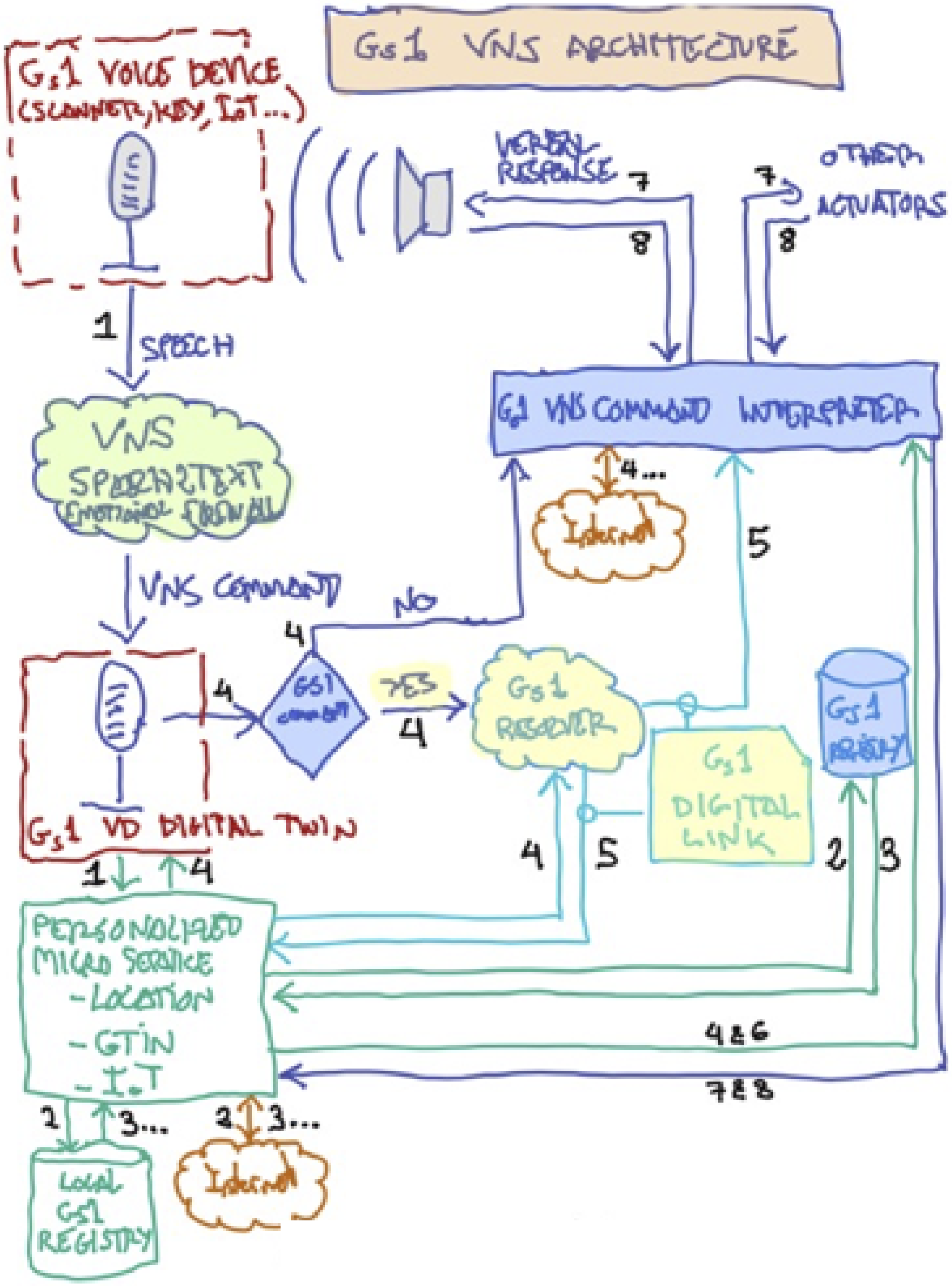

**Figure 6.** GS1 Reference Architecture Suggestion

We now proceed to describe these components of the architecture in greater detail.

[3]Thanks to Phil Archer for his support in understanding the possible role of GS1's Digital Link

## 4.1. *The Voice Name System (VNS)*

Analogous to the DNS on the Internet, this server could consist of a database containing all of the registered wake words as well as who owns them, in addition to a set of standard commands, definitions, settings and modes.

### *4.1.1. Definitions, Modes, Settings and Commands*

Tables 1-2 illustrate a standard set of commands, modes, and definitions that cover a broad range of possible situations a user might encounter when interacting with a device. This is just a starting point for the many interactions that could come default with a device.

| Definitions | |
|---|---|
| **Voice Recognition Device** | Any device that can respond to voice input. This could be a Google Home, Amazon Echo, desktop computer with Cortana, smartphone running Siri, or any device running software that allows the user to communicate with a computer using their voice. The primary feature of these devices is the ability to parse and interpret human speech. |
| **Activation Phrase** | A phrase or sound that wakes or commands a device. This could be "OK Google" on the Google Home. An activation phrase is very similar to an activation word. They only differ in the number of words; a wake phrase is greater than one word whereas a wake word is one word. For example, Google chooses to use an activation phrase to wake their smart speaker, "**OK Google**". Whereas Amazon chooses to use an activation word to wake theirs, "**Alexa**". |
| **Activation Action** | An alternative to using a wake word is to perform an action that awakes a voice recognition device. This includes both active and passive actions. Active actions require an intentional movement or change by the individual. Examples include pressing a button, swiping an RFID card, or scanning a fingerprint. Passive movements are based on the change of a continually examined property. This would include using a passive infrared motion sensor to detect movement, or running facial recognition software to continuously scan for users. |
| **Command** | An order that the device is given by the user. Ultimately there could be some commands that are universal across devices and some that are not. Some could come standard with each device, while others could be optional or customizable. This is similar to how there is a standard "cd" command across all Linux devices, but one has to install the "screen" command for it to do anything. Example: "**Order paper towels**" or "**What is the weather?** |
| **Ping** | An action taken by the device that gathers information about the device, user, and connection. This could return information like the WIFI settings, and location details. Example: "Computer, **ping** my device." |
| **Root Bypass** | A method that brings the device from a current state to a default state. If left in given mode by someone else, the current user could use a root bypass to reset the device to standard mode. Similar to running a computer as an administrator, or using the sudo command in a Linux machine. |

Table 1.: A list of preliminary definitions that will be applicable to discussing and researching conversational commerce and the voice space.

| **Modes** | |
|---|---|
| **Conversation** | A mode in which the device is either meant to operate without an activation word or phrase, or the activation word or phrase has already been spoken. In Conversation mode instead of saying “Alexa, what is today’s date?” one could say “What is today’s date?” |
| **Commerce** | A mode in which the device is temporarily controlled by a third party. |
| **Standard** | A default mode in which the device is initially set run. |
| **Voice Recognition** | An application or program that prompts an electronic device to start responding to voice input. |

Table 2.: A mode is a unique operational condition under which a device would respond to given commands differently. This table provides a handful of example modes.

| **Settings** | |
|---|---|
| **Boot** | The process of initializing a device or device mode. |
| **Light** | The visual effects that the device uses. |
| **Privacy** | The options regarding the recording, storage and processing of information collected by the device. |
| **Sound** | The audio that the device plays. |

Table 3.: A setting is the environment in which the device is operating, both local (specific to each device) and global (specific to each type of device). This table displays preliminary settings.

The commands listed above can be broken down into internal and external commands. Internal commands are often settings that tend to be device specific such as turning up volume. An Internal command doesn’t necessarily need to communicate outside of itself to successfully execute the instruction provided by a user. External commands on the other hand could generally involve routing information from the originally-commanded device to a server or other device. Routing information could be inferred based off a user’s inquiry or command, or routing information could be explicitly provided by the user.

| **Factory Commands** | |
|---|---|
| **Track** | This could be used to track a product or item and its status as it progresses through the factory. |
| **Kill** | This command could stop any machinery currently in operation. This allows the voice recognition device to become a verbal kill switch for important equipment. |
| **Speed** | This could be used to increase or decrease the line speed of a product in a production environment. |
| **Display** | This command could return information about the environment. This could be the temperature, humidity, gas concentrations, or electricity consumption. |

Table 4.: This table contains several example commands that could be used in a factory setting. They are used to both gather information from and control the machinery and environment.

The above table should be considered in conjunction with the factory use case outlined in section 2.3. These commands are examples of the implementation of voice recognition devices in a manufacturing plant.

Similarly to the way the DNS is operated in a device, the VNS provides the same capabilities to cache addresses that are used repeatedly by the user. The specifics for this would be determined by the device itself, as IoT devices can have very low storage capacity. We propose a way to bypass inefficient storage of cached addresses by standardizing the creation of macros. This would be tied to a set of commands specific to the VNS, a first approach is described in the table below.

| **Macro Creation Commands** | |
|---|---|
| **Pair [] to []** | A VNS address, which is specified as the first input, is assigned to a personalized wake word, specified in the second input. The user can access this address without querying the VNS servers by simply using the chosen word. |
| **Delete []** | A previously paired wake word is deleted. |
| **Reset** | All paired wake words are deleted. |
| **Define []** | The corresponding paired address to the specified wake word is spoken. If the input is “all”, all addresses and wake words are spoken to the user. |

Table 5.: Proposed Default commands for macro creation.

### *4.1.2. Ownership*

Certain applications, in the grocery shopping use case for example, devices should be configured including all the guidelines above in a seamless way. We propose an ownership command, which ties the device to the commands and activation actions/phrases defined in a specific address. For a specific grocery store company, all devices could be quickly configured by speaking the command *Ownership “Company Registered Brand”*. The device then sets the VNS address as a default for all commands and activation phrases.

Now that we have defined a few potential modes, settings, and definitions, we will move into a section that describes how one might implement an architecture that would support the previously defined items.

### *4.1.3. VNS Architecture*

As mentioned above, the Voice Name System architecture would be based directly on the current Internet DNS architecture.[31][41] We chose to do this because the DNS is a great example of a hierarchical, easily scalable and generalizable communication standard. In this section we give an adapted DNS architecture for an open Voice Name System.

The Internet DNS assigns an IP address to each domain name. Domain names are themselves divided into different subdomains, which are usually divided by a '.' in each address. In our VNS architecture, these dots are represented in the same way. A root server provides the host with an IP address corresponding to a language server. We suggest each language should be handled by a separate cluster of servers. When making the same request to the language server, the host receives an IP address corresponding to a wake server, which classifies different wake words such as "Hey" or "Hello". If no wake word is provided, the server defaults to an empty wake word. This server finally redirects the host to a Proprietary Server, which stores the full name of the domain. In that server, companies or individuals can store default commands, settings, modes and definitions which can be updated locally. Figure 7 shows a representation of this architecture for an example voice request.

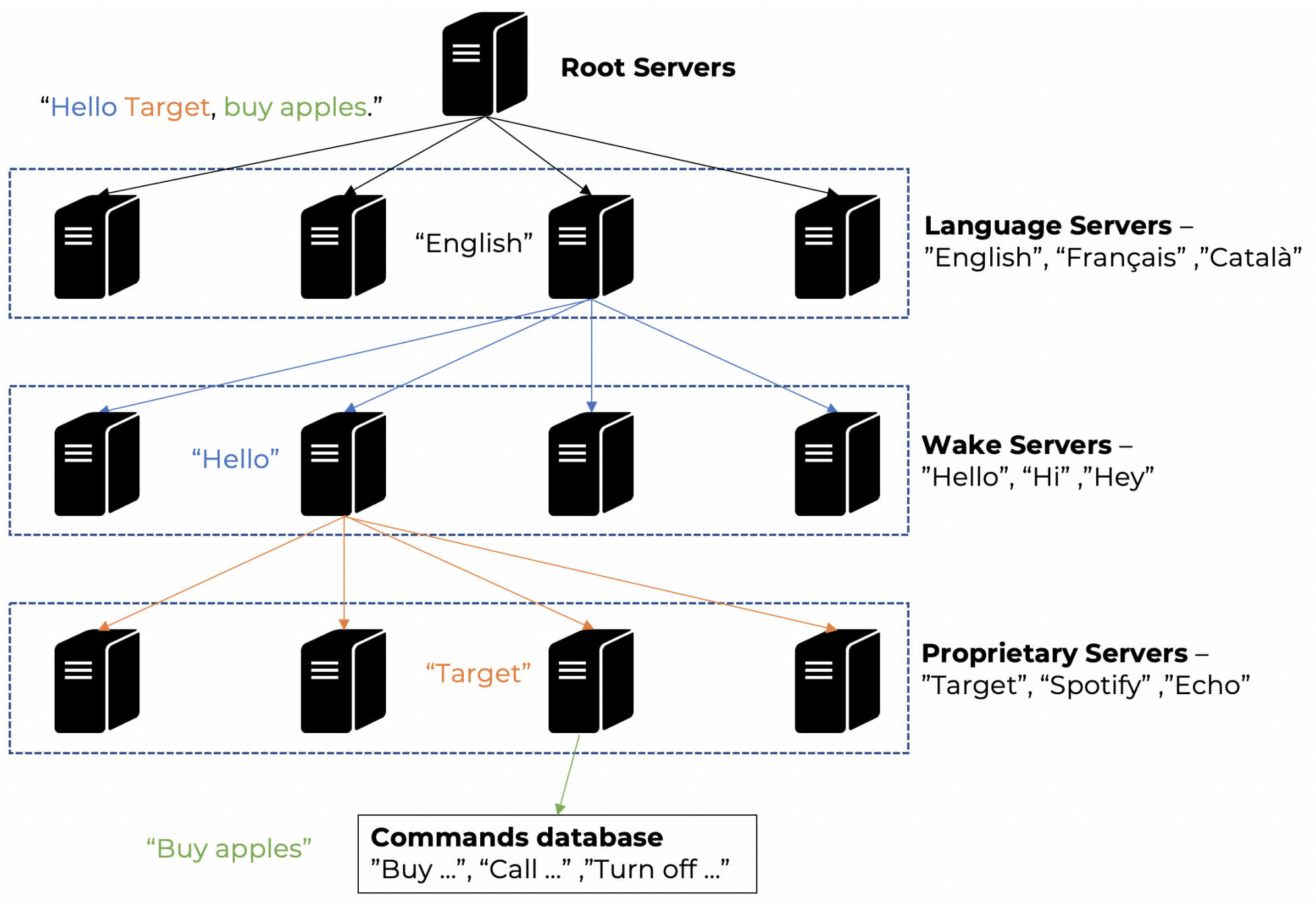

**Figure 7.** Possible VNS architecture following the command *"Hello <Grocery-Store-Name>, buy apples."*, with 4 hierarchical server levels including Root, Language, Wake and Proprietary

This architecture works under the assumption that language has been previously processed either locally or in an external server, which is to say no language processing happens in the servers themselves. Similarly to the way DNS works for websites and as mentioned in the previous sections, the VNS would encourage devices to cache frequent server requests to a local storage in order to speed up the process. Figure 8 shows how VNS would enable a user named Alice to use her Alexander assistant to ask Alexa to perform a task or answer a question.

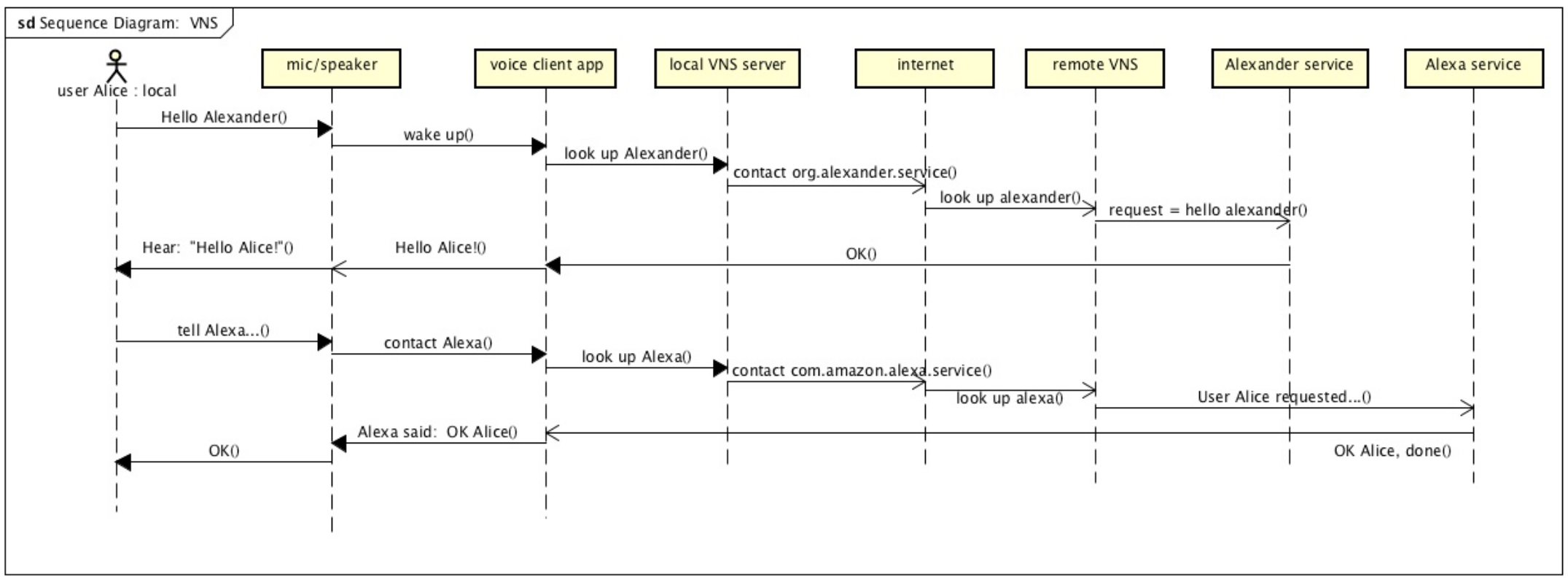

**Figure 8.** Sequence diagram for contacting a remote provider over VNS

| BRAIN UNIT | LAYER NAME | MIT OPEN VOICE | APPLE | AMAZON | GOOGLE | Communication Model Layer |
|---|---|---|---|---|---|---|
| Symbolic Comp. Mdls. | Conversation | Reg. Ex. / DNN | One shot | One shot | One shot | Application Layer |
| Cognitive Models | Use Cases | MIT OV Use Cases | Apps, VC | Skills | Smart Home Action | |
| | Language | HUEY, JAK | SiriKit | AWS Lambda | Oauth 2.0 | |
| Sensory Stream | Objects, People | VNS: CWC, CPF, SVC | Smart Home | Smart Home | Smart Home | Facilitation Layer |
| | Virtual Selves | | Siri | Alexa | OK Google | |
| Brain OS | Device | Sigma Open Source | Apple HW | Echos | Google Home | Connection Layer |
| | Communication | Link with OSI Layers | | | | |

**Figure 9.** Conversational Communications Model

## 4.2. *Conversational Communications Model*

We propose a Conversational Communications Model with the following three layers:

- Application Layer
- Facilitation Layer
- Connection Layer

This structure is an alternative to the representations from the TCP/IP and OSI models [34], and maps to the Brain model of the MIT Center for Brain Mind and Machines[24, 25]. Based on this general model, we present this specific mapping of technologies to the new layer model in Figure 9. The VNS facilitates communications between our new applications and both proprietary and open-source hardware.

## 4.3. *Wake Word Recognition*

A server could be open to the general public for use consisting of trained machine learning models conducting speech-to-text processing to identify words which could be cross-referenced with the wake word repository in compute cluster 1. It would also contain additional software that detects human speech and subsequently collects it. There are many programs that could be added to this software repository to make it more comprehensive such as language translation. We now illustrate how this may be done with a model we developed for the purpose of wake word recognition. Our model is intentionally kept simple because we aim to demonstrate that an open source approach can be the starting point for an elaborated system down the road. The choices we made throughout the design can also help guide a standardization effort.

#### *4.3.1. Objectives*

We designed this model with several objectives in mind. First off, we wanted to prevent collisions between registered wake-up-sentences. In this, our model could be adapted from any other speech to text software currently available in the market. The second objective was to use the same model to detect wake-up-sentences in real time applications on embedded devices. Current speech recognition solutions tend to be dependent on internet connection, all computations are performed on an external server and returned to the embedded device. We wanted with our model to allow for computations to be performed in the device. We made several technical assumptions on the nature of the input audio samples:

- **Audio Sample Rate** - Computers are available to store and play digital audio signals in different formats. Most formats are based on saving the intensity of sound for each period. Frequencies range from 5000 Hz to 50000 Hz depending on the use given to the audio. The sample rate we selected for our model is 16000 Hz as it is the most common in deep learning.
- **Voice Properties** - The human voice of a typical adult has a fundamental frequency from 85 to 255 Hz. However, the harmonics are present for the missing fundamental to create the impression of hearing the fundamental tone. It's important to note that when we hear a constant sound, for example the letter $E$ repeated over time, the recorded audio signal is not a constant but rather a sinusoidal signal of a frequency and all of its harmonics. A very common tool to visualize speech in deep learning is the spectrogram, which represents the intensity of the different sinusoidal frequencies over time. Research has shown that the logarithm of the frequency is even more useful for some representations, mimicking human loss of information in higher frequencies [40]. Since speech recognition aims to mimic how humans interact with each other, we chose to use Mel-Frequency Cepstrums (MFCs) of the audios as input [28], which achieves the characteristics described above.

Similarly, we identified the following challenges:

- **Lightweight Model** - This model needs to be run real time on embedded devices with limited computing power and storage. The model therefore needs to have a limited number of outputs, neural nodes and layers to allow for good performance.
- **Scalability** - The model should be able to recognize thousands of wake words and phrases that are added as they are registered in the VNS. Preferably, the number of output nodes won't change, so as to prevent the need of retraining the model.
- **Background Cancellation** - The model needs to be able to filter out background noise and conversation so as to minimize the number of false positives.

With these parameters in mind, we designed the following model.

### *4.3.2. Model Properties and Structure*

Given two audios, the model classifies whether two words are the same (outputting a 1) or different (outputting a 0). This allows for a very simple and straightforward way of registering a new wake-up sentence, without necessarily having to train the model again as a new output node is created.

The model itself is formed by an encoding section, which is performed both on the input and reference samples, and an inference section, which combines both outputs and analyzes them for similarities. The encoding section uses as input the sequential time windows of the Mel-Frequency Cepstrums of the samples. Two convolutional layers are combined with a dense layer and an LSTM layer. It's very important to point out that the weights for the encoding portion of the model are share between the sample and the input branches. The inference section of the model is simply a set of dense layers, which compare the two encoded audios. The structure can be seen in figure 10. For real time analysis, the reference audio encoding can be computed beforehand in an external server. Only the input audio needs to be computed locally. This can be done with real-time uninterrupted recording thanks to the LSTM layer, making the model a Recurrent Neural Network. This method saves the RNN's last state and feeds it back as an input in conjunction with the new audio recording. This method allows for a very efficient way of computing sample comparisons. The model then only has to compare both encoded audio samples using the two dense layers in the inference portion.

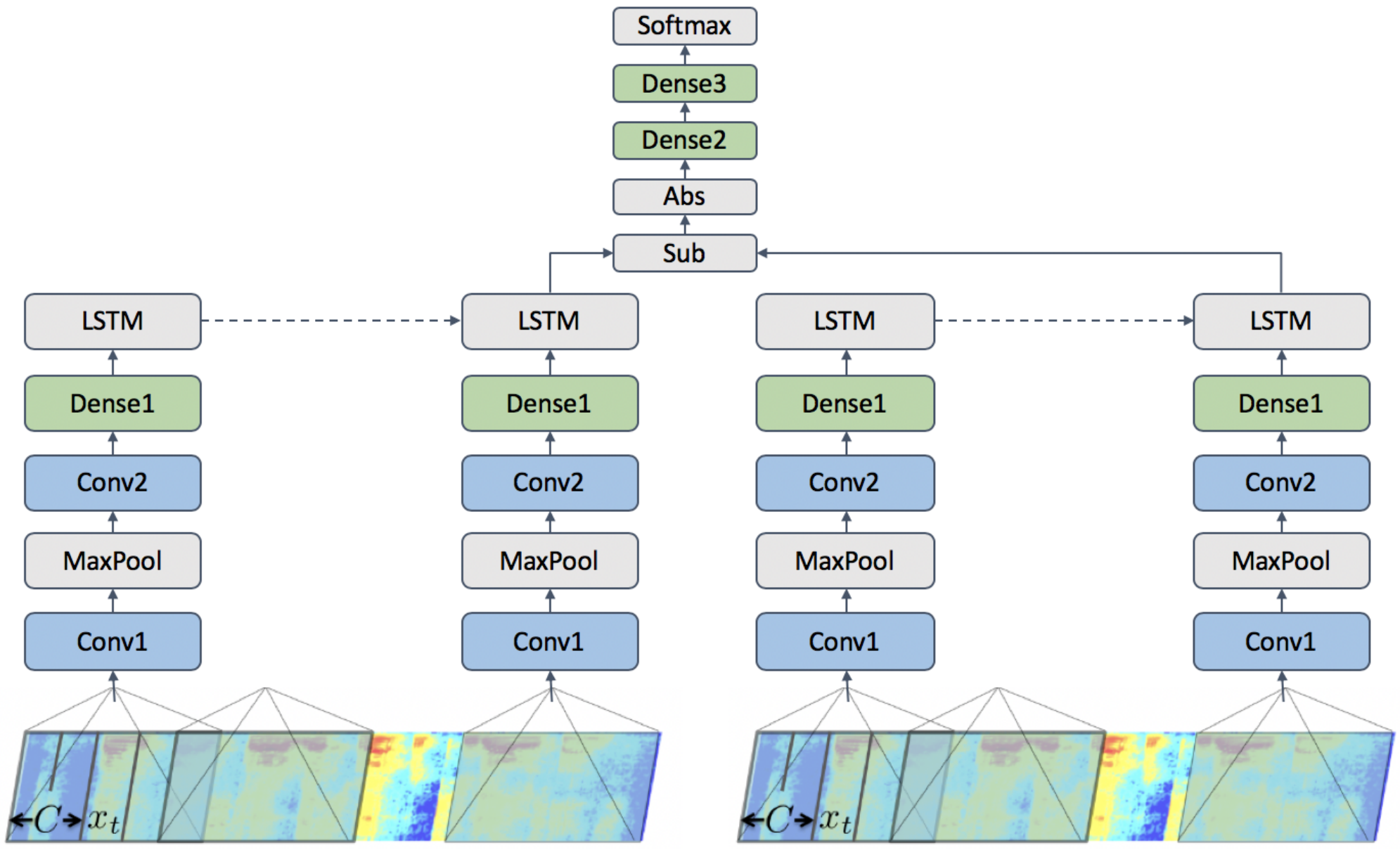

**Figure 10.** Box diagram of the model

### *4.3.3. Model Training and Performance*

We used a data set of 2000 spectrograms per word for 10 words: *[zero, one, two, three, four, five, six ,seven, eight, nine]*. For each iteration, we fit the model with a batch of 200 spectrogram pairs, specifying weather they were the same word or not. These pairs were selected such that there were the same number of correct and false pairs and equally distributed in classes. The training process took 2h 13min and 3730 iterations. We used an early stop technique with a validation set to avoid over-fitting. We used a cross entropy loss function [47] and Adam optimizer [22].

The data was obtained from a Kaggle competition [6]. The objective of it was to classify one second words into classes. The winner team obtained a categorization accuracy score of 91%. The vocabulary words they had to classify where *[yes, no, up, down, left, right, on, off, stop, go]*.

We obtained a score of slightly below 95% (see fig. 11) in the task of classifying whether the word was the same or not over the digits set. We are not working under the same set nor under the same task so we should take this comparison just as a reference.

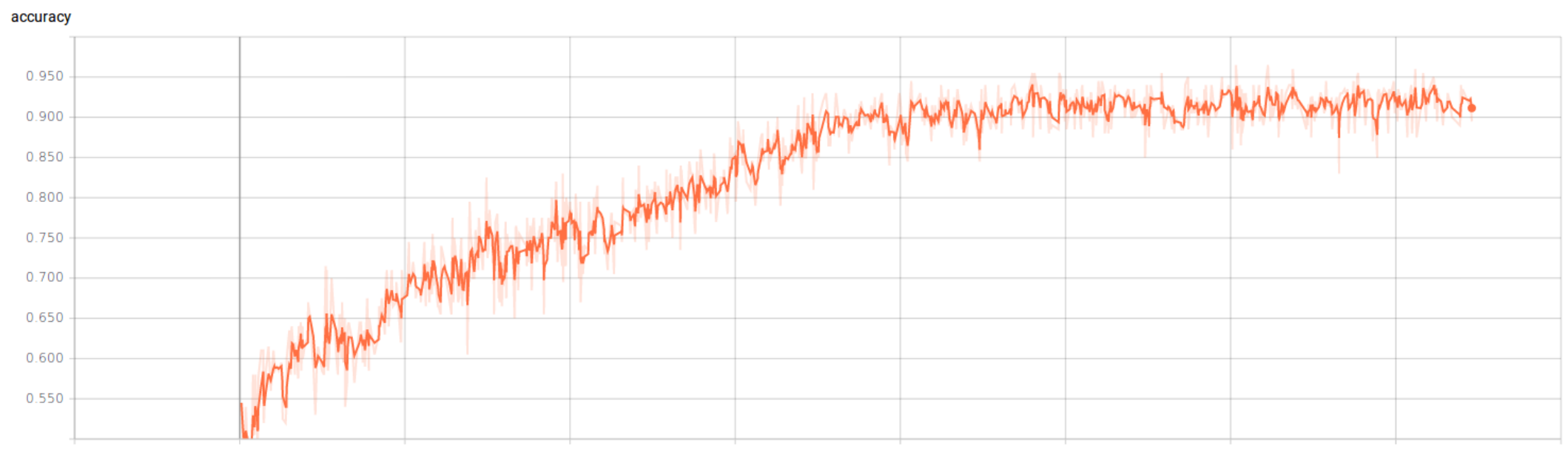

**Figure 11.** Model test data accuracy

### *4.3.4. Further research and scalability*

This model is only a proof of concept, a stepping stone for a permanent solution to wake word recognition. Eventually, we need a model capable of scaling to thousands of wake-up-sentences that can deal with different languages and accents - something that is probably not yet within reach of open source approaches. Currently we don't even have data of multiple languages and accents so we cannot even validate a model. Our model is designed to be very scalable, but, most likely many modifications and a lot of tinkering might be necessary in the future depending on performance.

On a side note, all tests were performed on laptops and desktop computers. One of the main objectives for this model is for it to be able to run real time in embedded systems. Further testing is also needed in this direction. Finally, the model fails to obtain accurate results when none of the wake-up-sentences is spoken. We are currently working on developing a function that returns the model to its initial state when background noise or conversation is being recorded.

### *4.3.5. Software architecture*

We have started the work to create a reference architecture for a vendor-neutral voice service, designed in Java. The Huey Voice Server will be available for downloading from a public repository, so developers may use this system as a library for creating voice applications.

We designed a standardized process for designing and using grammars in the service, as shown in Figure 12.

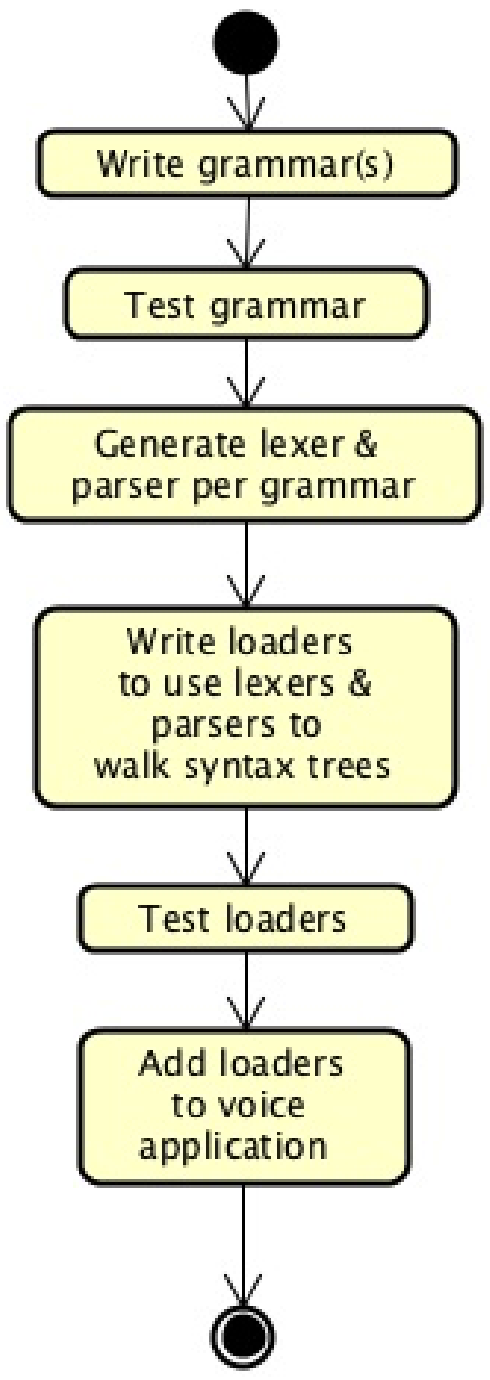

**Figure 12.** Activity diagram showing grammar and software development process

Each use case requires one or more grammars to support the correct processing of spoken or typed user commands. Even for apparently simple use cases, each grammar can be quite lengthy, as shown in the detailed grammar for the shopping list:

```
<input> ::= <meta_command> | <statement>
<meta_command> ::= (<meta_verb> <determiner>? <shopping_list>) | <exit_command>
<statement> ::= <ws_opt>
  (<select_statement> | <delete_statement> | <add_statement> | <create_statement>)
  <ws_opt>
<add_statement> ::= <add> <qty_optional> <item>
<select_statement> ::= <select> <small_number> | <select> <select_scope>
<delete_statement> ::= <delete> <small_number>
<create_statement> ::= <create> <shopping_list>
<select_scope> ::= "all" | "none" | <select_relative>
<select_relative> ::= (<determiner>? <position>) |
  (<small_number> "from"? <to> <small_number>)
<position> ::= "first" "one"? | "last" "one"? | <small_number>
<to> ::= "to" | "through" | "thru" | "-"
<meta_verb> ::= <open> | <save> | <share> | <send> | <print>
<exit_command> ::= <bye>
<determiner> ::= "a" | "an" | "this" | "the" | "my"
<add> ::= "add" | "append"
<delete> ::= "delete" | "remove"
<select> ::= "select" | "highlight"
<qty_optional> ::= <small_number> | " "
<shopping_list> ::= ("shopping" | "grocery" | "supermarket") "list"
<item> ::= <identifier> | (<identifier> <identifier>
  | <identifier> <identifier> <identifier>
  | <identifier> <identifier> <identifier> <identifier>)
<create> ::= ("create" | "make") <determiner>? <new_optional>
<open> ::= "open"
<save> ::= "save"
<share> ::= "share"
<send> ::= "send" | "mail" | "email"
<print> ::= "print" | "display" | "show"
<bye> ::= "bye" | "goodbye" | "good bye" | "exit" | "close"
<new_optional> ::= "new" |
<identifier> ::= <letter> <char>*
<char> ::= <letter> | <digit>
<small_number> ::= (<digit_no_zero> <digit> <digit> <digit> <digit>)
  | (<digit_no_zero> <digit> <digit> <digit>) | (<digit_no_zero> <digit> <digit>)
  | (<digit_no_zero> <digit>) | <digit_no_zero> | <digit_zero>
<digit_zero> ::= "0"
<digit_no_zero> ::= [1-9]
<digit> ::= [0-9]
<letter> ::= ( [a-z] | [A-Z] )
<ws> ::= " "+
<ws_opt> ::= " "*
```

(Whitespace tokens have been omitted from some statements above for clarity.)

Once the grammar has been drafted, an online tool (e.g. https://bottlecaps.de/rr/ui) or desktop application can draw syntax diagrams from the grammar, as shown in Figure 13.

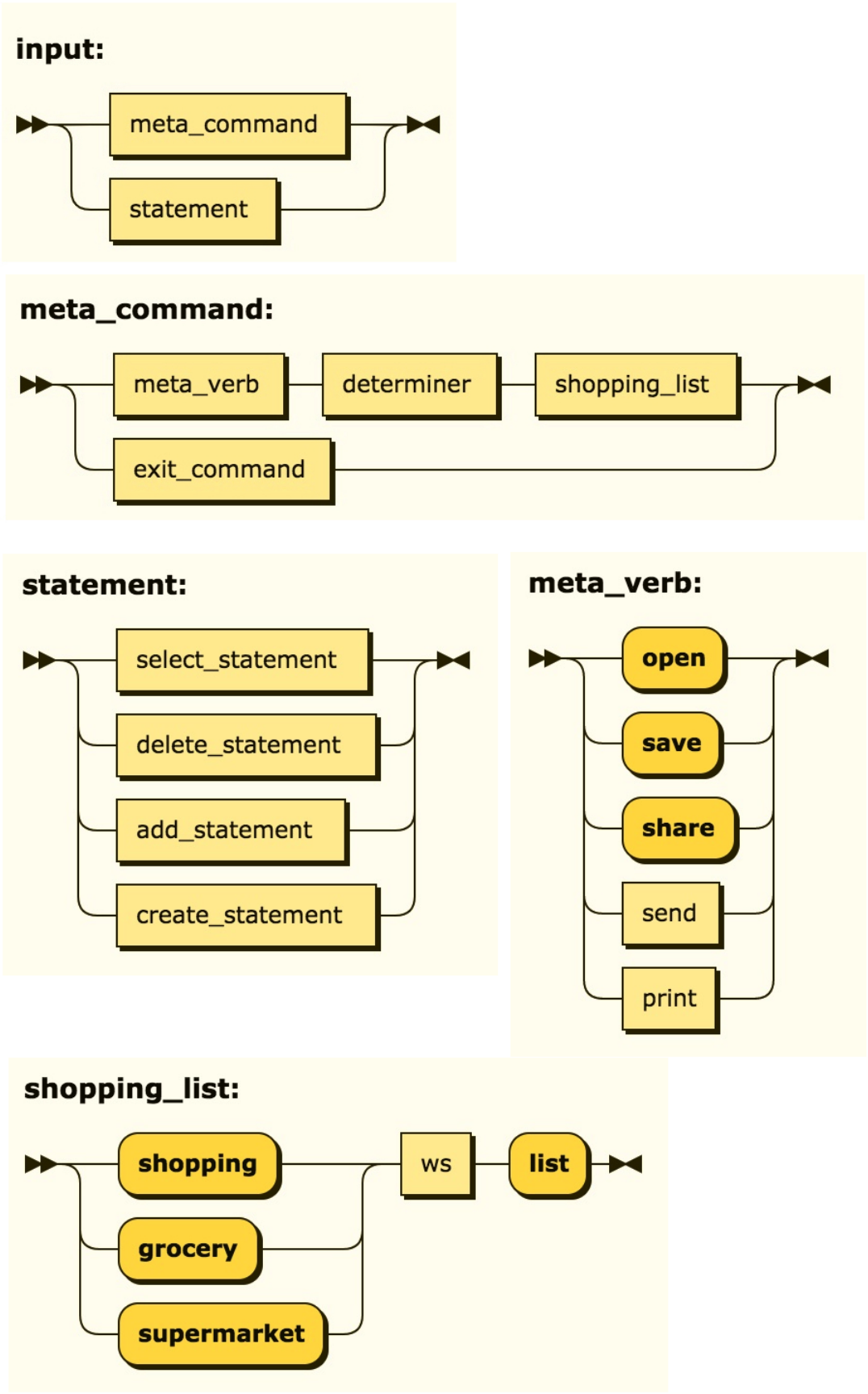

**Figure 13.** Railroad style syntax diagrams for the shopping list use case

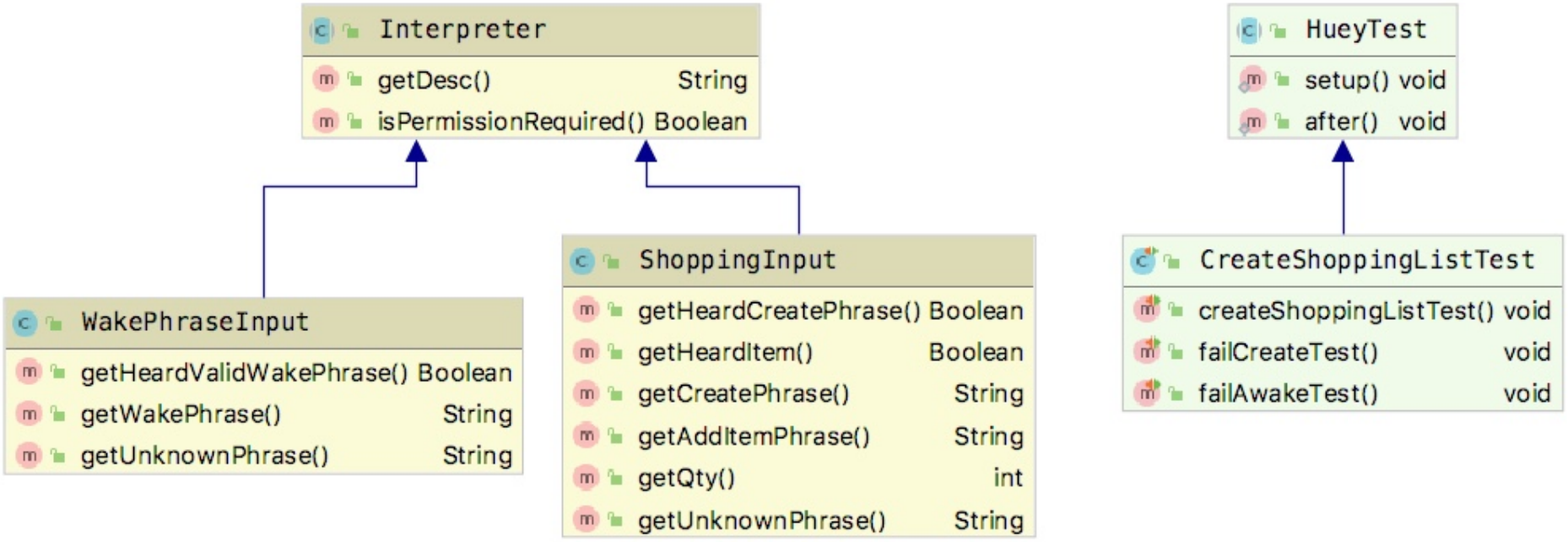

**Figure 14.** Partial class diagram for the Java implementation of a shopping list program

Developing working software to support the use case is an iterative process. One may start by drafting the *ontology* for the use case: a vocabulary that is specific for the task at hand. In our shopping list use case, after waking up a voice assistant, the user might say “create a new shopping list.” This would be followed by commands such as “add carrots” and so on. By listing sample statements like this, patterns emerge regarding the command structures and the words and phrases that form the lexicon within the ontology.

Because it’s very difficult to create software to understand all permutations of a human’s command (or sequence of commands), by necessity the designer must specify the structure or syntax of the commands. This constraint allows us to document the format of understandable commands, and write the grammars needed as the basis for the software which will interpret the commands and carry out actions. As mentioned above, the programmer must carry out a sequence of design, coding, and testing tasks to build up the code required to interpret the commands. Fortunately there are a variety of tools to generate lexers and parsers from grammars written in standard syntax. However eventually some hand-coding is required to integrate generated code with the main part of the application. An example of an implementation of the shopping list is shown in Figure 14.

In addition to the complexity just described, expanding the software functionality to include users who speak different languages or dialects becomes a significant parallel project. For example, the American shopping list program would need “cloning” and modification of its ontology to work in another place such as England or Spain. We cannot simply translate words or phrases one-for-one to the target language, due to syntactic and cultural differences between locales.

Some use cases also require programming features for safety, security, and user permissions. Consider a medical use case related to the Covid-19 pandemic, where a user wishes to take a medical test to find out if they have contracted the disease. During the voice interaction, the user might want to take a test, for valid health concerns. However for medical and legal liability reasons, it might not be permissable for a patient to simply order the test for themselves.

The *Decorator* software design pattern helps solve this problem. In our application, we indicate that some operations require additional permissions, and when this juncture is reached in a conversation, can print or speak a notice to the patient that additional approvals. Figure 15 shows a sample implementation using this design pattern.

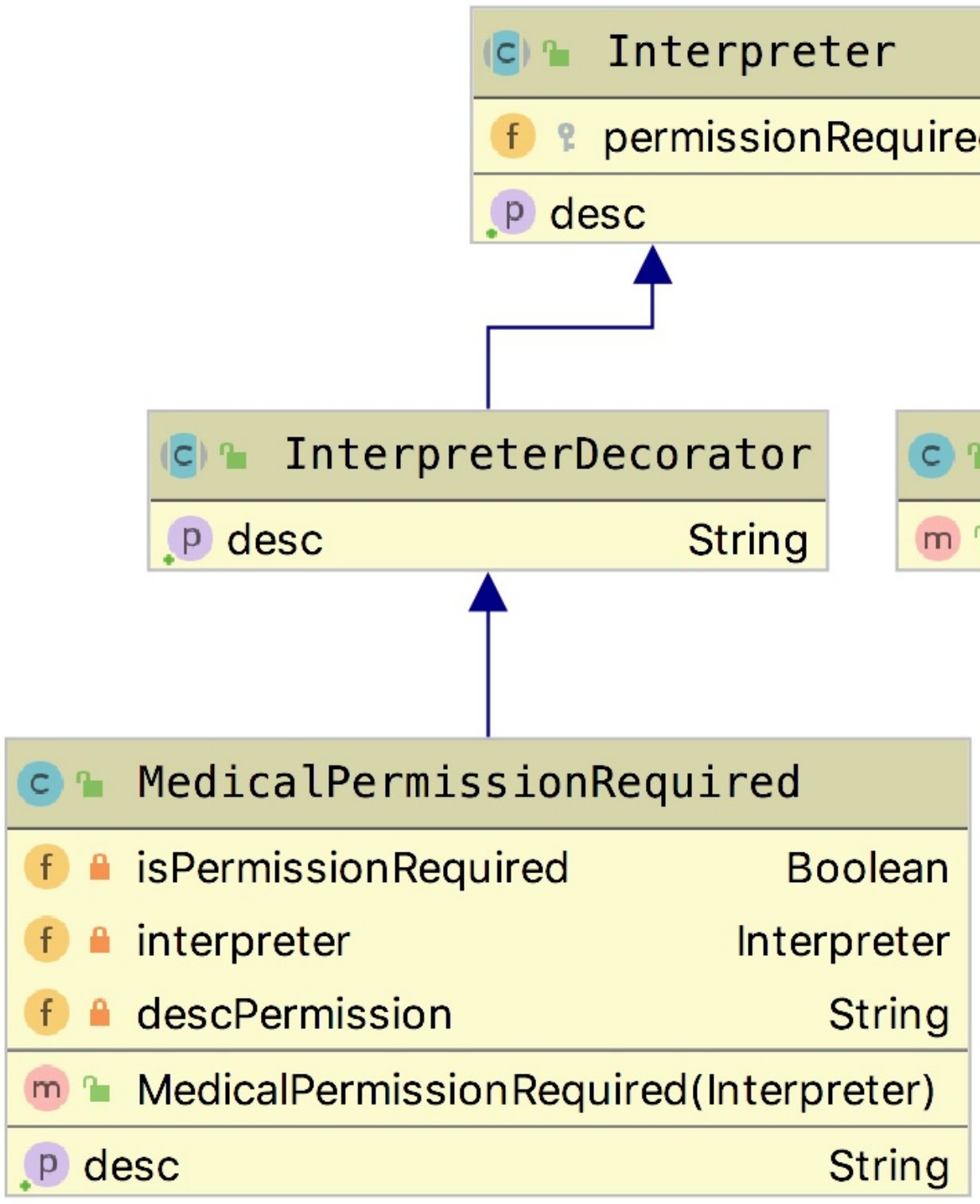

**Figure 15.** Using the Decorator design pattern to secure sensitive operations

## 4.4. *Speech Sample Storage*

This server would be used following a standardized method of uploading audio samples to the network. This would be done either through phone applications, website interfaces, desktop apps or hardware, through a front end gateway. These samples would consist of clearly labeled recordings of a selected wake word. Quality would not be checked for in this server. This server should evidently be prepared to deal with attacks, as well as possible hardware malfunctions, and be independent from any Cloud Services Provider. We propose in figure 16 an example for a potential architecture scheme.

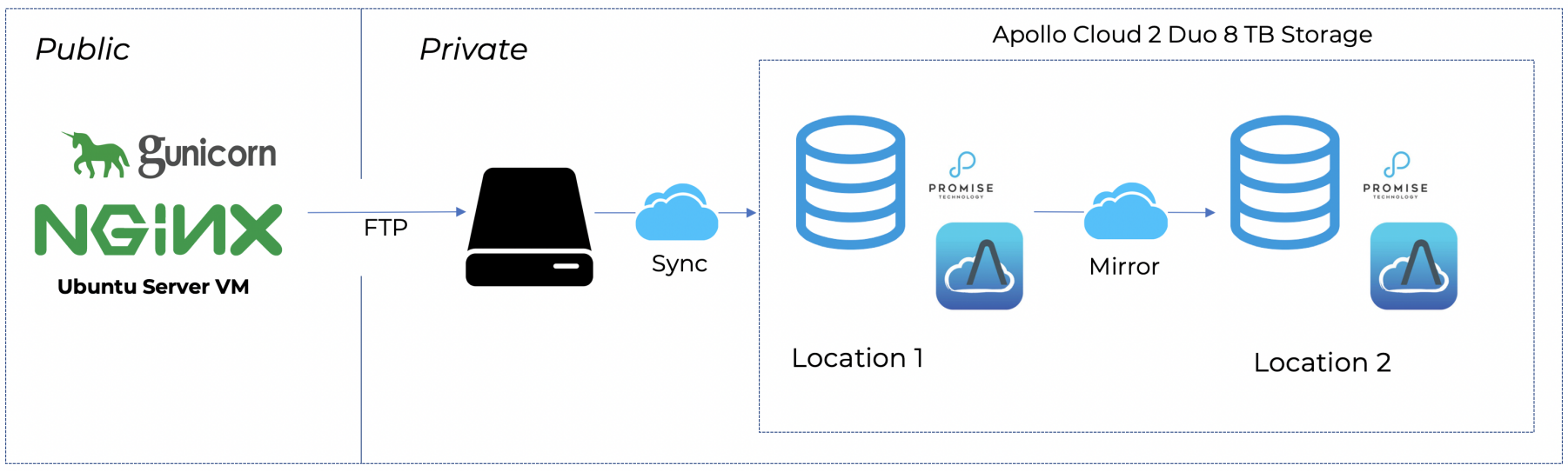

**Figure 16.** Proposed architecture including a public Ubuntu server, a private server, and a network of mirrored Network Attached Storage units in different locations. This architecture is easily scalable and robust to hardware failure or cyberattacks.

All data is uploaded to the public server through parsed HTTP or HTTPS requests. It is organized in hashed folders created daily and stored chronologically. A database tracks where these files are stored, as well as the characteristics from each one. No user information is stored. A daily FTP request is executed from a private server to obtain the data from one of the hashed folders. The contents in that server are synced automatically to a network of Apollo Cloud NAS distributed worldwide. This architecture responds well to the following threats:

- **Cyber Attack to the public server** - If the public server is compromised, the data will not be immediately corrupted throughout the rest of the network as the syncs only happen once a day. Furthermore, only data from that day is downloaded, so if the attack is reported after the sync happens, the corrupted folder can be identified through its unique hash and deleted.
- **FTP request failure** - If the FTP request fails halfway, the incomplete folder and files can both be identified and deleted from the private server and NAS network before repeating the procedure.
- **Hardware Failure** - Because of the high degree of storage redundancy in this architecture, if either the public or private server shuts off, they can both be easily rebooted and updated to the previous day's databases. If any of the hubs in the NAS network fails, the information would still be safe in the network. By delocalizing the NAS, environmental conditions or accidents are not a threat to the network's robustness.

Another advantage of this architecture is that it is extremely scalable, by simply increasing the available storage in the NAS network. We propose this structure to be mimicked by the sanitized speech storage computing cluster discussed below.

### 4.5. *Sanitized Speech Sample Storage*

This computer cluster would include cleaned and integrated speech samples to be used by organizations and researchers to build new machine learning models to further upgrade this technology. This repository will act like the MNIST database for handwritten digits and could also serve as a future foundation for benchmarks. These sanitized speech samples would come from the speech sample storage server discussed above, checked for the following quality indicators.

- **Standard Open Voice Format** - All audio files should be in wav format. Another interesting option would be to develop a file format specific for the standard which would allow to label the wake word, language and accent, and even provide multimedia representations such as static image, video and textual description.
- **Audio Length** - Audio samples would be checked for noise so that only the labeled word or sentences are contained in the sample. This could be achieved with either currently available speech recognition software to parse the samples, or mechanical turks manually reviewing them.
- **Volume and Clarity** - Audio samples should be checked for audio clarity. This could be achieved either by once again employing mechanical turks, or automating the process by identifying the amplitude peaks and minimums of each sample. It is important to note that not all recordings should be flawless, as deep learning networks would benefit from more challenging training data points.

Each training data set, comprised of a set number of checked audio samples, would be assigned to a hash block. This block would identify the institution or individual who performed the quality check. Being checked by a reputable institution or by multiple entities would indicate a healthy training set.

### 4.6. *Front End Gateway*

Finally, users could interface with the system using a front end gateway that would allow them to generate, access and download audio samples and open source code. We envision such gateways serving as access points for new developers attempting to become members of the standard. By providing access to data and code through the gateway, we hope to enable small organizations and teams to easily develop capabilities in the voice recognition space. This is particularly important because smaller organizations will find it difficult to enter this space without reliance on major cloud providers. The frontend gateway can help to alleviate this challenge by allowing developers to obtain data in order to create their own deep learning models for voice recognition. The gateway would allow users to contribute recordings of themselves to be used as training data, and help to build a open source repository of reliable samples that organizations can freely use.

Ideally, this gateway could be accessed through whatever platform the user finds convenient, particularly through any web browser or mobile application. Though application developers are trained to develop applications that work across browsers and platforms, it is worth noting that developing cross-browser audio recording functionality can introduce additional challenges. Certain browsers do not support popular Application Programming Interfaces (APIs) for recording, and mobile device browsers may have different support for such APIs than their desktop counterparts. To aid developers, we tested major browsers across devices and compiled the following figure, detailing what APIs can be used across major browsers and operating systems.

We used this information when building our own architecture for recording, consisting of iOS and Android applications and a website supporting cross browser recording on most major platforms.

Tables 7 and 8 specify all open software and versions we used to implement the code for all the items discussed above, including front end software based on a React/React-Native environment, and a Python/Flask backend.

| **Front-End Open Source Software** | |
|---|---|
| **React** | Version 16.02 for Web Version - Version 16.4.1 for iOS and Android<br>Downloaded from *https://reactjs.org/* |
| **ReactStrap** | Version 6.5.0<br>Downloaded from *https://reactstrap.github.io/* |
| **Bootstrap** | Version 4.1.3<br>Downloaded from *https://getbootstrap.com/* |
| **MobX** | Version 5.8.0<br>Downloaded from *https://github.com/mobxjs/mobx* |
| **React Native** | Version 0.56.1<br>Downloaded from *https://facebook.github.io/react-native/* |
| **NodeJS** | Version 6.4.1<br>Downloaded from *https://nodejs.org/en/* |
| **Android Studio** | Version 3.3 |
| | Downloaded from *https://developer.android.com/studio/* |
| **Xcode 10** | Version 10.1<br>Downloaded from *https://developer.apple.com/xcode/* |
| **Audio Recorder Polyfill** | Version 0.1.3<br><br>Downloaded from *https://github.com/ai/audio-recorder-polyfill* |
| **RC-slider** | Version 8.6.4<br>Downloaded from *https://www.npmjs.com/package/rc-slider* |
| **React Device Detect** | Version 1.6.2<br>Downloaded from *https://www.npmjs.com/package/react-device-detect* |
| **React Router** | Version 4.3.1<br>Downloaded from *https://github.com/ReactTraining/react-router* |
| **FontAwesome** | Version 4.7.0<br>Downloaded from *https://origin.fontawesome.com/* |
| **React Native Circular Progress** | Version 1.0.1<br><br>Downloaded from *https://github.com/bartgryszko/react-native-circular-progress* |
| **React Native Vector Icons** | Version 5.0.0<br><br>Downloaded from *https://github.com/oblador/react-native-vector-icons* |
| **React Native Fetch Blob** | Version 0.10.13<br><br>Downloaded from *https://www.npmjs.com/package/rn-fetch-blob* |
| **React Native Audio** | Version 4.1.3<br>Downloaded from *https://github.com/jsierles/react-native-audio* |
| **React Native Sound** | Version 0.10.9<br>Downloaded from *https://github.com/zmxv/react-native-sound* |
| **React Native Splash Screen** | Version 3.1.1<br><br>Downloaded from *https://github.com/crazycodeboy/react-native-splash-screen* |
| **React Native SVG** | Version 6.5.2<br>Downloaded from *https://github.com/react-native-community/react-native-svg* |
| **React Native Popup Dialog** | Version 0.16.6<br><br>Downloaded from *https://github.com/jacklam718/react-native-popup-dialog* |

Table 7.: A list of open source front-end software used in the project to deploy the different components of the standard discussed above.

| Back-End Open Source Software | |
|---|---|
| **Python** | Version 3.5.2<br>Downloaded from *https://www.python.org/* |
| **Wheel** | Version 0.32.2<br>Downloaded from *https://pythonwheels.com/* |
| **Flask** | Version 1.0.2<br>Downloaded from *http://flask.pocoo.org/* |
| **Pydub** | Version 0.23.0<br>Downloaded from *https://github.com/jiaaro/pydub* |
| **Gunicorn** | Version 19.9.0<br>Downloaded from *https://gunicorn.org/* |
| **Numpy** | Version 1.15.4<br>Downloaded from *http://www.numpy.org/* |
| **Nginx** | Version 1.10.3<br>Downloaded from *https://www.nginx.com/* |
| **Supervisor** | Version 3.2.0<br>Downloaded from *http://supervisord.org/* |
| **Tensorflow** | Version 1.12.0<br>Downloaded from *https://www.tensorflow.org/* |
| **Keras** | Version 2.1.6<br>Downloaded from *https://keras.io/* |
| **CMUSphinx** | Downloaded from *https://cmusphinx.github.io/* |
| **DeepSpeech** | Downloaded from *https://github.com/mozilla/DeepSpeech* |

Table 8.: A list of open source back-end software used in the project to deploy the different components of the standard discussed above.

Having considered an architecture of computing clusters as a potential solution to implementing the open voice standard, we will now propose a reference architecture approved by GS1, the worldwide institution for barcodes.

### 4.7. *GS1 Reference Architecture*

We now introduce a sample architecture, developed around GS1's resolver structure, which grants access to product information through GTINs (bar-codes), as seen in figure 5. The first component of this system is a voice capturing device (labeled as GS1 voice device).

We propose that the voice capturing device operates based on a short, command-like syntax. Though voice recognition platforms have recently been notable for their ability to capably understand natural communication, we are of the opinion that this is not in the best interest of the end user. While the use of rigid commands for interfacing with devices may be less convenient than speaking naturally, we believe that privacy concerns greatly outweigh such considerations. Recent research [cite] has shown that it is possible to apply machine learning to speech data in order to detect symptoms of mental conditions such as depression. While such technology has the potential to be beneficial, it also increases the potential for intrusion into users' personal lives by anyone with access to their speech data. A robust command-like syntax could combine the benefits of natural language processing to reduce rigidity and allow for some variations in phrasing, while simultaneously limiting conversation to eliminate unnecessary data collection.

| **Platform** | **Browser** | **getUserMedia** | **MediaRecorder** | **MediaRecorder Polyfill** | **AudioContext** |
|---|---|---|---|---|---|
| MacOS Mojave | Google Chrome | + OS: v10.14.1<br>Browser: v71.0.3578.98 | + OS: v10.14.1<br>Browser: v71.0.3578.98 | N/A | + OS: v10.14.1<br>Browser: v71.0.3578.98 |
| MacOS Mojave | Mozilla Firefox | + OS: v10.14.1<br>Browser: v63.0.1 | + OS: v10.14.1<br>Browser: v63.0.1 | N/A | + OS: v10.14.1<br>Browser: v63.0.1 |
| MacOS Mojave | Apple Safari | + OS: v10.14.1<br>Browser: v12.0.1 | - OS: v10.14.1<br>Browser: v12.0.1 | + OS: v10.14.1<br>Browser: v12.0.1<br>Polyfill: v0.1.3 | + (WebkitAudioContext)<br>OS: v10.14.1<br>Browser: v12.0.1 |
| MacOS Mojave | Opera | + OS: v10.14.1<br>Browser: v58.0.3135.47 | + OS: v10.14.1<br>Browser: v58.0.3135.47 | N/A | + OS: v10.14.1<br>Browser: v58.0.3135.47 |
| Windows 10 | Google Chrome | + OS: v10.0.17134<br>Browser: v71.0.3578.98 | + OS: v10.0.17134<br>Browser: v71.0.3578.98 | N/A | + OS: v10.0.17134<br>Browser: v71.0.3578.98 |
| Windows 10 | Mozilla Firefox | + OS: v10.0.17134<br>Browser: v64.0.2 | + OS: v10.0.17134<br>Browser: v64.0.2 | N/A | + OS: v10.0.17134<br>Browser: v64.0.2 |
| Windows 10 | Microsoft Edge | + OS: v10.0.17134<br>Browser: v42.17134.1.0 | - OS: v10.0.17134<br>Browser: v42.17134.1.0 | + OS: v10.0.17134<br>Browser: v42.17134.1.0<br>Polyfill: v0.1.3 | + OS: v10.0.17134<br>Browser: v42.17134.1.0 |
| Windows 10 | Opera | + OS: v10.0.17134<br>Browser: v57.0.3098.116 | + OS: v10.0.17134<br>Browser: v57.0.3098.116 | N/A | + OS: v10.0.17134<br>Browser: v57.0.3098.116 |
| Android Nougat | Google Chrome | + OS: v7.1.2<br>Browser: v71.0.3578.99 | + OS: v7.1.2<br>Browser: v71.0.3578.99 | N/A | + OS: v7.1.2<br>Browser: v71.0.3578.99 |
| Android Nougat | Mozilla Firefox | + OS: v7.1.2<br>Browser: v64.0.2 | - OS: v7.1.2<br>Browser: v64.0.2 | + OS: v7.1.2<br>Browser: v64.0.2<br>Polyfill: v0.1.3 | + OS: v7.1.2<br>Browser: v64.0.2 |
| Android Nougat | Microsoft Edge | + OS: v7.1.2<br>Browser: v42.0.2.2864 | - OS: v7.1.2<br>Browser: v42.0.2.2864 | + OS: v7.1.2<br>Browser: v42.0.2.2864<br>Polyfill: v0.1.3 | + OS: v7.1.2<br>Browser: v42.0.2.2864 |
| Android Nougat | Opera | + OS: v7.1.2<br>Browser: v49.2.2361.134358 | - OS: v7.1.2<br>Browser: v49.2.2361.134358 | + OS: v7.1.2<br>Browser: v49.2.2361.134358<br>Polyfill: v0.1.3 | + OS: v7.1.2<br>Browser: v49.2.2361.134358 |
| Android Nougat | Samsung Internet | + OS: v7.1.2<br>Browser: v8.2.01.2 | - OS: v7.1.2<br>Browser: v8.2.01.2 | + OS: v7.1.2<br>Browser: v8.2.01.2<br>Polyfill: v0.1.3 | + OS: v7.1.2<br>Browser: v8.2.01.2 |
| Apple iOS | Apple Safari | + OS: v12.1.2<br>Browser: v12.1.2 | - OS: v12.1.2<br>Browser: v12.1.2 | + OS: v12.1.2<br>Browser: v12.1.2<br>Polyfill: v0.1.3 | + OS: v12.1.2<br>Browser: v12.1.2 |
| Apple iOS | Google Chrome | - OS: v12.1.2<br>Browser: v71.0.3578.89 | - OS: v12.1.2<br>Browser: v71.0.3578.89 | - OS: v12.1.2<br>Browser: v71.0.3578.89<br>Polyfill: v0.1.3 | + OS: v12.1.2<br>Browser: v71.0.3578.89 |
| Apple iOS | Mozilla Firefox | - OS: v12.1.2<br>Browser: v14.0 | - OS: v12.1.2<br>Browser: v14.0 | - OS: v12.1.2<br>Browser: v14.0<br>Polyfill: v0.1.3 | - OS: v12.1.2<br>Browser: v14.0 |
| Apple iOS | Microsoft Edge | - OS: v12.1.2<br>Browser: v42.10.3 | - OS: v12.1.2<br>Browser: v42.10.3 | - OS: v12.1.2<br>Browser: v42.10.3<br>Polyfill: v0.1.3 | - OS: v12.1.2<br>Browser: v42.10.3 |
| Apple iOS | Opera Touch | - OS: v12.1.2<br>Browser: v1.3.1 | - OS: v12.1.2<br>Browser: v1.3.1 | - OS: v12.1.2<br>Browser: v1.3.1<br>Polyfill: v0.1.3 | - OS: v12.1.2<br>Browser: v1.3.1 |

**Table 6.** A table showing the different APIs supported based on device operating system and browser. The data presented was obtained through our testing and use of online resources [13] [11] [12]. The package used to polyfill the MediaRecorder API on unsupported browsers was Audio-Recorder-Polyfill, a free open source package. [16]

# 5. Huey Language Design

## 5.1. *Introduction*

We have designed a new, unambiguous natural language named *Huey*. With this language, we aim to standardize voice interactions to a universal reach similar to that of other systems such as phone numbering. To achieve this we anticipated the various types of requests that humans may make of an intelligent system by speaking, as illustrated in the use cases outlined in Chapter 1.

A user may be speaking to a local system, such as smart home appliances, or to a remote service, to ask questions or make task requests that are known to be handled by other applications.

An ideal intelligent system would perfectly understand, at any moment, what the user's intentions are, and if necessary switch modes to carry out a request within a given context. Unfortunately the technology has not developed to a point where this is possible for a majority of cases, so we take the typical *divide and conquer* approach to break the problem into smaller parts.

## 5.2. *Ontologies*

Multiple ontologies are required to respond to requests that are related to a specific task context, as well as to a global context. Each ontology is comprised of elements touched upon earlier:

- grammar: the rules for forming sentences or phrases
- lexicon: a representative set of words and phrases
- semantics: the meaning of statements or commands
- predicates: conditions that precede actions

Based on the elements above, one or more *actions* are performed by the system in response to an input statement.

Part of building the lexicon and grammar involves selecting *keywords* that have special meanings to an intelligent assistant. These words are listened for carefully when a user speaks a command, to narrow down the possibilities of the intent.

The grammars we design to work within assistant skills as well as outside a particular skill, are generally formulated to follow this pattern:

```
{verb-phrase} {noun-phrase}
```

A *verb-phrase* starts with a keyword serving as a verb. This is the signal to the system to know how to match a request with the templates for commands expressed via one or more grammars.

A *noun-phrase* is comprised of one or more words, identifying a person, thing, concept, etc. It may also contain keywords, depending on the needs of a particular rule.

The keywords were also designed to sound different from each other, at least in American English. This helps with designing the language interpreters to distinguish words more easily.

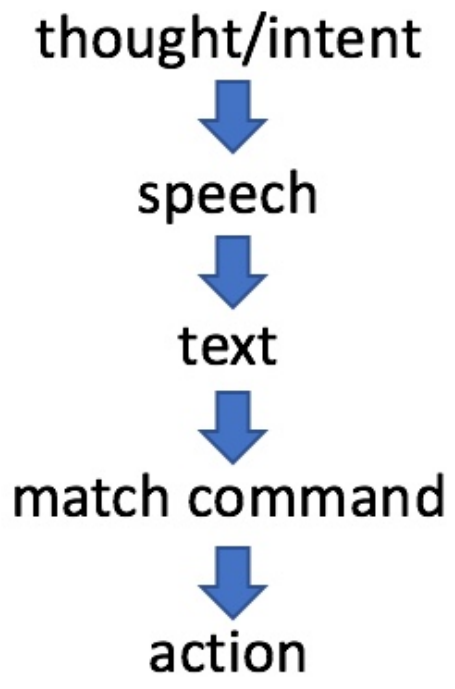

**Figure 17.** Transforming user's thoughts and intents to system actions

## 5.3. *Design Philosophy*

We need to consider that the human speaker will use a kind of natural language, regardless of training or any instructions about how to use this system.

Accordingly, we need to make transformations as shown in Figure 17.

This is an over-simplified version of the processing that takes place, as we might need one or more retry loops to arrive at the correct interpretation and action, including coping with noisy environments. Here are capsule descriptions of each phase.

**Thought/Intent.** We cannot ever know the true thought or intent of the user. In fact, the user may be tired and thinking one thing, and say another. What is the true intent in this case?

**Speech.** Despite the previous caveat, we will assume that in most instances the user's speech represents their true intent. There may be background noise mixed in with the speech, or we may have timing issues with acquiring a useful utterance (e.g. a full command, phrase, or sentence as needed). In some cases spoken requests might be typed instead of spoken, however this presents its own challenges (e.g. typos, abbreviations, etc.).

**Text.** Conversion from speech to text can be done with high accuracy in a low-noise environment and other pre-conditions, such as support for spoken dialects.

**Match command for current context.** Ideally Huey needs to find the one command that is the best match for the text received. At any one moment, when interpreting the received text, Huey will have some implicit (inferred) or explicit context for that text. An implicit context can be formed based on a variety of criteria, including:

- the current page being displayed or most recent spoken feedback
- the path the user took to reach the current state
- the most recent few commands heard

An explicit context can be set by the user alternating between voice, keyboard, and mouse, for example to select user interface (UI) elements to choose a mode. Or the user may speak a special command that switches modes, a *meta command*, described below.

Sometimes before proceeding to do one or more actions, Huey will need multiple pieces of information. These data can be placed into a frame based on the most recent user requests and actions. Once a programmed threshold is reached, including confidence of speech-to-text, Huey will judge the current frame complete (or complete enough) to do the corresponding action.

Note that some requests will require confirmation or clarification, e.g. destructive operations on a file system or requests needing approvals from other people or agents.

**Action.** Finally we have collected the necessary info and can perform the requested action. Huey will log all requests, to provide an undo capability.

### 5.4. *Natural vs. Unnatural Language*

Humans do not necessarily speak in robotic patterns, always repeating the same sentences in the same way. There are also stray, non-word sounds, and background noise in most environments. Word usage may vary a lot from time to time, even when expressing an identical request given on another day.

While our grammars and lexicons are tuned to hear requests in the command format given above, we also design for more natural language forms such as

```
{extra-words} {verb-phrase} {more-words} {noun-phrase} {etc.}
```

### 5.5. *User Interface Variability*

Some applications will have a visual display capability, such as on a mobile device, laptop, or dedicated digital assistant. Others may have no text display and offer only auditory feedback, or sound plus some kind of light or motion display. The innovations and standards discussed here are broad enough to encompass a variety of usage situations. Programmers who implement digital assistant programs will need to provide the typical, main expected output, while also considering addition of fallback mechanisms to provide feedback under device constraints. For example, without a visual display, the system will be in text-to-speech mode by default. So care must be taken to avoid flooding the user with too much spoken text at any one time.

### 5.6. *Meta Commands*

A *meta command* is a user request that is either global in nature, or relevant to an entire document or other composite object. They may also serve as directives to change context from the current one to another.

Huey provides the capability to define commands using the following keywords, to let the user request something “above” the task level. These are intended to be short and intuitive, so relatively easy to remember. They also closely model commands familiar to many users, e.g. from using common desktop document applications.

- open: open a document, connection
- kill: immediately halt a request in progress
- disconnect: break the connection with a system
- save: save a document, note, etc.

- edit: change the current or specified item
- create: make a new document, entry, or object
- find: locate an existing document or system
- send: email or forward information to a person or system
- list: speak or show existing items
- rename: change the name of an existing item
- run: execute a batch job, launch a program, etc.
- undo: go back one or more steps to a previous state
- help: get help on a specified topic, or get the index

Depending on the current assistant the user is communicating with (e.g. the one nearest at the moment), the commands will be interpreted as necessary to relate to the current or most recent context. Some commands will require additional arguments – for example *save* may first be spoken as *save as [name]*.

## 5.7. *Navigation Commands*

A *navigation command* is a user request that is specific to a current context, and enables choosing one or more items or going to another part of a list of items. For a visual UI, the user will see feedback such as highlighting or borders around parts of a page. For a non-visual UI, the skill may announce confirmation that an item has been selected etc. Not all keywords below will make sense in each output modality.

- select: choose one or more items
- next: go to the next item or group of items
- previous: go back to an earlier item or group of items
- check: mark a checkbox or radio button, or confirm an audio prompt
- erase: blank out a field, or the currently selected item
- uncheck: unmark a checkbox or radio button
- field: jump to a named field or page
- go to: go to a specific place in a document or list
- locate: search for something in the current document or list

Navigation commands will be most useful for operating devices with display, in an entirely or mostly hands-free fashion.

## 5.8. *Huey Language Specification*

We have designed a new, unambiguous natural language named *Huey.* With this language, we aim to standardize voice interactions to a universal reach similar to that of other systems such as phone numbering. To achieve this we anticipated the various types of requests that humans may make of an intelligent system by speaking, as illustrated in the use cases described above.

See [27] for a complete language specification for Huey.

## 5.9. *Huey Use Cases*

## 5.10. *Introduction*

In this section, we examine a variety of use cases that may be handled by Huey with Jak. We show how we can use the grammars of a Huey system, both the standard base sets plus additional ones designed for specific use cases.

### *5.10.1. Switching Assistants*

Near the beginning of this paper, we provided a long natural language request scenario – a more complex version of the example just described above. The scenario imagined a user making multiple requests in a short span of time, talking with several intelligent assistants to add items to a shopping list, add information to an expense report, set an alarm with music, and turn on some lights.

To program this using currently available speech dialog tools would involve a great deal of programming. Furthermore, this custom code might not be easily expandable to use cases without significant rewriting. Also, as discussed in this paper, most assistants currently will handle a one-shot request or a request with a short follow-up confirmation. E.g., if you ask Alexa to set your alarm for 7am tomorrow, she may ask if you want to make this your standard alarm. At this point you can answer, and the conversation is over. To make another request, you need to wake the assistant again.

We would like to provide a better user experience, allowing more natural language, and design mechanisms to freely switch from one assistant to another. (Assuming the user has already set up accounts for different services.)

Although our long request is whimsical, here we present some grammars to show how a Huey application can understand complicated requests, and use switch to alternate assistants that are registered in the VNS.

```
<input> ::= <meta> | <meta_shop> | <meta_sheet> | <meta> <stmt>
  | <meta_other> <stmt> | <meta_other> | <stmt>

<stmt> ::= <stmt_sheet> | <stmt_expense>
  | <stmt_shop> <and> <stmt_shop> | <stmt_shop>
  | <meta> <ignore>* <meta_sheet> <from>
      <det>? <topic_expense>? <topic_expense>? <template>
  | <tell_assistant> | <tell_assistant_compound>

<meta_other> ::= <ignore>* <connect> <assistant> <for_of>
     <det>? <skill_remote> <ignore>?
  | <ignore>* <wake>? <connect>? <assistant> <for_of>?
     <det>? <skill_remote> <ignore>?
  | <ignore>* <wake>? <connect>? <assistant> <ignore>?
```

The grammar shown above can act as a *root grammar*, the top of a syntax tree for a natural language understanding system to hear a variety of requests after the user's initial wake command.

When the user says they want to add an item to a specific shopping list, they trigger the `<stmt_shop>` case included above, which leads to the shopping list grammar (see ShoppingList.bnf in the appendix).

In this example, we hardcoded the names of a few supermarkets, including Lidl, mentioned by the user in our example. In the production application, we would use a more data-driven approach for literal strings of this nature. Also, it is important to note that less common words and phrases (such as "Lidl") may not be understood by generic speech recognition software. In this case we might need to train some recognition models or use phonetic matches to catch unusual sounds and interpret them correctly.

However, via the grammars shown so far, we cleverly provided a specific slot for `<store>` in the `<add_item>` rule. So if the surrounding words are heard by the system, the store name phrase will be passed along for further processing by the application. So even if a spoken "Lidl" is interpreted as "little", we can catch this downstream as long as the correct rule is triggered. Furthermore, recording usage data will allow us to use reinforcement learning to improve the system's accuracy over time.

Returning to our long use case, the user asked to send requests to other assistants registered in the VNS. The grammar below (from RootVNS in the appendix) allows us to understand the request and make the connection:

```
<tell_assistant_compound> ::= <tell_assistant> <and> <tell_assistant>

<tell_assistant> ::= <meta> <ignore>* <connect> <assistant>
  | <connect> <assistant> <service_action>?
  | <tell_assistant_at_time> <service_action>

<tell_assistant_at_time> ::= <meta> <ignore>* <connect> <assistant>
    <day_relative> <time_of_day>? (<for_of> <time_str>)?

<service_action> ::= <to>? <action> <item> <for_of> <service> <music>?
  | <to>? <action> <det>? <room> <thing> <ignore>?

<meta_other> ::= <ignore>* <connect> <det>? <assistant> <for_of>
    <det>? <skill_remote> <ignore>?
  | <ignore>* <wake>? <connect>? <det>? <assistant> "assistant"?
      <robot>? (<for_of|to>)? <det>? <skill_remote> <ignore>?
  | <ignore>* <wake>? <connect>? <assistant> <for_of>? <det>?
      <skill_remote> <ignore>?
  | <ignore>* <wake>? <connect>? <assistant> <ignore>?

<meta> ::= <attn>? <wake> <ignore>?

<attn> ::= "hey" | "ok"
  | "hello" | "hi"

<wake> ::= "huey" | "sigma" | "sigmoid" | "sigmund"
  | "alexander" | "alex"

<connect> ::= "ask" | "tell"
  | ("connect" <to>?)
  | <switch_to>
  | ("go" <to>?)

<switch_to> ::= "switch" ("back")? <to>?

<assistant> ::= <google>
  | "amazon"? "alexa"
  | "apple"? "siri" | "cortana"
```

```
  | <wake> | <item>

<robot> ::= "robot" | "bot" | "android" | <item>

<google> ::= "google" "assistant"?
  | "google" "home"?

<service> ::= "spotify" | "pandora"
  | "amazon"
  | "apple" "itunes"?
  | "youtube" | "yt"
  | "tidal"

<music> ::= "music" | "songs" | "playlist"

<skill_remote> ::= "forecast" | "weather" | "weather" "forecast"
  | "calendar" | "appointments"
  | "schedule" | "alarms"
  | (("daily"  | "flash" | "today" | "todays")? "briefing")
  | ("sing"    | "singing") (<item>)?
```

For illustrating the mechanics of the conversation flow, we've continued to hardcode some values. In this example we've used the names of popular existing assistants, as well as some we've proposed here. In a production application, we can store some of this data in configuration files or a database to avoid hardcoding them in grammar rules.

The set of grammars listed in this section demonstrate that we can chain together short instruction sets to provide sophisticated functionality that is not currently available in commodity smart assistant services.

### *5.10.2. List Management*

Huey can be used to manage a variety of lists using spoken commands or the text equivalent. Here we show some examples including shopping lists, to do lists, and designs for sharing lists from Huey to other systems.

***5.10.2.1. Creating Lists by Voice.*** Using a Huey-enabled web site, a user may create different types of lists. Earlier in this paper we examined the grammars supporting adding items to a shopping list using a command such as

```
add cookies
```

Without naming a particular list, items are added to a default list that is not associated with a store. Figure 18 shows how this works in a web browser powered by the Huey engine. The user can type in an “add” command, or if the microphone is enabled, speak the command instead.

Huey supports basic operations to add, remove, and display list information. The engine also provides a mechanism to add new *handlers* to add functionality based on user actions. For example, a programmer may have written a recommendation option to help a shopper remember to choose items that are related to items already added to the shopping list. The recommendation engine receives the current list info and the most recent additions made by the user, and returns a list of recommendations, as shown in Figure 19.

Besides shopping lists, we can also create a “to-do list”. In Figure 20 we see how this would work on a laptop or tablet screen. The user can use their voice to manage the list or switch to using a real or virtual (on-screen) keyboard.

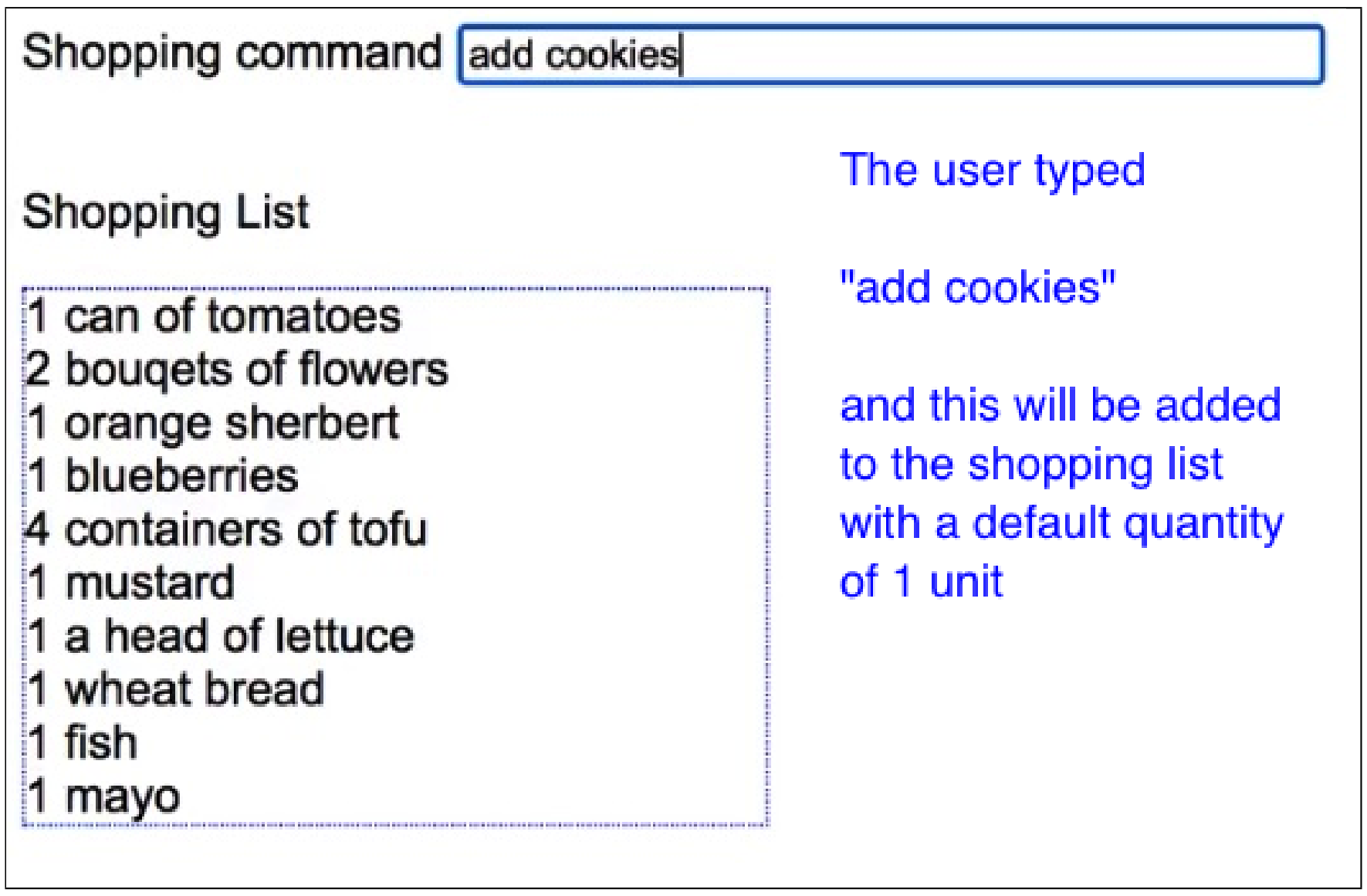

**Figure 18.** Adding an item to a shopping list

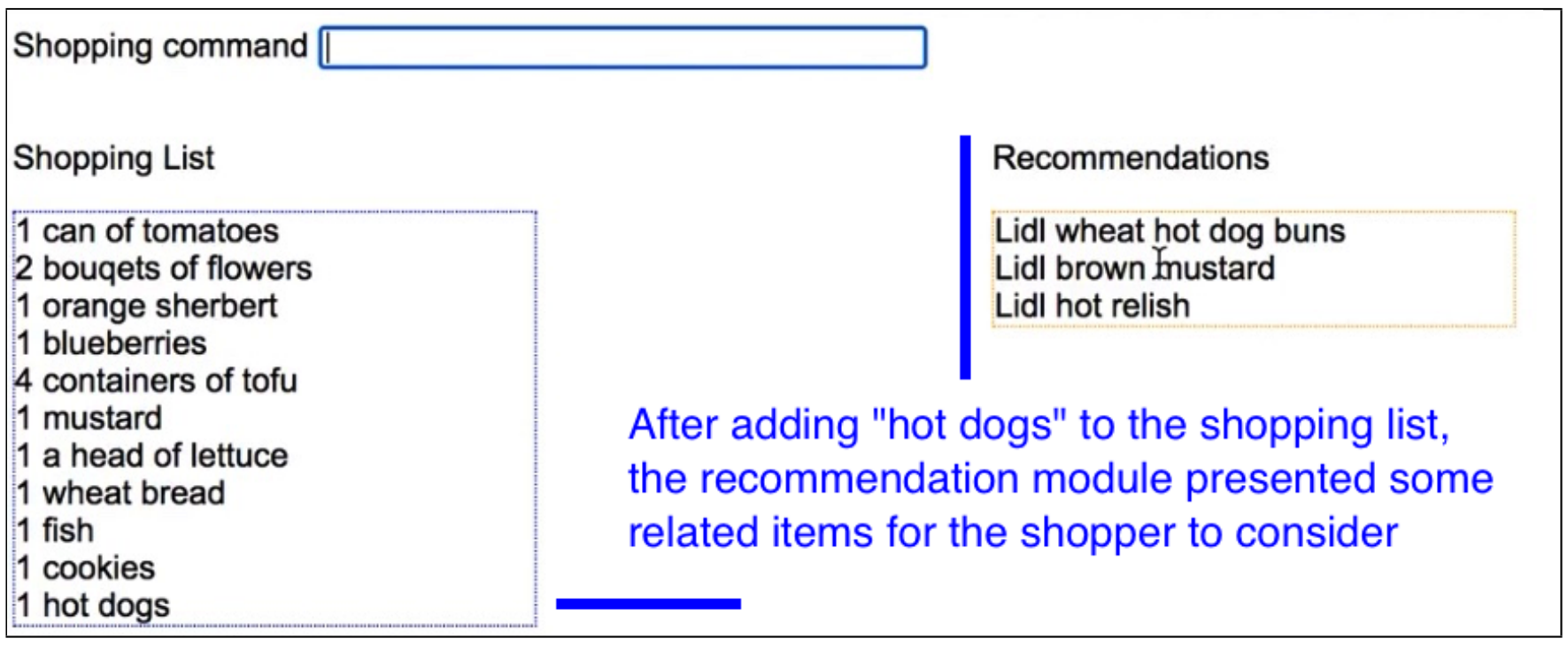

**Figure 19.** Recommendations handler connected to the shopping list

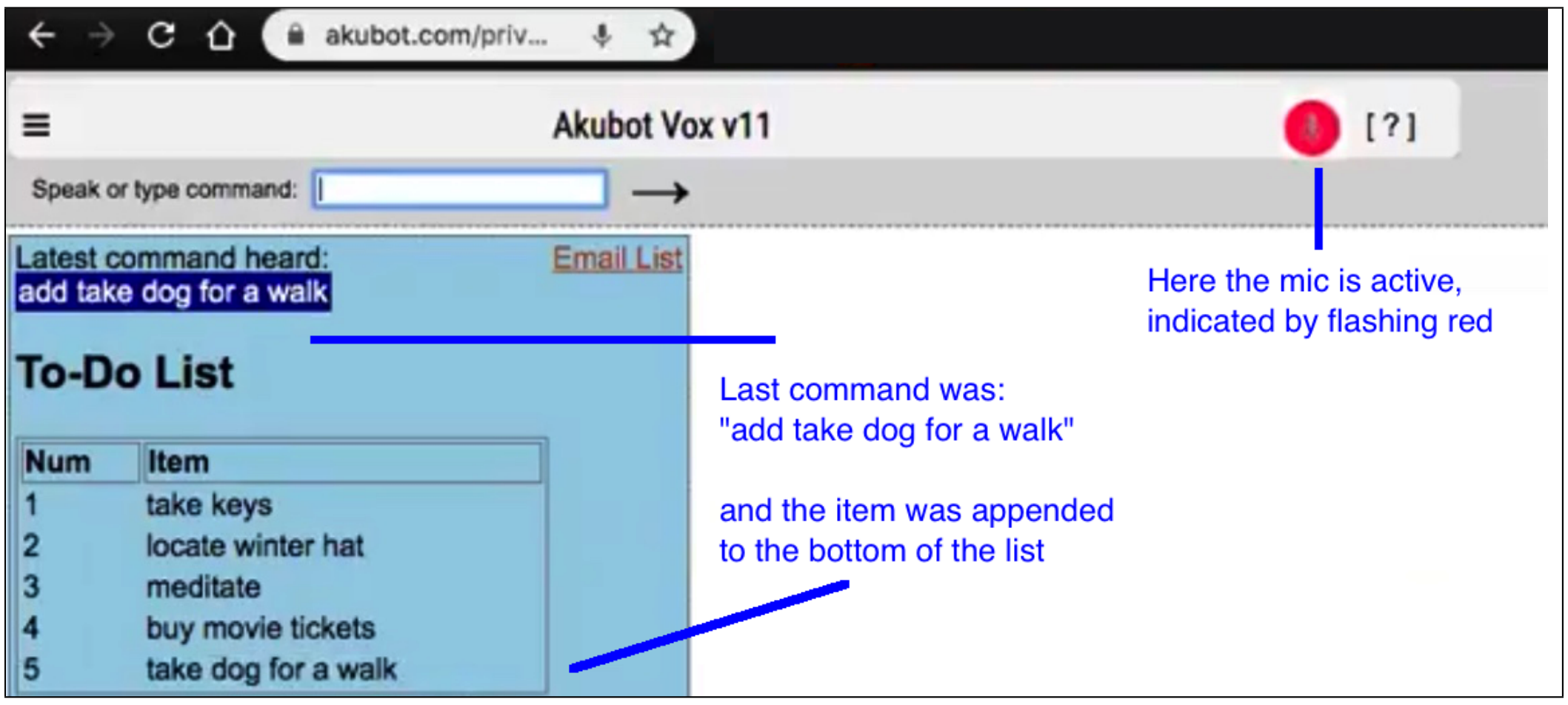

**Figure 20.** To-Do list managed by voice

***5.10.2.2. Creating Named Lists.*** The user can make a new list using a command such as

```
create a Target shopping list
```

Once a named list is created, a user can view items on that list

```
display my Target shopping list
```

assuming they are using a device with a screen.

### *5.10.3. Sharing Lists by Anonymous URL*

Many people are rightfully concerned about data privacy. A user may want to keep their shopping list separate from any particular stores, or even their internet or phone service provider.

Huey allows the user to share a shopping list via an anonymous URL. This command

```
share my shopping list
```

will generate a random hash string that is appended to the end of a URL, e.g.

```
https://www.example.com/lists/share/R7CC6N
```

On the web server this maps to a static text file with the contents of the list. After the user copies this URL (via their clipboard or from an email), they can send it to a store as a one-time link to download the shopping list text. The store's server can use a Linux command such as

```
wget https://www.example.com/lists/share/R7CC6N
```

to get the list.

The Huey server should have a maintenance process to delete the temporary links and the files from its file system, to protect user data.

### *5.10.4. Sharing Lists via Messenger Apps*

The user can send the list to a store by sending an email message or using almost any messaging app.

***5.10.4.1. WhatsApp messenger app.*** One popular messaging app is WhatsApp. Some supermarkets use WhatsApp to communicate with their customers, who use their mobile phone to chat with an employee or a chatbot. Either way, once the user has their shopping list URL, they can paste this into WhatsApp so the store may download the list. Figure 21 shows this scenario.

Alternatively, the store may prefer to have the contents of the list pasted into a messenger session, rather than a URL to get the list. Figure 22 shows this scenario.

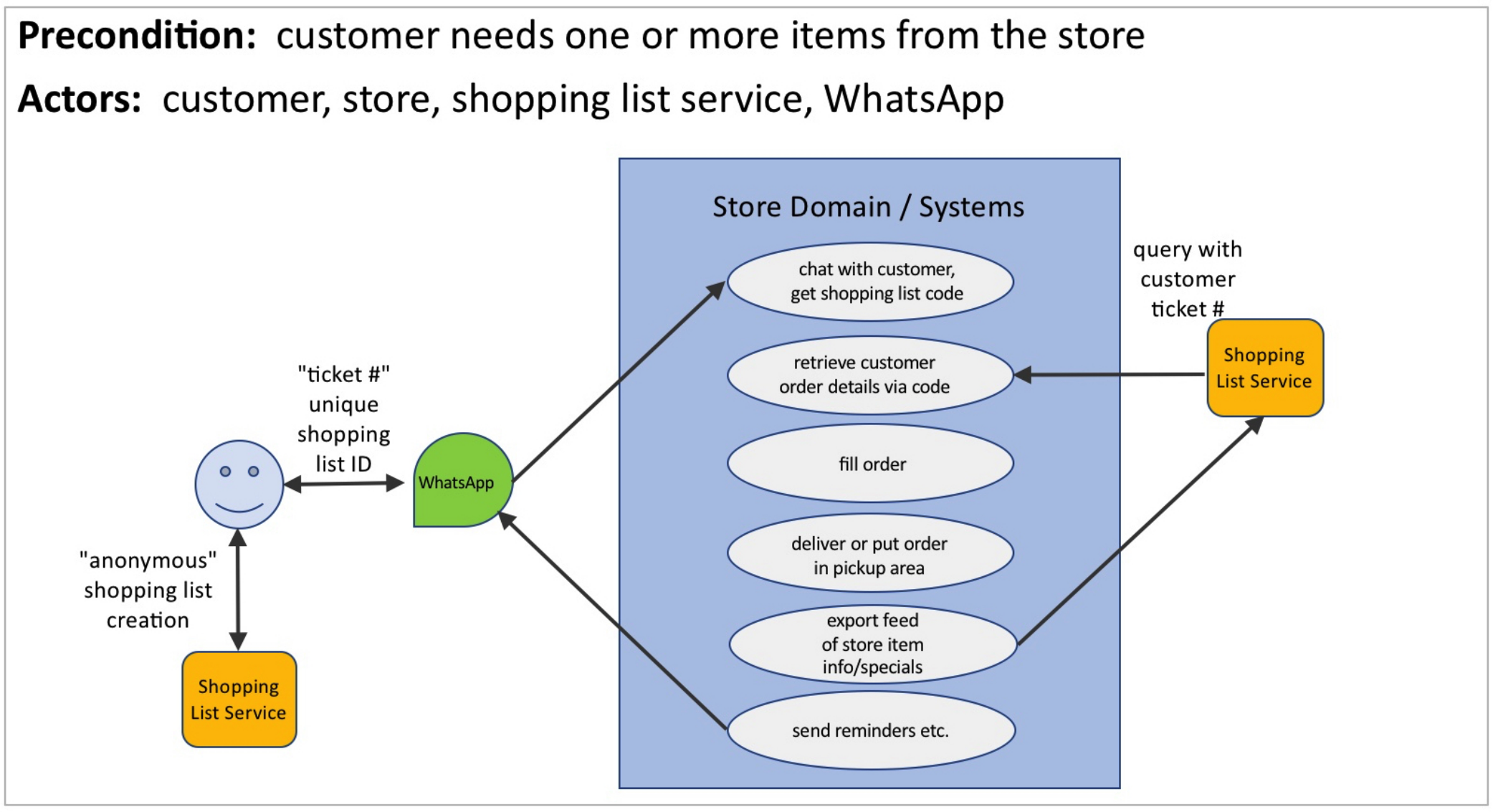

**Figure 21.** Sharing a shopping list URL via WhatsApp

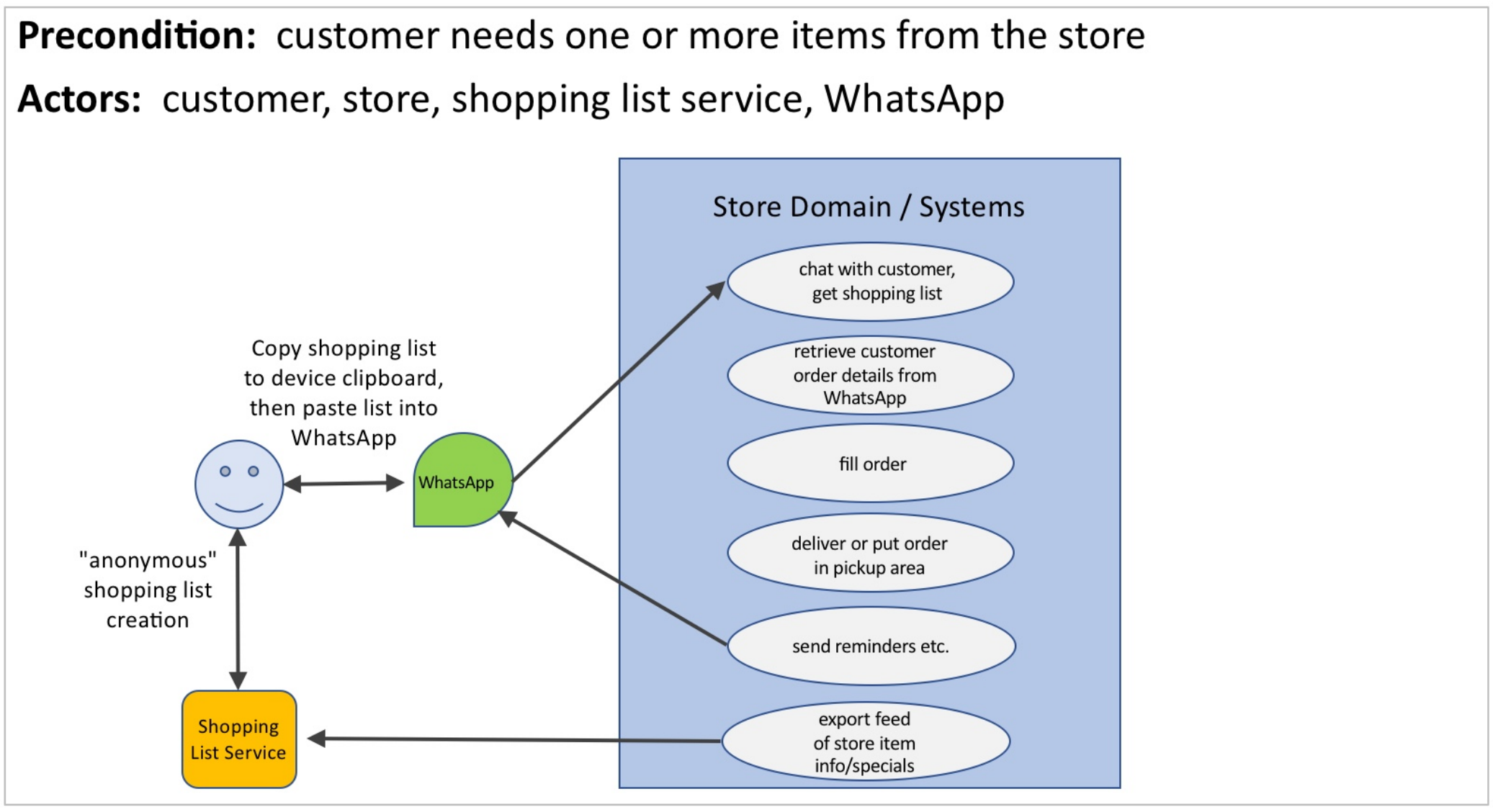

**Figure 22.** Pasting a shopping list into WhatsApp

#### *5.10.5. Navigating a Single-Page Web Application*

In some web applications, only a few data items are needed to fill out a form or a request. The data items may be collected in a linear fashion, the same for any user and each session. Or there might be a different logical flow, with branching logic depending on responses to questions along the way (e.g. in a user survey).

In recent years some developers have created web sites using only one HTML page, known as *single-page web applications* (SPA). The page reloads with new content as the user goes forward (or back) through a sequence of steps. The web page is programmed to display the text and fields by enabling pre-loaded items with JavaScript, or retrieving new instructions from the web server.

Any smart phone will have a web browser, and since many browsers have the ability to convert speech to text, we can enable a web application to use this text. The text can be inserted into a form field, or interpreted as a command to click a button.

The Huey grammars and engine can be leveraged to build this kind of application, as shown in Figure 23

In this example, after the user says "Start" or taps the button, the next page is displayed as in Figure 24.

Note how the constraints of the

- small form factor of the phone, and
- one data item per page

benefit us when voice-enabling an application. The reason is, that on any page the user will see a very limited number of options, and the program will expect a small number of inputs.

Once the user provides all the data, they can be shown a summary page to confirm and be given a chance to go back to fix – using their voice if they like (Figure 25).

After the user selects Finish by voice or by touching the screen, they can be sent back to beginning of the application or to another page on the site (see Figures 26 and 27).

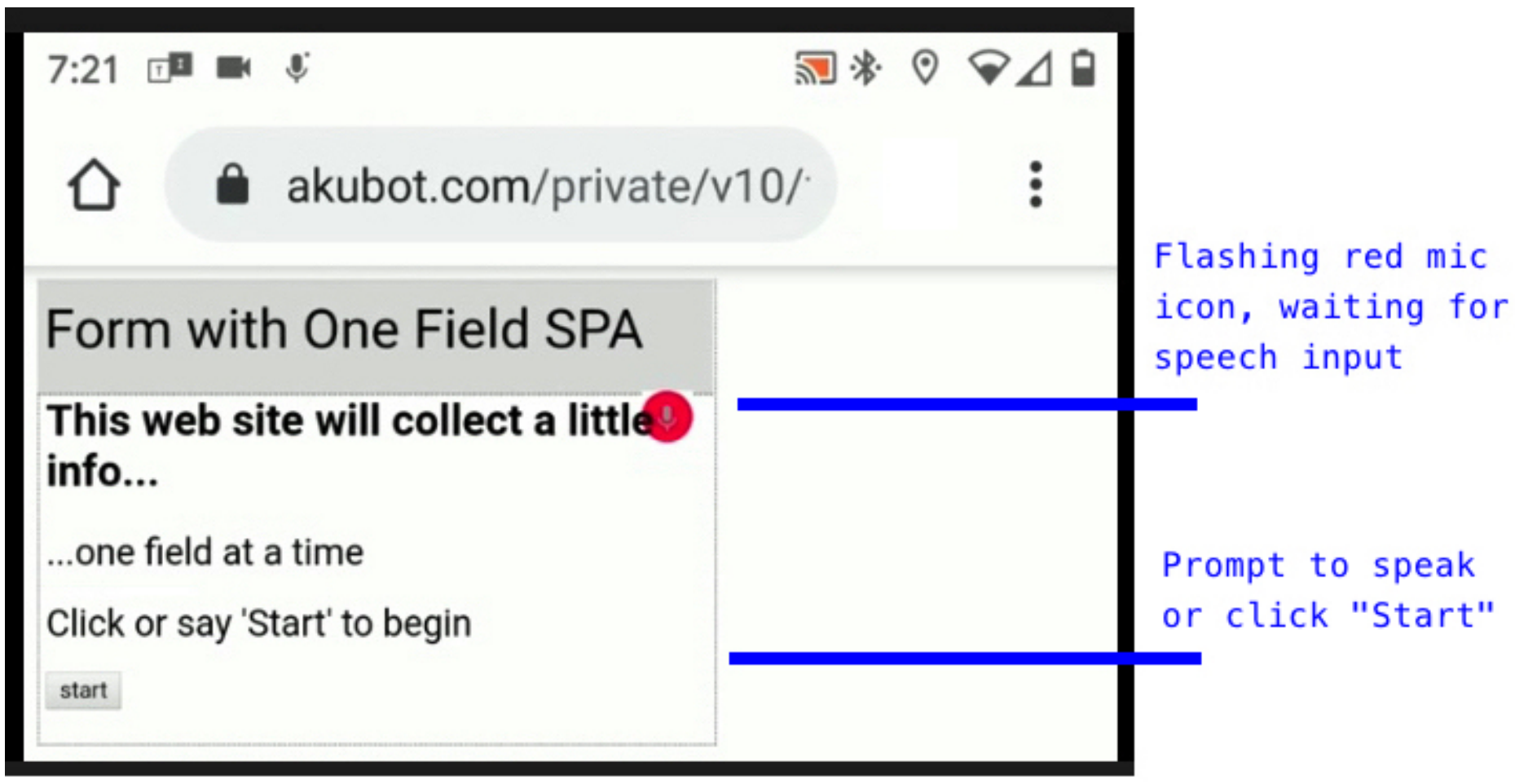

**Figure 23.** Single-page web application on a smart phone with voice control

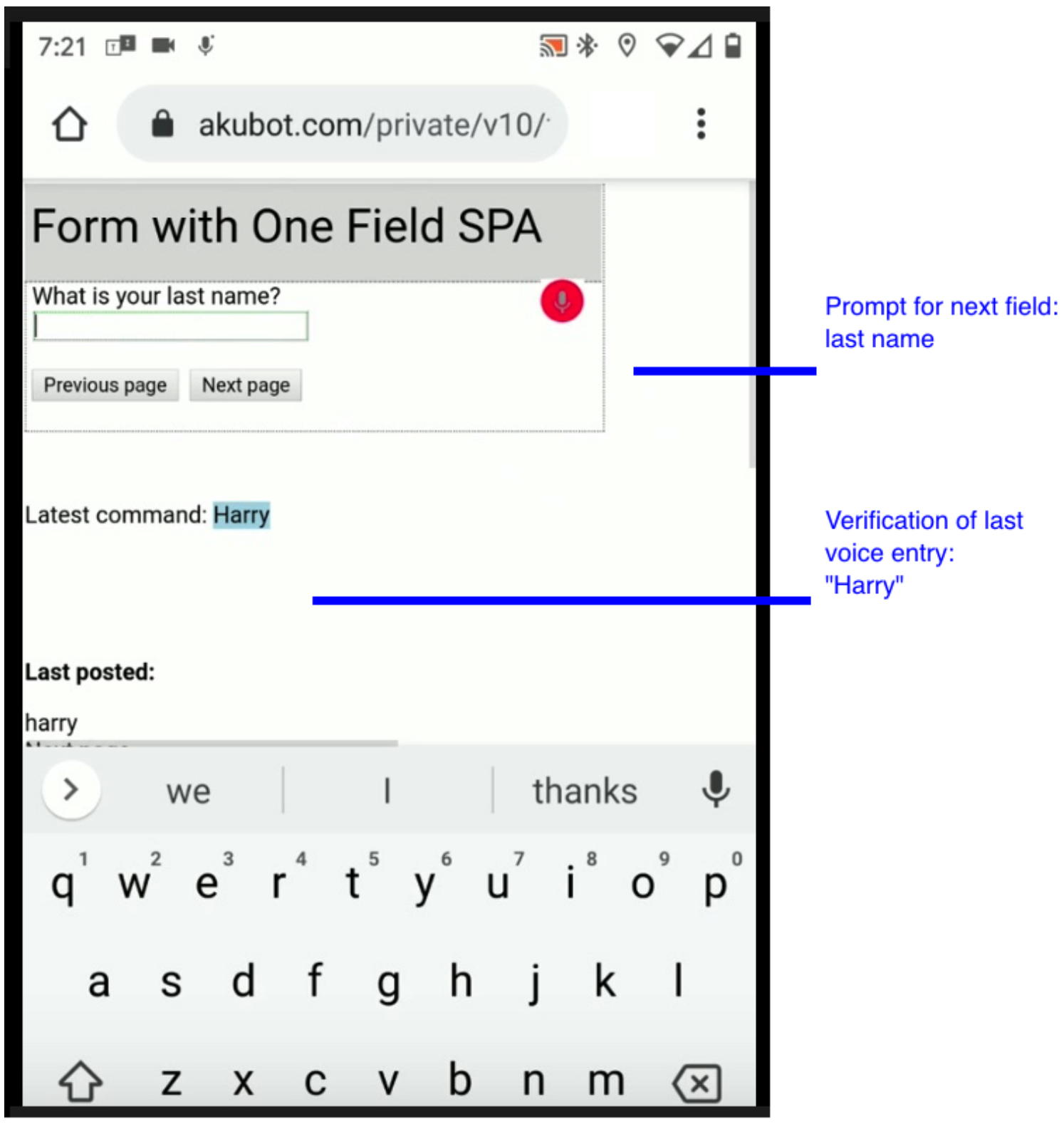

**Figure 24.** Page 2 of the SPA, with one field

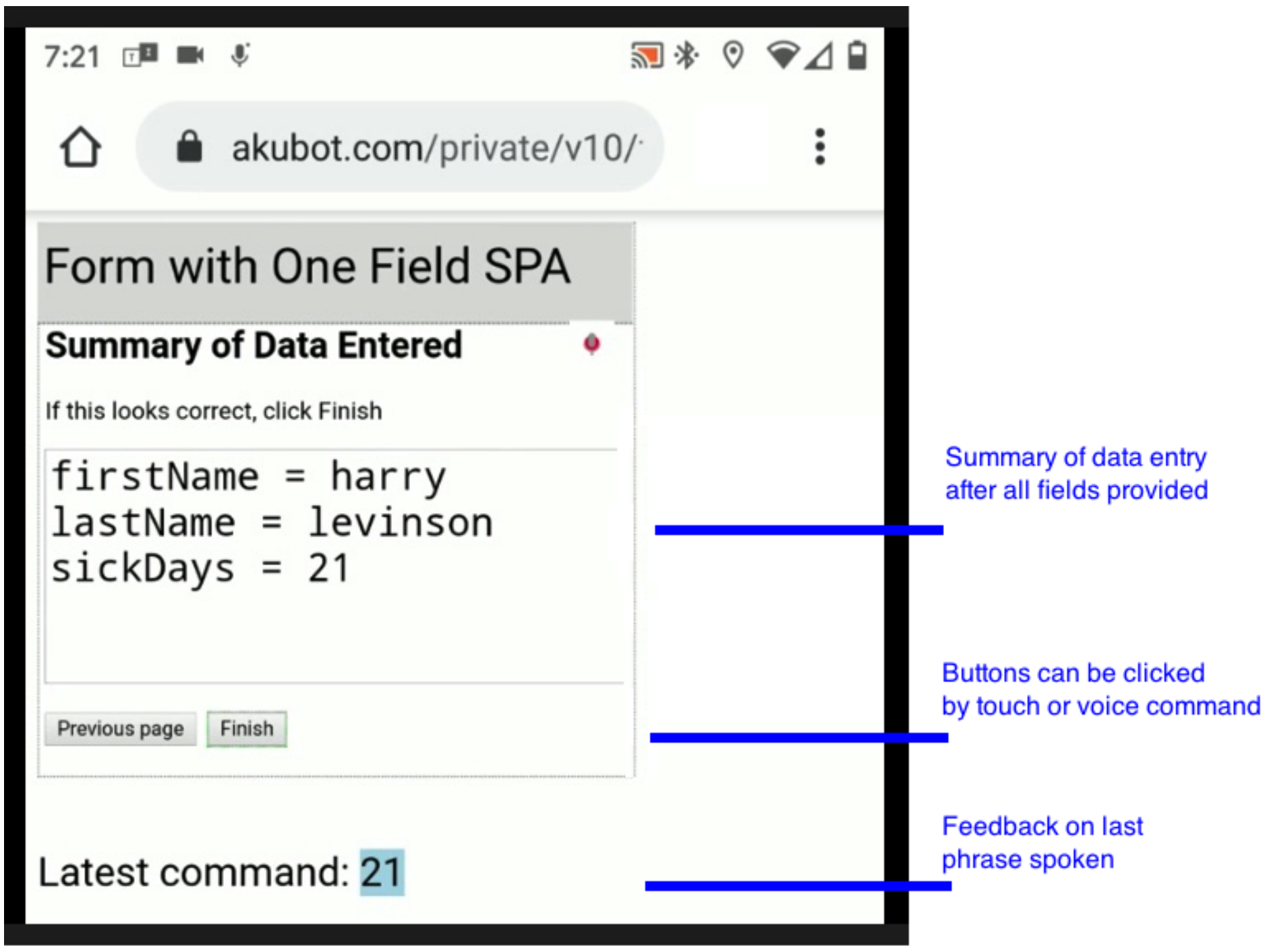

**Figure 25.** Summary page of the SPA

### *5.10.6. Navigating Web Forms by Voice*

On a laptop or desktop computer, we have more screen "real estate" to work with. Here Huey can be used to navigate forms with more than one field.

Consider a simple web page such as the one in Figure 28.

We can connect a web application to the Huey engine, and provide form navigation by voice. The form shown in Figure 29 allows the user to speak commands to add text to the form, as well as move between fields. The right side of the screen contains a unit test rig to exercise the application by running a sequence of commands. Here you can see a mix of data entry and navigation commands.

The same Huey engine can also be used to choose items from a menu, such as on an index page like the one in Figure 30.

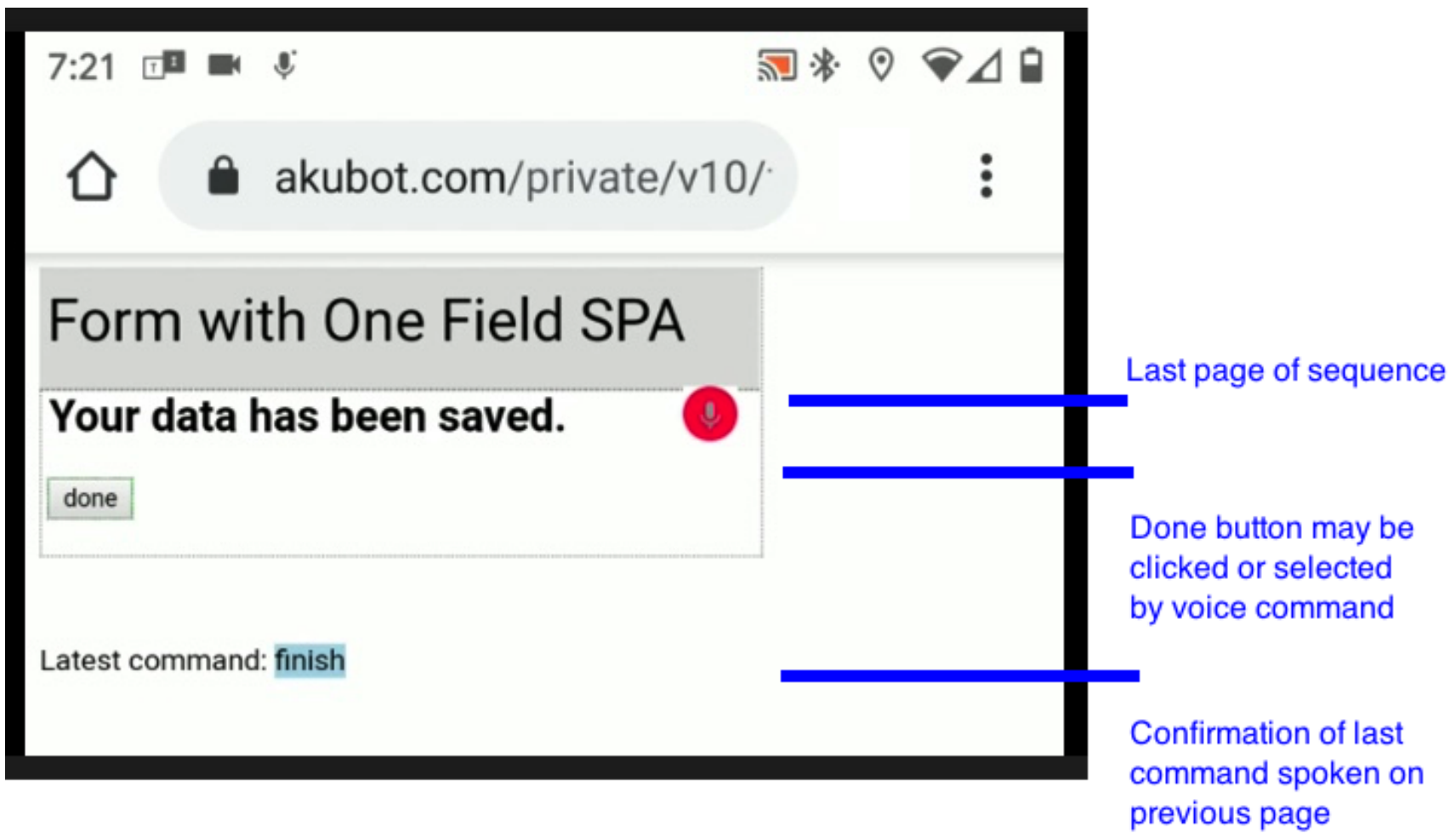

**Figure 26.** Summary page of the SPA

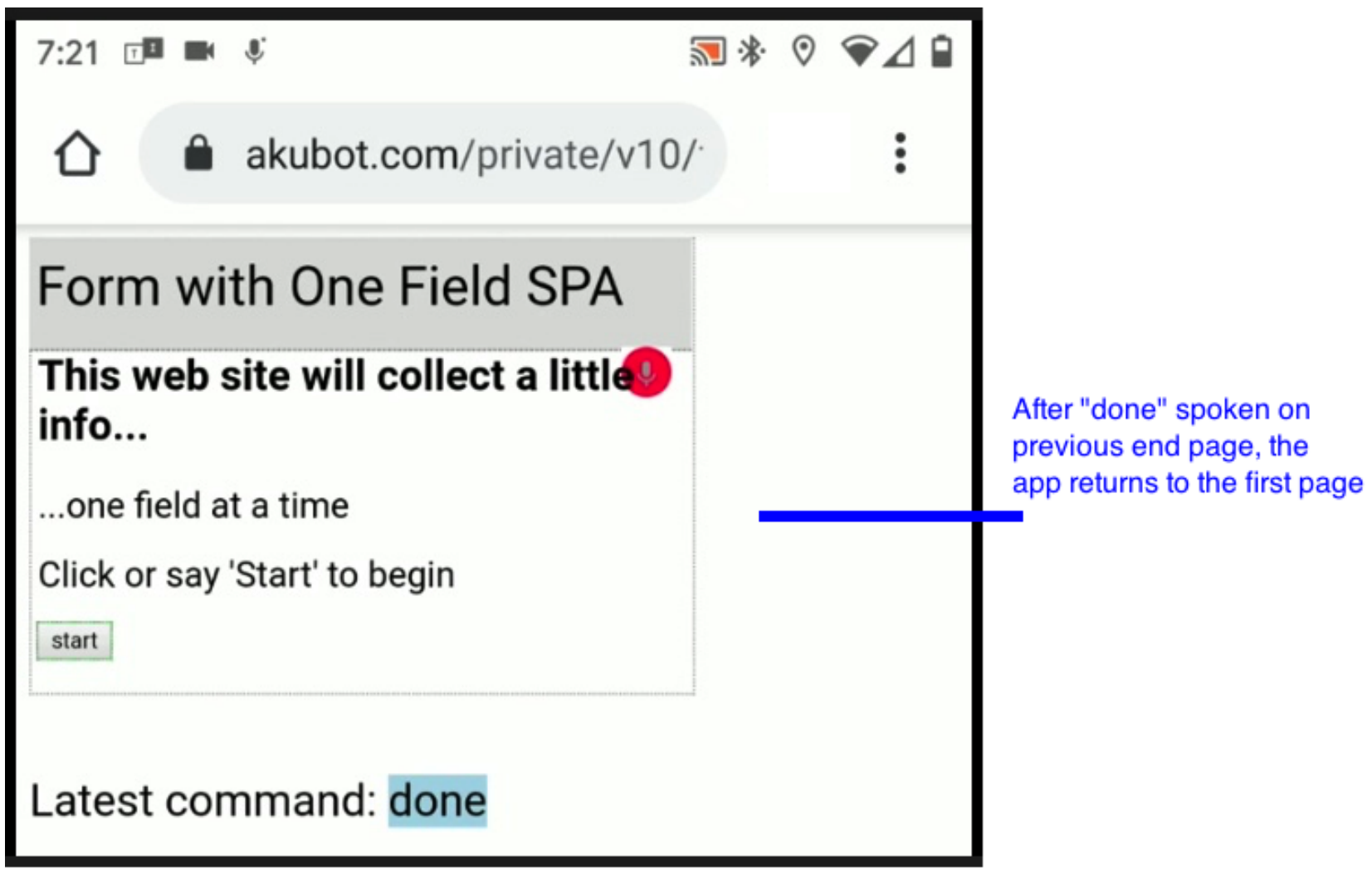

**Figure 27.** Return to first page of the SPA

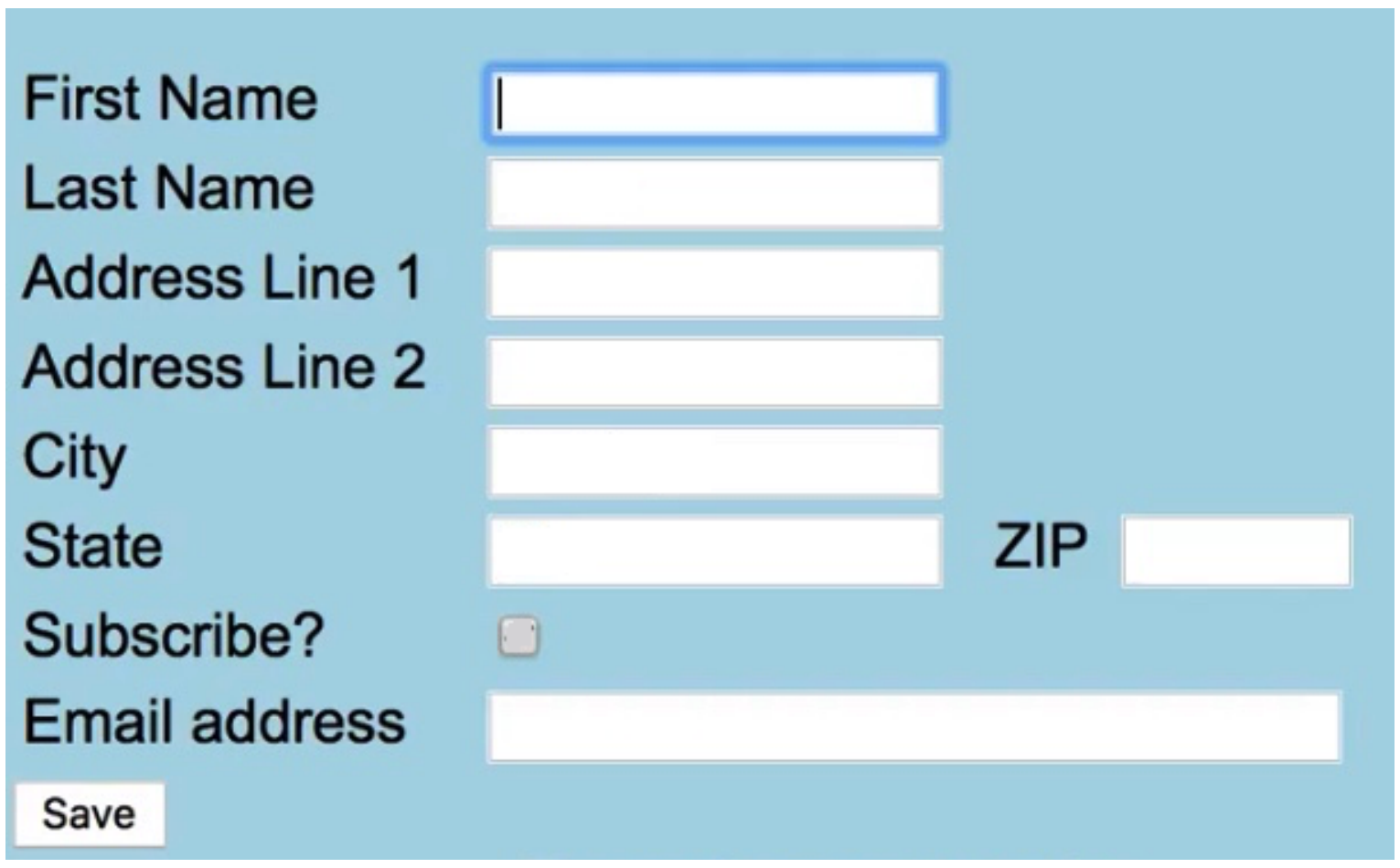

**Figure 28.** A simple web form

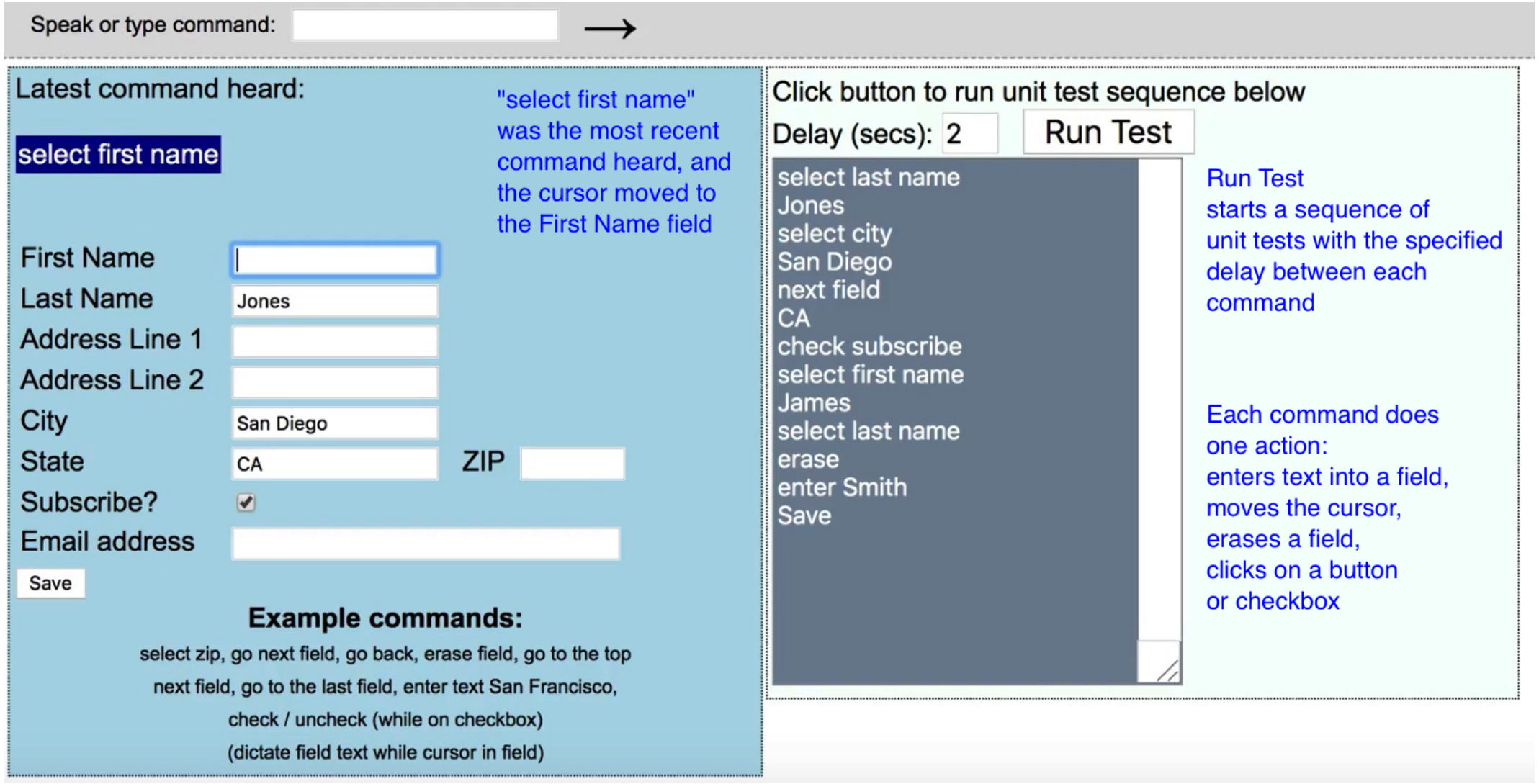

**Figure 29.** A Huey-enabled web form

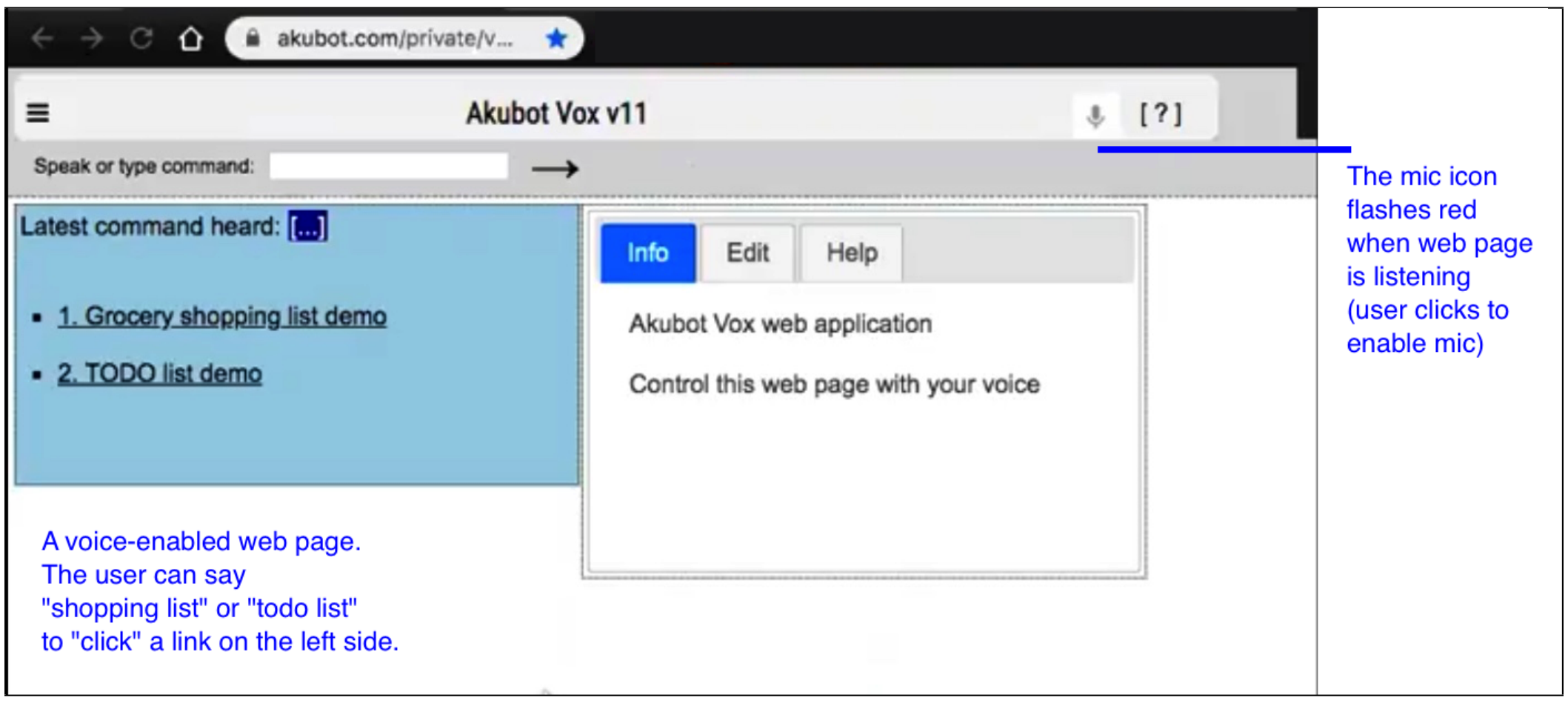

**Figure 30.** A Huey-enabled index page

**Vehicles**

| Num | A | B | C | D | E | F |
|---|---|---|---|---|---|---|
| 1 | Year | Make | Model | Description | Price | |
| 2 | 1997 | Ford | E350 | ac, abs, moon | 3000.00 | |
| 3 | 1999 | Chevy | Venture "Extended Edition" | | 4900.00 | |
| 4 | 1999 | Chevy | Venture "Extended Edition, Very Large" | | 5000.00 | |
| 5 | 1996 | Jeep | Grand Cherokee | MUST SELL! air, moon roof, loaded | 4799.00 | |
| 6 | | | | | | |

**Example commands:**

**Figure 31.** Selecting a column by voice in a spreadsheet

### *5.10.7. Navigating and Enhancing a Spreadsheet by Voice*

Consider a small spreadsheet with sample data from a CSV file [4] on some vehicles for sale displayed in a web page, as in Figure 31.

The figure shows the results of speaking the command

```
select the price column
```

A common scenario with spreadsheets containing numbers, especially financial data, is to add summary rows to the bottom of a data table. To accomplish this, the user may speak the command

```
add a formula to sum it
```

and Huey will infer that "it" means "the price column" referenced by the previous command. The browser displays a new row as shown in Figure 32.

Huey allows us to implement SQL-like commands such as

```
select rows where price is less than 5000
```

to highlight or select rows that match criteria relevant to the data and structure (metadata) for the current document; see Figure 33.

Taking this concept one step further, we can group together Huey navigation commands to create Huey macros. These user-defined macros may be stored in the user's

[4] Data source for CSV: https://web.archive.org/web/20200501232301/https://en.wikipedia.org/wiki/Comma-separated_values

**Vehicles**

| Num | A | B | C | D | E | F |
|---|---|---|---|---|---|---|
| 1 | Year | Make | Model | Description | Price | |
| 2 | 1997 | Ford | E350 | ac, abs, moon | 3000.00 | |
| 3 | 1999 | Chevy | Venture "Extended Edition" | | 4900.00 | |
| 4 | 1999 | Chevy | Venture "Extended Edition, Very Large" | | 5000.00 | |
| 5 | 1996 | Jeep | Grand Cherokee | MUST SELL! air, moon roof, loaded | 4799.00 | |
| 6 | TOTAL | | | | 17699 | |

**Example commands:**

**Figure 32.** Spreadsheet display after "add formula" voice command issued

**Vehicles**

| Num | A | B | C | D | E | F |
|---|---|---|---|---|---|---|
| 1 | Year | Make | Model | Description | Price | |
| 2 | 1997 | Ford | E350 | ac, abs, moon | 3000.00 | |
| 3 | 1999 | Chevy | Venture "Extended Edition" | | 4900.00 | |
| 4 | 1999 | Chevy | Venture "Extended Edition, Very Large" | | 5000.00 | |
| 5 | 1996 | Jeep | Grand Cherokee | MUST SELL! air, moon roof, loaded | 4799.00 | |
| 6 | TOTAL | | | | 17699 | |

**Example commands:**

**Figure 33.** Spreadsheet display after "select rows" voice command issued

cloud account to allow code re-use across documents.

This powerful technique – using voice to navigate and enhance documents – can be extended to other types of documents that may be edited in a browser context.

### *5.10.8. Adding Data to a Spreadsheet by Voice*

Suppose we want to track our travel expenses on a business trip, by using voice commands and a mobile device. It would be convenient to create a new spreadsheet based on an existing template, such as the one in Figure 34.

After creating a new spreadsheet document from the template, the traveler might want to issue a long natural language command such as

```
add an expense for category lodging with a
description of hotel on june first for
two hundred dollars and fifty seven cents
```

If we knew that the traveler would always speak the various phrases in the same order the same way, we could write one or more grammars to handle such input, then parse it to populate a semantic model [17].

Unfortunately, while hoping to translate natural spoken language, we need to be prepared to accept many different ways of saying the same thing (often true in English). Long commands or requests are especially difficult, both for the user to remember all the rules for making a structured request, and for a programmer anticipating all the possible variations of one sentence.

| | A | B | C | D |
|---|---|---|---|---|
| 1 | Category | Description | Date | Amount |
| 2 | airfare | | | |
| 3 | lodging | | | |
| 4 | ground transportation | | | |
| 5 | meals | | | |
| 6 | other | | | |
| 7 | TOTAL | | | |

**Figure 34.** Travel expenses spreadsheet template

Using a parser we could identify many phrases and a few clauses in the sample sentence above. Then we could write a complicated grammar that would handle this sentence and ones that have a very similar structure. But when another sentence type comes along that is also comprised of many words, we would almost certainly need to write an additional grammar to handle that second structure.

This is clearly a losing battle, because “add an expense” is only one of several kinds of statements we’d like this voice application to handle. As long as a human is writing the code to go along with the grammars, it’s much work to write and maintain this quantity of code.

Another way for the user to accomplish the same thing – add a new expense item – would be to *progressively capture* the different data values in the item. Here is our strategy for writing some reasonable grammars:

- Capture the user intent to add an expense
- Add a spreadsheet row and highlight it in the UI
- Let the user continue the conversation and provide more data values
- Update the UI to show new cell values
- Listen for other commands

Here is a sample conversation between the user and the assistant, which breaks up the long request into several smaller ones. Now we have several manageable, natural language phrases. We add onto this a sudden jump to another intent and finally return to add another expense.

```
hi sigma
please create a new spreadsheet using the
  travel expenses template
add an expense for lodging
now set the description to hotel
the date is june first
the amount was two hundred dollars
and fifty seven cents

hey sigma, ask google for my forecast

add another expense for breakfast
```

Figure 35 shows the spreadsheet after each sub-request described above (not shown

| | A | B | C | D |
|---|---|---|---|---|
| 1 | Category | Description | Date | Amount |
| 2 | airfare | | | |
| 3 | lodging | | | |
| 4 | lodging | | | |
| 5 | ground transportation | | | |
| 6 | meals | | | |
| 7 | other | | | |
| 8 | TOTAL | | | |

| | A | B | C | D |
|---|---|---|---|---|
| 1 | Category | Description | Date | Amount |
| 2 | airfare | | | |
| 3 | lodging | | | |
| 4 | lodging | hotel | | |
| 5 | ground transportation | | | |
| 6 | meals | | | |
| 7 | other | | | |
| 8 | TOTAL | | | |

| | A | B | C | D |
|---|---|---|---|---|
| 1 | Category | Description | Date | Amount |
| 2 | airfare | | | |
| 3 | lodging | | | |
| 4 | lodging | hotel | 6/1/2020 | |
| 5 | ground transportation | | | |
| 6 | meals | | | |
| 7 | other | | | |
| 8 | TOTAL | | | |

| | A | B | C | D |
|---|---|---|---|---|
| 1 | Category | Description | Date | Amount |
| 2 | airfare | | | |
| 3 | lodging | | | |
| 4 | lodging | hotel | 6/1/2020 | 200.00 |
| 5 | ground transportation | | | |
| 6 | meals | | | |
| 7 | other | | | |
| 8 | TOTAL | | | 200.00 |

**Figure 35.** Spreadsheet data added via several short commands

are the diagrams for the forecast and breakfast commands).

We do have a few long phrases, but using grammar and heuristics we can handle them. Nevertheless we pulled out part of the original content into several more manageable pieces. The listing below shows a partial BNF grammar that will handle the sample conversation in parts:

```
<input>       ::= <wake> | <meta> | <sheet_new>
                  | <add_row> | <set_cell>

<wake>        ::= <attn> <wake> | <wake>

<meta>        ::= <sheet_new> | <meta_other>

<sheet_new>   ::= <ignore>* <create> <det>? <sheet>
                  (<with>+ <ID> <template>)

<sheet>       ::= "spreadsheet" | "spread sheet" | "sheet"

<meta_other>  ::= <wake>? <connect>? <assistant>
                  <for>? <det>? <skill_remote>

<add_row>      ::= <ignore>? "add" <det>? <row_entity>
                    <for>? (<category> | <thing>)

<set_cell>     ::= <ignore>? "set"? <det>?
                   <col_entity> <to_be>? <cell_value>

<cell_value>   ::= <phrase_str> | <phrase_num> | <phrase_date>

<skill_remote> ::= "forecast" | "weather" | "calendar"
  | "appointments" | "schedule" | "alarms"
  | (("daily" | "flash")? "briefing")

<row_entity> ::= "expense" | "row" | "entry" | "item"

<category>   ::= "airfare" | "lodging" | "meals"
  | "other" | "miscellaneous" |
    ("ground" "transportation"?)

<col_entity> ::= "category" | "description"
  | "date" | "amount"

<thing>    ::= "breakfast" | "dinner" | "tip" | "tips"
  | "lunch" | "ride" | <phrase_str>

<det>      ::= "a" | "this" | "the" | "my" | "new"
  | "another" | "that" | "this"

<ignore>   ::= "please" | "can you"
```

```
  | "would you" | "now" | "um" | "uh"

<with>     ::= ("with" | "using") "the"?
<template> ::= "template"

<for>   ::= "for"  | "of"
<from>  ::= "from" | "with"
<to>    ::= "to"   | "through" | "thru" | "-"

<to_be> ::= "to" | "to be" | "to become"
  | "set to" | "is" | "was"



<connect> ::= "ask" | "tell" |
  ("connect" <to>?) | ("go" <to>?)

<assistant>::= ("google" "assistant"?)
  | "Amazon"? "Alexa" | "Apple"? "Siri" | "Cortana"
```

Key parts of the grammar are shown below in Figure 36. You can see in the diagram and in the grammar text, we hardcoded some strings that normally should be handled more flexibly by an application. This could be the approach if we created one grammar per spreadsheet template. But unlike attempting to parse variants of the original long command, we now can handle a variety of requests that are segmented as described.

Contrast this approach with the typical one-shot functionality described earlier, which is how most commercial digital assistants work. A user wakes up the assistant, makes one command, and gets one answer. In our application dialog, once the user begins the conversation about the spreadsheet (and a particular new row), they can continue using their voice to speak to the assistant. The assistant keeps track of the skill context, while listening for commands that might require it to shift to another skill. Note we also defined some words to safely “ignore”. Our divide-and-conquer strategy makes it easier to insert `<ignore>` terms because we can deal with those per phrase.

Finally, we’ve omitted details regarding parsing of date descriptions and dollar amounts. Regardless of our overall grammar strategy, we need sub-grammars or other code to properly lex and parse such expressions. However our application architecture provides the means to share grammar snippets between handlers, so we can externalize common code from the expense use case code. Our grammar strategy still requires additional logic in our application regarding timeouts, recovering from errors, and similar housekeeping code.

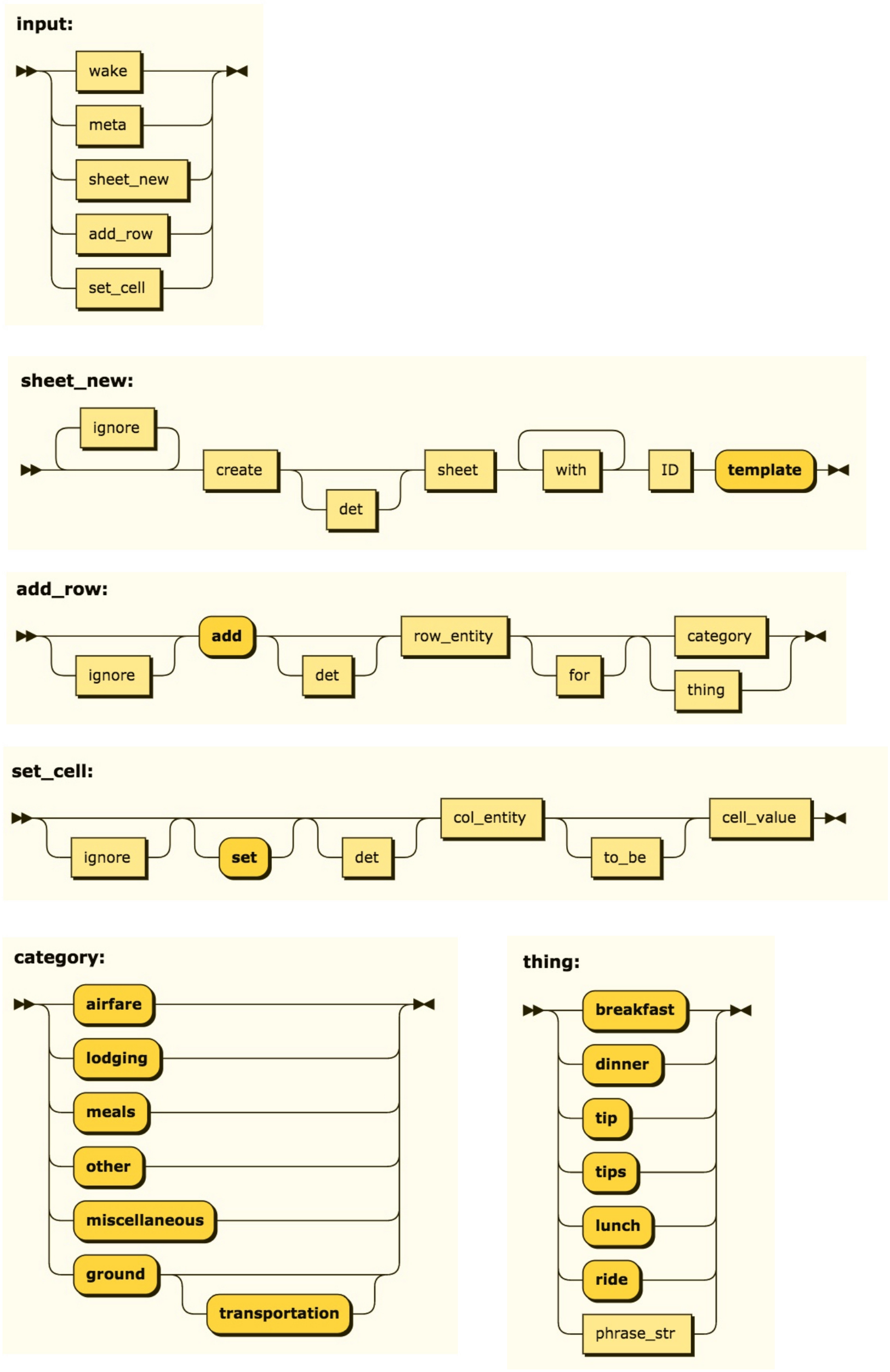

**Figure 36.** Spreadsheet command syntax summary

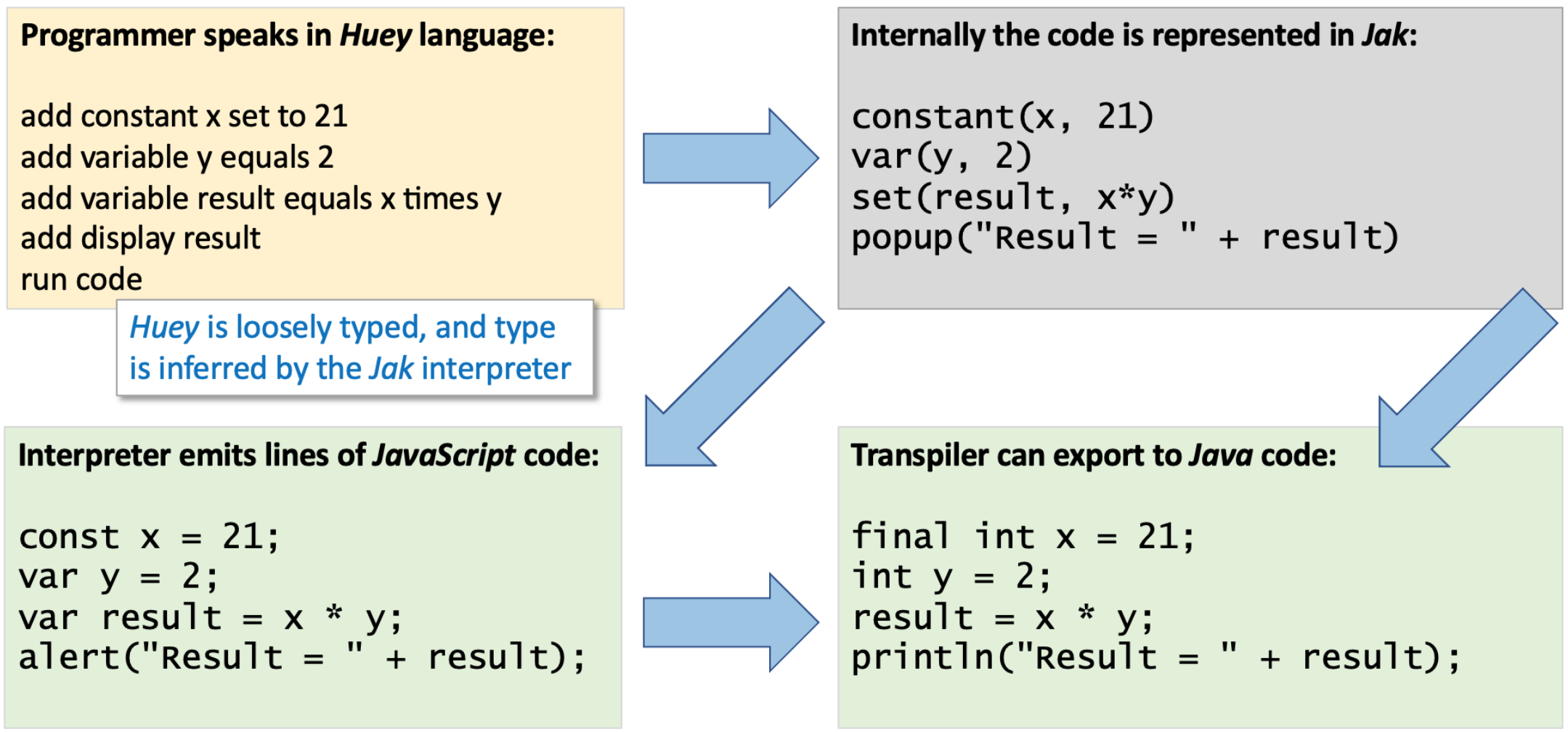

**Figure 37.** Coding by voice with Huey and Jak

### *5.10.9. Coding by Voice*

Huey can assist a programmer by providing voice tools in a web browser.

Our research goals center around creating a system design to provide software developers the ability to create programs and applications using their voice and natural language. We attempt to bridge the divide between spoken conversation between humans, and speech between a human and a computer. Humans are very good at understanding each other, and computers can understand a limited amount of speech from humans [21].

*Conversational computer programming* (CCP) is a new approach that allows software developers to talk to their computer while creating source code for applications [51]. Speech recognition services are now available on laptops and desktops via web browsers, opening possibilities for voice commands to supplement the use of keyboards and pointing devices such as mice. A developer can access a voice-enabled web application that is backed by cloud microservices to provide AI functionality to interpret spoken or written editor commands (see Architecture section).

In a software development context, in contrast to human conversational speech, the vocabulary of programming languages is far more limited. Likewise the range of topics that a programmer might want to speak with the computer about while writing code (opening or saving documents, compiling, etc.) are far more limited. From this perspective, making the computer understand a programmer's speech might seem like a simpler problem to solve than programming general-purpose digital assistants. Yet there don't appear to be successful, widespread development environments for using voice for dictating code in a structured format, or for controlling that environment [15].

Despite many advances in tools for software engineers, programmers today still typically engage in much physical labor to create an application. Whether sitting or standing, they usually have their hands on a keyboard or mouse (or other pointing device). Most programmers will need to switch back and forth between the keyboard and mouse hundreds or thousands of times in a given workday. There is significant risk of a high level of tedium, and for the less fortunate, repetitive stress injuries from overuse of the hands in limited range of motion.

This situation has been entrenched since widespread usage of desktop computers with mice in the 1980s. Most code is still written "from scratch," or put together using snippets from existing applications or examples seen on the internet (e.g., StackOverflow.com). Although countless programming languages, paradigms, and tools have been created, in many ways programming today is very similar to writing code in a text editor during the mainframe and minicomputer eras. Programmers will benefit from newer, smarter tools that at the very least can ease rapid prototyping ([26]; [8], [9]).

Nevertheless, today's laptops are more powerful than even supercomputers in earlier days of computing. While a programmer is thinking, typing, or clicking, their multi-CPU computer with gigabytes of memory is mostly idle – waiting for a keyboard or mouse interrupt in the milliseconds between user actions. There are "wasted cycles" in the sense that these powerful computers can be doing more to help the programmer be more productive, and thereby be less distracted by the mechanics of putting words into source code documents. There is an opportunity to leverage commodity speech recognition tools to improve the user experience for programmers. Instead of manually typing every character of every word, a programmer can translate their higher-order ideas into code more quickly by speaking commands to the computer. In this way, some of the mechanical labor is removed, allowing the developer to generate sections of code with less interruption in their thought processes.

To enable this style of interaction, we propose designs for control languages for the programmers to request the computer create lines or blocks of code using voice commands. We also describe how code created in this environment can be converted to other popular languages. We present an architecture including new languages and tools to construct a speech-controlled IDE.

In Figure 37 a programmer speaks Huey commands such as

```
add constant x set to 21
```

Huey interprets each command into a line of Jak code, as shown in the upper right of the figure – in this case

```
constant(x, 21)
```

Since we're operating in a web browser environment, we do another translation step from Jak to JavaScript, producing

```
const x = 21;
```

Continuing in a likewise manner, a programmer can quickly dictate prototypes for programs that are stored internally as Jak, but displayed and executed in the browser as JavaScript.

Once we have a program written in Jak and stored in the browser or in the cloud, we can also *transpile* (also known as *transcompile*) code into other languages such as Java, Python, or Kotlin.

### 5.11. *Implementing Huey*

Huey's specifications will be elaborated in future publications to enable developers to create their own libraries and systems for the applications discussed in this paper.

Developers may use the popular *Model-View-Controller (MVC)* pattern to create systems, as depicted in Figure 38.

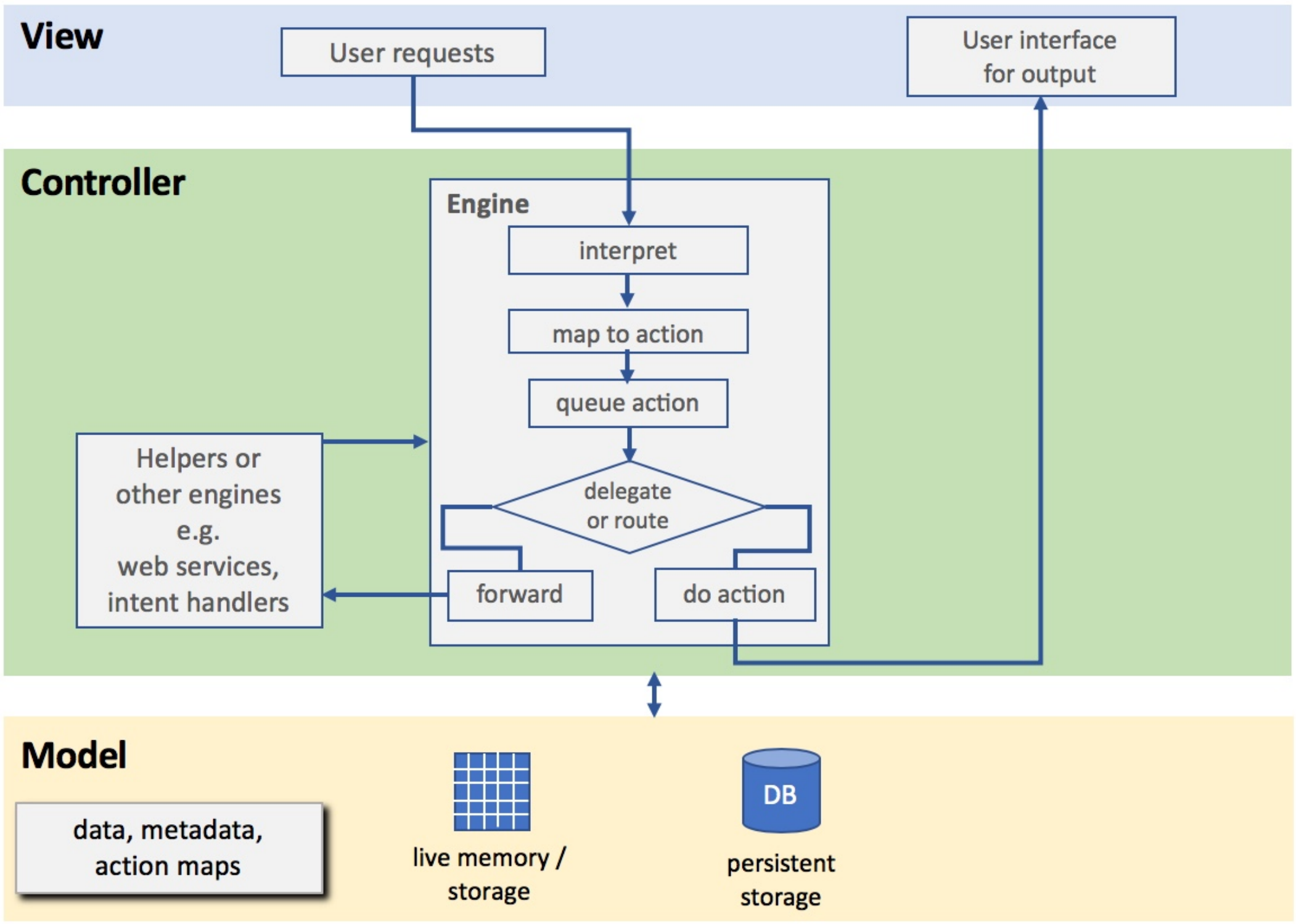

**Figure 38.** Model-View-Controller (MVC) pattern usage

### 5.12. *Huey Language Design Summary*

The design philosophy and approach described above represents a flexible way to anticipate a wide variety of user inputs and convert them to rational system requests. The Huey and Jak languages provide the tools for designing and implementing components that will interact with the VNS to provide distributed processing and communication between humans and machines, as well as machine to machine.

### 5.13. *Introducing Jak*

Jak is a new programming language designed to represent instructions and data in a source code file. It can be used to translate Huey statements into an intermediate form that is convenient for compact storage, that is also human-readable. This code can be executed by a Huey system to carry out system actions.

The name Jak is an acronym for *Jak Ain't Kotlin*, a naming style inspired by GNU, *GNU's Not Unix!*.

In addition to being a convenient form for representing programs, a Jak program can be converted to code using other popular programming languages. This will allow programmers to connect different applications together, whether originally created only with Huey and Jak, or Huey plus other systems.

As defined here, Jak could be implemented as a standalone language. The grammars are sufficiently described to create the standard language support: linters, compilers, interpreters, and packagers. However in its current implementation, it requires a host system to run code. The host system can be built in a JVM language such as Java,

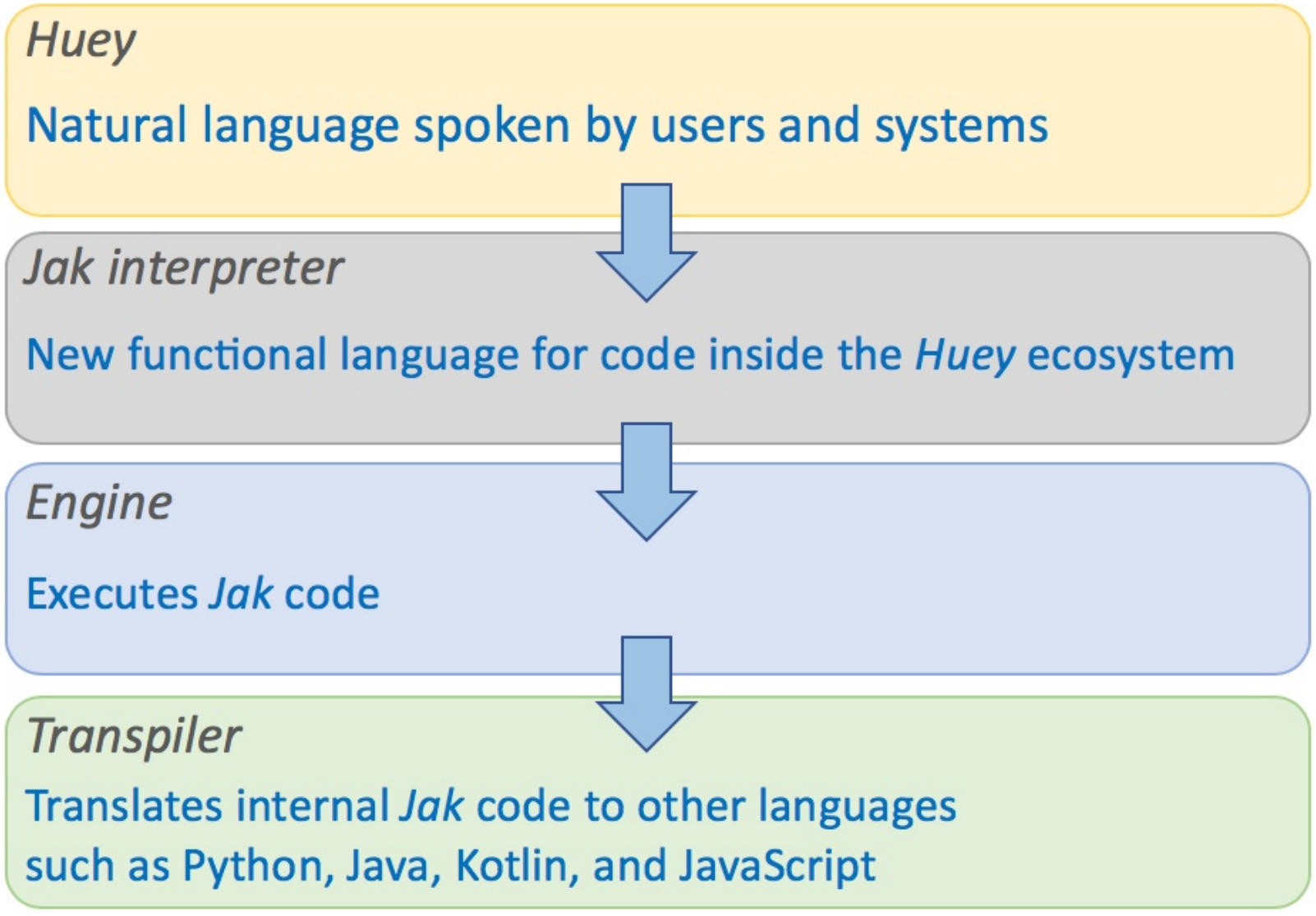

**Figure 39.** Layered domain-specific language (DSL) model

Kotlin, or Scala, or other popular languages such as Python or JavaScript.

### *5.13.1. Domain Specific Languages*

A key element of the Huey architecture is the usage of *domain specific languages* (DSLs) [58].

A DSL is usually smaller in scope than a typical programming language, and much smaller than a predominant human language. They may exist entirely inside another system as an *internal DSL*, or be usable across different kinds of systems – an *external DSL* [17].

The Jak language was initially designed as internal DSL for Huey to represent code and data in a compact form. Also, it shares some syntactic forms used in Java, Python, and JavaScript, each used very widely for about two decades. This makes it easier to read and understand, and provides an opportunity to translate Jak code to these other languages more easily. (See Figure 39.)

Beyond the passing resemblance to the languages mentioned, Jak was specifically designed to share some features from functional Java and Kotlin.

The Jak language was designed such that each line of code (beyond the usual headers) is a function call. This allows us to generate and execute code using a consistent format, while avoiding too much dependence on side effects. Higher-order functions enable us to pass functions as first-class objects as arguments.

Notable features of Jak include:

- Data types of constants & variables are inferred during interpreting or compiling
- Many functions are mostly pure: we avoid side effects
- Higher order functions (such as loop above) enable other functions (such as `bodyFn` above) to be used as first-class objects
- Closures { } are used for defining bodies and lambda functions

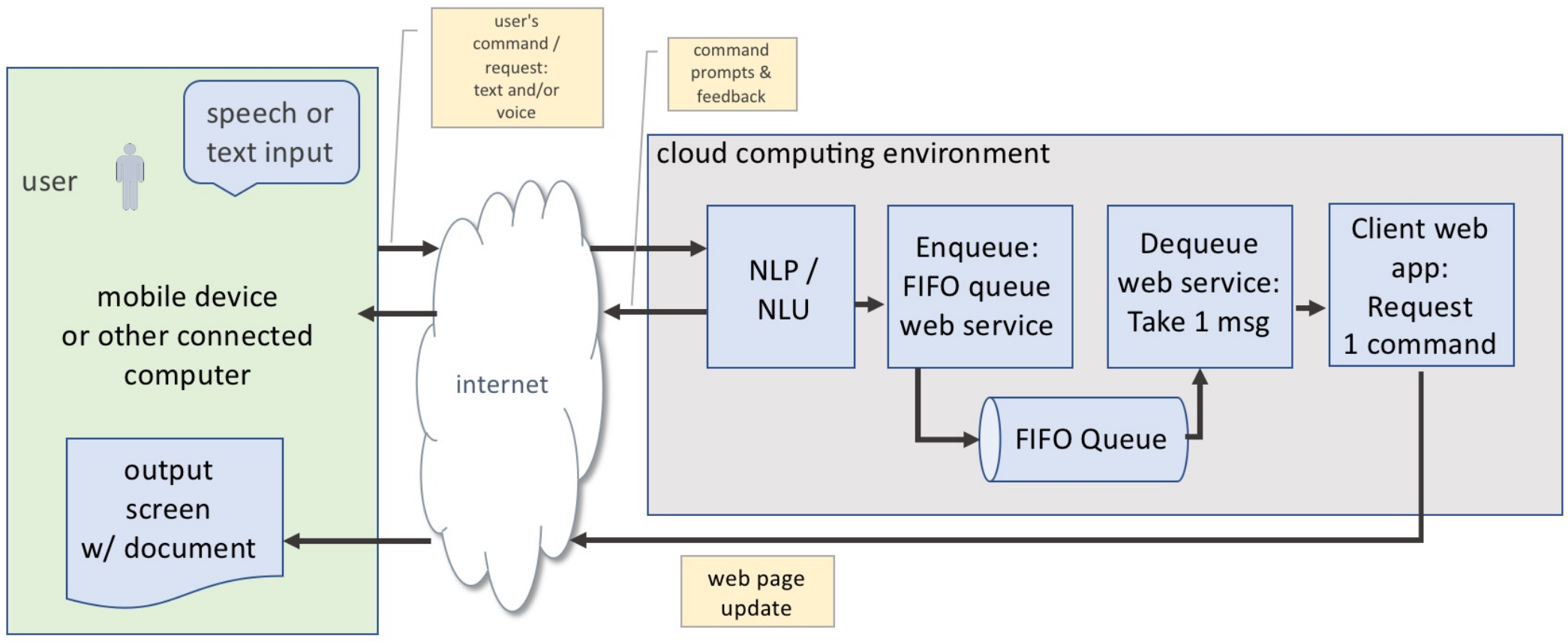

**Figure 40.** Functional architecture for a speech-controlled web application

### *5.13.2. Jak Language Specification*

The Jak language design uses ideas and grammatical forms from JVM languages such Java, Kotlin, and Scala, as well as other languages including Python and JavaScript. Some grammar statements included in this specification were borrowed from the open source GitHub repository with ANTLR grammars for various languages [2] as well as *The Definitive ANTLR 4 Reference* by Terence Parr [39].

For more details on the Jak language, see the language specification in [27].

## 5.14. *Huey Functional Architecture*

The system design consists primarily of software components used as building blocks. These may be implemented as managed services in a cloud environment (to serve users via the internet), or for smaller numbers of users, deployed to a laptop or on-premise data center (see architecture Figure 40).

Cloud computing companies sell managed services to provide functionality such as queues, databases, software-defined networks, and NLP. Nevertheless, there are many open-source software projects that can be used instead. This design configuration would provide the ability to use cloud environments to run a set of containerized applications that can be configured to be independent of the cloud providers vendor-specific solutions [37].

In Figure 40, many details of the physical implementation, such as clustered servers and networking, are hidden to focus on the application building blocks. This system will let a user issue verbal commands or a type commands into a chat window to perform actions as described in the use cases documented in this paper.

For a more detailed description of the Huey functional architecture, see [27]. The following sections provide some highlights.

### *5.14.1. Huey Speech Interface*

The Huey Speech Interface is a web application with speech recognition provided by the web browser. Most commonly used browsers support a speech API that enables any computer (laptop, smartphone, tablet) with an internet connection to speak instead

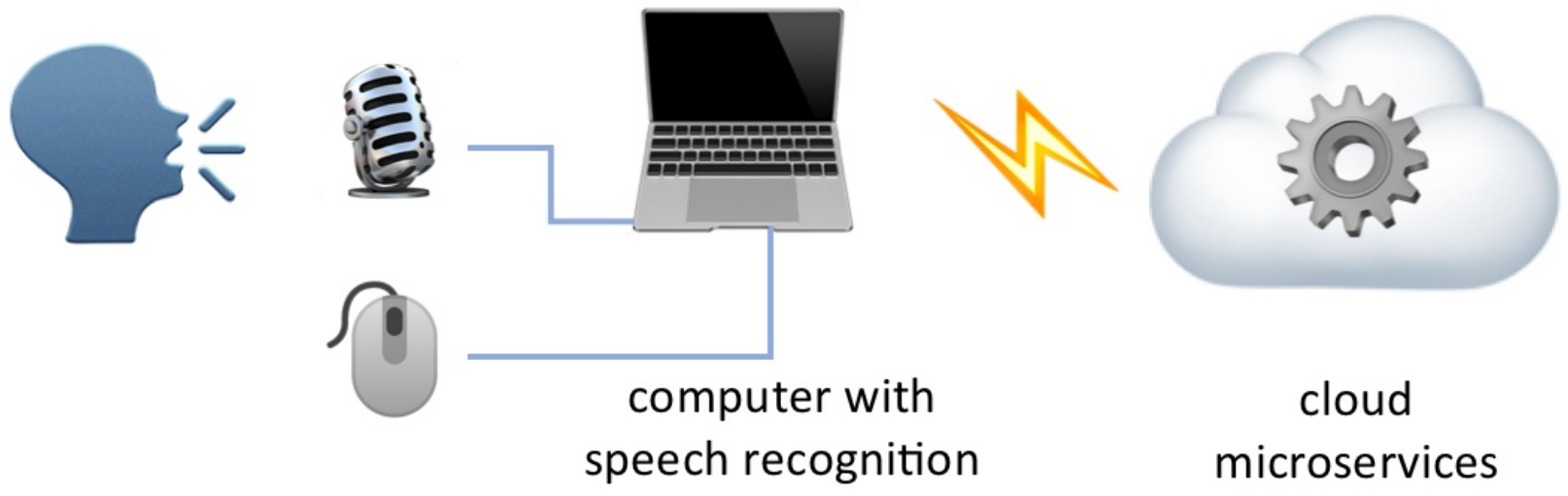

**Figure 41.** Cloud AI access via laptop

of type.

Once the user grants permission for the browser to access the computer's microphone, speech may be converted to text. The text seen by the proposed application will be an expression of natural language. While unstructured "conversations" between the user and the computer can generate text on the screen quickly, this is not particularly useful beyond dictation for regular computer documents or form fields.

We call this the Huey Voice User Interface (UI). The system as illustrated in Figure 41 will expect the programmer to use a pre-defined language to request the application perform an action. The complexity level of this language will be much less than conversational speech between two people, but also slightly less complicated than the strict syntax required by all contemporary programming languages.

**Pseudo-code Description of High-Level Huey Shell Algorithm in Huey Server**
*Key entities are in [brackets]*

**Huey Server**:
- Loads [rules] into memory from rules file (rules.tsv)
- Each [rule] maps an [action] for an [engine] to execute per [input] type
- Launches **HueyShell**, which runs until terminated by user/admin command or fatal error

**HueyShell**:
- Loads [handlers] in priority order
- Starts main [event loop]
- **If** there is something in the [input stack], pop it and process as [user][input] as in **Else** here
- **Else** on receiving a line of [user] text [input]:
  - Iterates over list of [handlers] in priority order:
    - **If a high-priority** [handler] (e.g. help, wake, VNS) [handler] attempts to process [user] [input]
      - -> see details below
    - **Else If the [peek][handler]** sees a [compound request]:
      - [peek] splits the [input] and pushes the individual [request] items onto the [input stack]
    - **Else If a regular [skill][handler]** can handle this [input]:
      - [handler] process [input], performs [action], returns [result] to client for [user]
    - **Else the [catch-all] [handler]** attempts to process the [user] [input]:
      - Currently just reports an error, tells [client] that [input] could not be processed

**Figure 42.** High-Level Huey Shell Algorithm in Huey Server

### *5.14.2. Huey Command Line Interface*

The **Huey Command Line Interface** (CLI) is a set of programs that accepts user input (from text or speech-to-text) and sends it to the **Huey Server** for processing.

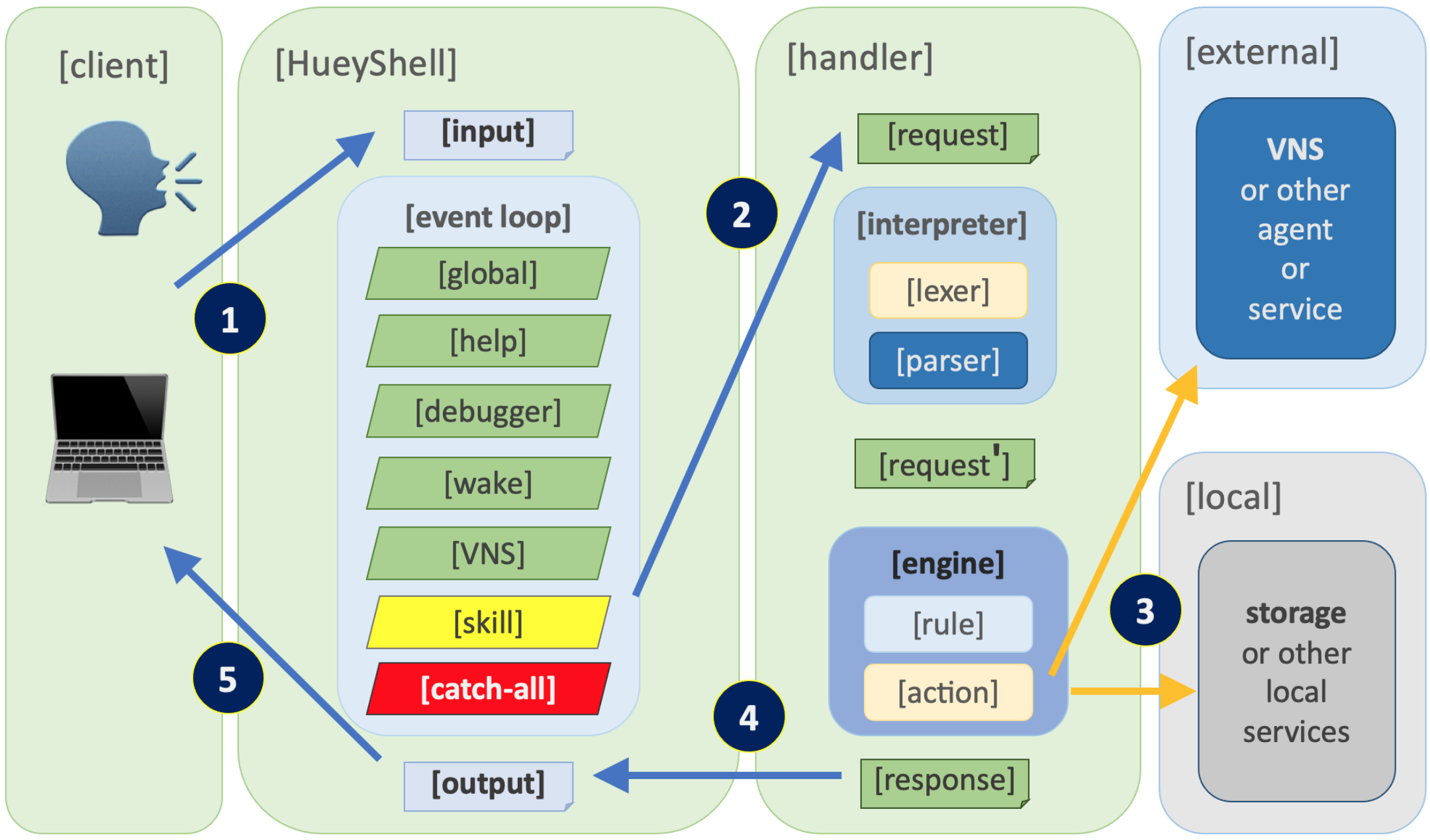

**Figure 43.** Huey Client / Server Interaction

Depending on the nature of the input, the server may carry out an action, such as adding an item to a user's list or sending a request to another server.

After processing, the server sends back a response to the CLI for display to the user. There are settings for the CLI to adjust how verbose the dialog is, so if a request succeeded normally the user may just see the CLI prompt displayed, waiting for the next input.

A pseudo-code description of this process is shown in Figure 42.

The **high-priority handlers** recognize commands related to these inputs in the order listed:

- Global requests [Login, Quit etc.]
- Help requests [Help handler]
- Debug commands [Debugger handler]
- Wake commands [Wake handler]
- VNS switching requests [VNS handler]

These handlers are given high priority so they may intercept special commands before potentially letting a `[skill][handler]` try to process the request.

This way, if the user is attempting to run a `[meta command]` (e.g. **:h** for help or **:f** to see the current frame) the system will "see" that immediately, then quickly provide a response and/or do an action. The request is marked as handled, so no other lower priority handlers will see it.

If a `[skill][handler]` can handle this [input], it continues as follows:

- [handler] loads [interpreter] and orchestrates [request] processing with [engine]
- [handler] populates [request] object using [key words]
- [handler] gives [request] to [interpreter]
- [interpreter] loads [lexer] and [parser] to process [request]
- [interpreter] extracts [key words] from [input] based on parsing & interpretation

- [handler] starts an [engine] designed to implement one or more [action] for this specific [skill]
- [handler] passes [request] to [engine]
- [engine] attempts to find an [action] [rule] that corresponds to [request]
- if [rule] found, [engine] executes [action] and captures output
- [engine] sends [response] back to [handler]
- [handler] reports [result] to caller, typically the Huey Shell [HueyShell]
- [HueyShell] communicates with user regarding response and potential follow-up

See Figure 43 which shows the interactions between client and servers via handlers.

The [VNS handler] will attempt to communicate with other servers and agents as described in the next section.

## 5.15. *Contributions from Huey Architecture*

In this section we have described the Huey language, the Jak language, and the architecture of Huey systems. Although the reader will recognize common software design patterns, this architecture provides the foundation to build systems that are greater than the sum of its parts.

Huey is the framework for encapsulating the objects, processes, and connections to turn textual requests into system actions or messages to be forwarded to other agents or systems. Features can be assembled with this framework by writing new grammars to interpret user's requests or even other system outputs from remote agents. Once a grammar is written, software tools can generate parser and lexer code, allowing the developer to focus on language elements and functionality rather than byte-wrangling. These parsers and lexers are the main components of interpreters, digesting sentences or phrases into more compact representations. A handler orchestrates the interpreter and asks an engine to take the digested request and carry out actions on data, send a message, or simply provide a response to the requesting user.

As the Huey system is developed further, Jak programs can be used to aid in these software development tasks. For developers this will provide a level of abstraction above custom-built solutions created by working directly in Java or Python.

These vendor-neutral design and runtime components are stepping stones to creating intelligent agents that participate on the emerging Open Voice Network. Organizations and individuals who wish to create smarter computers can do so without making major investments in learning one or more proprietary commercial systems.

Applications built with Huey will be deployable to nearly any computing environment, ranging from nanocomputers to cloud environments. There are no inherent dependencies on any vendor's compute, storage, networking, or security services or hardware.

The Open Voice Network will be available for developers to use the existing public internet infrastructure that allows computers and users in nearly all countries to communicate. Huey applications will be first-class citizens in this ecosystem while not depending on any particular operating system or vendor-specific development tools.

## 6. Policy Considerations

Table 9 lays out example solutions to topics that might be encountered in a regulatory environment when implementing the conversational commerce standard. Some of those listed may require government oversight to implement, and the majority have software solutions that could be implemented by device manufacturers and others.

| Regulatory Constraints | |
|---|---|
| **Neutral Navigation** | A user's service could not be discriminated against based on the user, the content being delivered, or the owner of the content. |
| **PII Removed** | A voice can tell a lot about a user, safe-guarding a user's PII could allow for more trust from users. |
| **Privacy** | Companies could consider Privacy by Design [46] where privacy protection is built into products, services, and organizations from inception. |
| **Voice "Incognito" Mode** | Users could use an incognito browsing mode to limit tracking by organizations and further protect their privacy. |
| **Data Transferability** | Data could be device agnostic, meaning data could have the ability to travel from one device to another, be accessed, and used. |
| **Voice Synthesization** | A device could alter the way its voice sounds resulting in the potential need for user permission settings to either accept or decline this behavior. |
| **Command Standardization** | To avoid confusion when switching between devices, a standard list of commands could come pre-installed on every device. |
| **Psychiatric Diagnosis** | A person's mental state, deduced by voice, could inform excellent user interactions. Corresponding privacy concerns could be considered. |
| **Learned Models** | If a machine learning model is built with the intention of sharing, what company's cloud could the model be trained, executed and stored on? |
| **Awareness** | If devices are constantly aware of what is going on around them, to what extent could they interfere if they sense wrongdoing? |
| **Session Transfer Neutrality** | A user's session with a device could be able to be transferred to another regardless of device of origin, user, and content being engaged. |
| **Consumer Protection** | Current consumer protections for online commerce and physical retail could be extended to conversational commerce to prevent fraud and more. |
| **Ad Tracking and Cookies** | A device could store cookies similarly to a computer browser, restrictions should be specified as to whether passive listening can be used to determine consumer behavior. |
| **Biasless** | A device could hold no prejudices against a user. |

Table 9.: A summary of potential regulatory concerns that could arise with the proposed conversational commerce standard.

Incorporating the aforementioned regulatory constraints from the standard's inception could significantly speed the process of creation and adoption. Having discussed the possible constraints enforced by a potential regulatory body, we will now broaden our vision and consider a potential architecture for conversational commerce devices as presented in the grocery shopping use case.

## 7. Conclusion

This paper began outlining a simple use case that illustrates possible benefits of an open voice standard. In writing it, we attempted to initiate a discussion about how to make a conversational commerce space, and increase the adoption of related technologies. By no means do the points covered in this paper represent a finite list, rather, hopefully they signify to all of us the beginning of a dialogue that will vastly improve conversational commerce.

It is worth reiterating that by proposing a standard, this paper's motivation is not to derail current, large device manufacturers progress, initiated over a decade ago by Apple with the introduction of Siri. Rather, the paper aims to help find an optimal competitive ground enabling an ecosystem with as many players as possible including third-party companies and large device manufacturers. Through implementation of the standard, there would be increased adoption of devices and thus additional monetary opportunities for both device manufacturers and third-party companies. Cisco is a great example of this potential shift as it capitalized on the open model of the Internet brought on by TCP/IP.

Before our vision of a standard is fully realized, perhaps the most important pending issue to resolve is how it will be improved going forward and why would others use it. We believe there are multiple possible scenarios that may address successfully these two questions. The following is a list that illustrates what may need to happen for one of them to succeed:

- **New Standards Organization:** A new organization, like the Open Voice Network, which started as a result of research in our lab, may grow an architecture that becomes part of Linux and as such is integrated in all devices becoming the de facto standard.
- **FAANGS:** The big players may decide it's time to be compatible so that Siri, Alexa, OK Goole and other talk to each other and accept other friends to their club.
- **Health Care:** Health care provides may have a great interest in this space since voice has the potential to transform healthcare in many ways as we have shown in our research.
- **Regulation:** We feel regulators need to step in and make sure voice becomes a source for public good. The retail and technology sectors are not the only ones that would benefit. Before health care providers step in, very simple regulation could help create large datasets for longitudinal voice-based ai models to address dementia, airborne pandemics and many more.
- **Open Source Community:** We feel the technology is there for a WWW-style standard that takes over the world.

## Appendices

ANTLR version 4 grammar files are presented in this appendix, along with their BNF equivalents. To learn more about ANTLR (pronounced *antler*), please visit

*https://www.antlr.org*

ANTLR allows the programmer to define parser grammar files and lexer grammar files. All the files presented here are parser grammar files, except for CommonLexer.

## Appendix A. ANTLR Grammars

### *Minimum VNS ANTLR Grammars*

We provide multiple grammar hierarchies as described earlier in this paper.

In this appendix we present RootVNS, the minimal set of grammar statements to support assistants in catching a wake phrase and switching between assistants or remote services. This is the top of the VNS grammar hierarchy, and via the `import` statement we include Switch, CommonParser, and CommonLexer.

Some grammar files also include the `options` and `@header` directives, which are used by the ANTLR code generator to integrate generated code with hand-written code.

All the ANTLR grammar files have atoms listed in lowercase, to reduce the amount of duplicated text to maintain. A pre-processor in Java compares inputs as lower case when employing the grammars.

Following the instructions in the ANTLR documentation, you may generate Java lexer and parser programs starting with one of the hierarchies of grammar files (starting with RootVNS, RootShop, or RootExpense). Be sure to name each grammar file with a ".g4" extension to match the ANTLR v4 convention.

### *Dynamic Assistant Name Support*

The VNS ecosystem supports an arbitrary number of assistants which may be located anywhere and have any name, as long as they are registered in the name service. For these reasons we do not hard-code all these assistant types and names in our source code or the grammar files.

Instead, in the Switch grammar `assistant` statement, we include the `item` element, in addition to commonly known assistants such as Alexa, Siri, et al. The `item` element in turn is defined in CommonParser as a phrase consisting of between one and six words.

At run time, an application communicating with assistants on the VNS may receive a user request (spoken, via chat, etc.) that mentions an assistant. If the name is not matched to the short list hard-coded in the grammar, the application will perform a lookup using a VNS query. As with the existing DNS system, different hosts may cache VNS info locally to speed query resolution. Also, there may be other non-authoritative name servers that the host can query if there is not a local cache hit. Finally, if those name servers do not provide the addressing info, the query will be propagated to an authoritative name server.

```
grammar RootVNS;

// Top-level ANTLR4 parser grammar for catching wake phrase
// and switching between assistants
//
// Revised 4/4/2021
//
// MIT License
//
// Directives for ANTLR generators to use when creating Java code
// Must stay at top of file
options {
    superClass = HueyParser;
}

@header {
import huey.lang.common.HueyParser;
}

// Grammar Precedence
// Statements defined in the first grammar file processed
// (e.g. typical this file) take precedence
// over statements imported from other files.

// TODO modify this import list for each application
// Import on one line to avoid compiler errors
// Lexers go to the end, and CommonParser should be right before that
import Switch, CommonParser, CommonLexer;


// Include all top level command inputs here
// TODO modify this top-level definition to include the statement types
// defined in your grammar list above
input   :   meta
        |   meta stmt
        |   meta_other stmt
        |   meta_other
        |   stmt
        ;

stmt    :   tell_assistant
        |   tell_assistant_compound
        |   stmt_login
        |   stmt_import
        ;
```

```
grammar Switch;

// ANTLR4 grammar for catching wake phrase
// and switching between assistants
//
// Revised 9/27/2020
//
// MIT License
//
//

tell_assistant_compound    : tell_assistant and tell_assistant
                           ;

tell_assistant     : meta ignore* connect assistant
                   | connect assistant service_action?
                   | tell_assistant_at_time service_action
                   ;

tell_assistant_at_time     : meta ignore* connect assistant day_relative
                                    time_of_day? (for_of time_str)?
                           ;

service_action     : to? action item for_of service music?
                   | to? action det? room thing ignore?
                   ;

meta_other   : ignore* connect det? assistant for_of det? skill_remote ignore?
             | ignore* wake? connect? det? assistant 'assistant'? robot?
                       (for_of|to)? det? skill_remote ignore?
             | ignore* wake? connect? assistant for_of? det? skill_remote ignore?
             | ignore* wake? connect? assistant ignore?
             ;

stmt_login   : login username ;


// Atom literals should typically be all lower case, as we match that way
// in the Java code

// Note we cannot use keyword "import" as a lexer or parser element in ANTLR
// as it's a keyword
stmt_import : imp item ('as' item)?
            | imp item 'from' item ('as' item)?
            | 'from' item imp item ('as' item)?
            ;

meta     : attn? wake ignore?
         ;
```

```
attn    : 'hey' | 'ok'
        | 'hello' | 'hi'
        ;

wake    : 'huey' | 'sigma' | 'sigmoid' | 'sigmund'
        | 'alexander' | 'alex'
        ;


connect : 'ask' | 'tell'
        | ('connect' to?)
        | switch_to
        | ('go' to?)
        ;


switch_to : 'switch' ('back')? to?
          ;

// NOTE:  expression [item] can match more than one word -- see CommonParser grammar
assistant : google
          | 'amazon'? 'alexa'
          | 'apple'? 'siri' | 'cortana'
          | wake | item
          ;

robot   : 'robot' | 'bot' | 'android' | item
        ;

google  : 'google' 'assistant'?
        | 'google' 'home'?
        ;

service : 'spotify' | 'pandora'
        | 'amazon'
        | 'apple' 'itunes'?
        | 'youtube' | 'yt'
        | 'tidal'
        ;

music   : 'music' | 'songs' | 'playlist'
        ;

skill_remote : 'forecast' | 'weather' | 'weather' 'forecast'
             | 'calendar' | 'appointments'
             | 'schedule' | 'alarms'
             | (('daily' | 'flash' | 'today' | 'todays' )? 'briefing')
             | ('sing' | 'singing') (item)?
             ;
```

```
parser grammar CommonParser;

// Parser grammar imported by Root grammar and shared across use cases
//
// Revised 9/21/2020
//
// MIT License

item : ID ID ID ID ID ID
     | ID ID ID ID ID
     | ID ID ID ID
     | ID ID ID
     | ID ID
     | ID
     ;

num  : NUMBER ;

email_addr : 'email' | 'email address'
           ;

template   : 'template' ;

verb     : add | del | sel
         ;

login    : ('login' | 'log in' | 'sign in') to?
         ;

username : ID ;

imp      : 'import' ;

add      : 'add'
         | 'append'
         ;

merge    : 'merge' | 'combine' | 'join'
         ;

// del and purge must not overlap
del      : 'delete' | 'remove'
         ;

purge    : 'purge' | 'clear everything' | 'clear all' | 'clear'
         ;

sort     : 'sort' | 'arrange' | 'reorder'
         ;
```

```
sel      : 'select'
         | 'highlight'
         ;

qty      : NUMBER
         ;

// Avoid using [item] as a catch-all here, and instead use [ID], because of
// interpreting [item] elsewhere, e.g. Java handler or interpreter code

unit     : 'pound' | 'pounds' | 'ounce' | 'ounces' | 'oz' | 'quarts'
         | 'gallon' | 'gallons'
         | 'dozen' | 'few' | 'lot' | 'some' | 'group' | 'groups'
         | 'pint' | 'pints' | 'half-pint' | 'half pint' | 'half-pints' | 'half pints'
         | 'container' | 'containers' | 'basket' | 'baskets' | 'bushel' | 'bushels'
         | 'sack' | 'sacks' | 'bag' | 'bags' | 'box' | 'boxes' | 'tray' | 'trays'
         | ID
         ;

create   : 'create'
         | 'make'
         ;

share    : 'share'
         | 'send'
         ;

save     : 'save'
         ;

print    : 'print'
         | 'display'
         | 'show'
         ;

bye      : 'bye'
         | 'goodbye'
         | 'good bye'
         | 'exit'
         | 'close'
         ;


det      : 'a' | 'an'
         | 'this'
         | 'the'
         | 'my'
         | 'our'
         | 'new'
         ;
```

```
and      : 'and' | 'plus'
         ;

for_of   : 'for' | 'of' | 'at' | 'by'
         ;

from     : 'from' | 'with' | 'using' ;

to       : 'to' | 'to my' | 'through' | 'thru' | '-' ;

to_be    : 'to' | 'to be' | 'to become' | 'set' | 'set to'
         | 'is' | 'was'
         ;

all      : 'all' | 'everything'
         ;

ignore   : 'please'
         | 'can you' | 'would you' | 'could you'
         | 'i' | 'want' | 'need'
         | 'now' | 'um' | 'uh' | 'ah'
         | COMMA | COLON | DOT | BANG | SEMI
         ;

place    : 'home' | 'work' | 'office'
         ;

room     : 'kitchen' | 'front door'
         | 'living room' | 'den'
         ;

action   : 'play' | 'search' | 'turn on' | 'turn off'
         ;

thing    : 'breakfast' | 'dinner' | 'tip' | 'tips'
         | 'light' | 'lights'
         | 'fan' | 'air conditioner'
         | 'lunch' | 'ride' | item
         ;

// Date expressions
date_exp : month '/' day '/' year
          | month_name (day | day_name) year
          | month_name (day | day_name)
          ;

month     : MONTH_NUM ;

day       : DAY_NUM ;
```

```
year      : YEAR_NUM ;

day_name : 'first' | 'second' | 'third' | 'fourth' | 'fifth' | 'sixth' | 'seventh'
         | 'eighth' | 'ninth' | 'ten' | 'tenth' | 'eleven' | 'eleventh'
         | 'twelve' | 'twelfth' | 'thirteen' | 'thirteenth'
         ;

month_name : 'january' | 'february' | 'march' | 'april' | 'may' | 'june' | 'july'
           | 'august' | 'september' | 'october' | 'november' | 'december' ;

day_relative : 'today' | 'tomorrow' | 'yesterday'
             | 'next week' | 'last week'
             ;

time_of_day : 'morning' | 'afternoon' | 'evening' | 'night'
            ;

time_str    : time_num WS? ('am' | 'AM' | 'pm' | 'PM')
            ;

time_num : NUMBER
         | NUMBER COLON NUMBER
         ;

num_word : 'one' | 'two' | 'three' | 'four' | 'five' | 'six' | 'seven' | 'eight'
         | 'nine' | 'ten'| 'twenty' | 'fifty' | 'hundred' | 'thousand' | 'million'
         | 'billion' | 'trillion' | 'zillion' | 'quadrillion'
         ;

currency : 'dollar' | 'dollars' | 'buck' | 'bucks'
         ;

coin_phrase : and? num_word? num_word? coin
            ;

coin        : 'cent' | 'cents'
            ;
```

```
lexer grammar CommonLexer;

// Common lexer grammar
// Revised 9/24/2020
//
// MIT License
//
// Borrowed from ANTLR 4:  ANTLRv4Lexer.g4
// -----------
// Punctuation
//
// Character sequences used as separators, delimters, operators, etc
//
DOT   : '.' ;
SEMI  : ';' ;
COLON : ':' ;
COMMA : ',' ;
BANG  : '!' ;

// Lexer rule, not a fragment
NUMBER    : DIGIT_NO_ZERO | DIGIT
          | FLOAT
          | LEAD_ZERO_TWO_DIGITS | NO_LEAD_ZERO_TWO_DIGITS
          | DIGIT_NO_ZERO DIGIT*
          ;

// Use num or Number in parser rules, not "INT"

INT2 : DIGIT DIGIT;

INT4 : DIGIT DIGIT DIGIT DIGIT;

// Borrowed from ANTLR 4 example Java.g4

FLOAT     : ('0'..'9')+ '.' ('0'..'9')* ;

MONTH_NUM : DIGIT_NO_ZERO | ('0')('1'..'9') | ('1')('0'..'2') ;

DAY_NUM   : DIGIT_NO_ZERO | ('0')('1'..'9') | ('1'..'2')('0'..'9') | '30' | '31' ;

YEAR_NUM  : (NINETEEN | TWENTY)? (DIGIT DIGIT) ;

fragment
DIGIT     : '0'..'9' ;

fragment
DIGIT_NO_ZERO : '1'..'9' ;

fragment
LEAD_ZERO_TWO_DIGITS    : ('0')('1'..'9') ;
```

```
fragment
NO_LEAD_ZERO_TWO_DIGITS : ('1'..'9')('0'..'9') ;

fragment
ZERO_ONE     : [0-1] ;

fragment
ZERO_ONE_TWO : [0-2] ;

fragment
ONE_TWO      : [1-2] ;

fragment
ZERO         : '0' ;

fragment
NINETEEN     : '19' ;

fragment
TWENTY       : '20' ;

// Borrowed from ANTLR 4 example Java.g4
LETTER       : [a-zA-Z] ;

ID           : LETTER (LETTER | DIGIT)*  ;

// Borrowed from ANTLR 4 example FuzzyJava.g4

// match anything between /* and */
COMMENT      :   '/*' .*? '*/'    -> channel(HIDDEN)
             ;

WS           :   [ \r\t\u000C\n]+ -> channel(HIDDEN)
             ;

LINE_COMMENT : '//' ~[\r\n]* '\r'? '\n' -> channel(HIDDEN)
             ;
```

## Appendix B. ANTLR Shopping Grammars

For shopping use cases, such as those described in this paper, start with the RootShop grammar instead of RootVNS. RootShop is the top of the shopping hierarchy. Add to this grammar ShoppingList as well as the shared CommonParser and CommonLexer grammars described earlier.

```
grammar RootShop;

// Top-level ANTLR4 parser grammar for shopping commands
//
// Revised 3/21/2021
//
// MIT License
//
// Directives for ANTLR generators to use when creating Java code
// Must stay at top of file
options {
    superClass = HueyParser;
}

@header {
import huey.lang.common.HueyParser;
}

// Grammar Precedence
// Statements defined in the first grammar file processed
// (e.g. typical this file) take precedence
// over statements imported from other files.

// TODO modify this import list for each application
// Import on one line to avoid compiler errors
// Lexers go to the end, and CommonParser should be right before that
import Switch, ShoppingList, CommonParser, CommonLexer;


// Include all top level command inputs here
// TODO modify this top-level definition to include the statement types
// defined in your grammar list above
input   :   meta | meta_shop
        |   meta stmt
        |   meta_other stmt
        |   meta_other
        |   stmt
        ;

stmt    :   stmt_shop_top
        |   tell_assistant
        |   tell_assistant_compound
        |   stmt_login
        |   stmt_import
        ;
```

```
parser grammar ShoppingList;

// Huey Shopping list grammar
// Conversational commerce example
//
// CRUD operations for a shopping list in memory
//
// Revised 9/21/2020
//
// MIT License
//
// NOTES:
// 1. All atoms should be all lower case, so they match statements in Java code
// 2. If grammar is changed, regenerate Java code with ANTLR
// 3. Rebuild app and test
//
// Meta command
// e.g. "create new shopping list" or "create shopping list"
// NOTE:  for "create", must say what they want to create
meta_shop : create det? (store | place)? shoppingList
          | share  det? shoppingList?
          | save   det? shoppingList?
          | print  det? (store | place)? shoppingList?
          | bye    det? shoppingList?
          | connect
          ;

stmt_shop_top : stmt_shop and stmt_shop
              | stmt_shop
              ;

stmt_shop : ignore? add_item | merge_lists | del_item | send_list
          | purge_list | sort_list
          ;

// should NOT match sheet/expense command:  add an expense for lodging

add_item : add det? to_shop_list? add_qty_item more_items?
         | add_qty_item more_items? to_shop_list?
         | add qty? bundled_item to_shop_list?
         | add qty? item to_shop_list?
         | add store? item to_shop_list?
         | ignore* to? buy det? bundled_item to_shop_list?
         ;

//     ignore* buy ignore? det? ignore? qty? ignore? bundled_item to_shop_list?

buy           : 'buy' | 'purchase' | 'get'
              ;
```

```
to_shop_list : to? det? (store | place)? shoppingList?
             ;

// Just an "item" or item with units, optional add additional items
bundled_item : det? unit? for_of? det? item more_items?
             ;

add_qty_item : ignore? add ignore? qty? ignore? det? ignore? item
             ;

more_items   : (and add? qty? det? item)*
             ;

merge_lists  : ignore* merge det? (store | place) shoppingList from
                  det? (store | place) shoppingList ignore*
             ;

// Cannot use del here, because delete and purge must not overlap
purge_list   : purge all? from? det? (store | place)? shoppingList?
             ;

sort_list : sort det? (store | place)? shoppingList?
          ;

del_item  : ignore* del det? item from? det? (store | place)? shoppingList?
          | ignore* del bundled_item from? det? (store | place)? shoppingList?
          ;

send_list : ignore* share det? place shoppingList to det? place email_addr
          ;

phrase    : (ID WS?)+
          ;

store     : 'whole foods' | 'lidl' | 'target' | 'star market'
          ;

// First word before 'list' is required
shoppingList : ('shopping' | 'grocery' | 'supermarket')? 'list'
             ;
```

## Appendix C. ANTLR Expense Sheet Grammars

For expense worksheet use cases, start with the RootExpense grammar instead of RootVNS. Add to this grammar Sheet and ExpenseSheet, as well as the shared CommonParser and CommonLexer grammars described earlier.

```
grammar RootExpense;

// Grammar that includes sheet and expense but
// not other use cases (such as shopping).
//
// Revised 3/21/2021
//
// MIT License
//
//
// Directives for ANTLR generators to use when creating Java code
// Must stay at top of file
options {
    superClass = HueyParser;
}

@header {
import huey.lang.common.HueyParser;
}

// Import on one line to avoid compiler errors
// Lexers go to the end
import Switch, Sheet, ExpenseSheet, CommonParser, CommonLexer;


// Include all top level command inputs here
input   :   meta | meta_sheet
        |   stmt_sheet_top
        |   meta stmt
        |   meta_other stmt
        |   meta_other
        |   stmt
        ;

stmt    :   meta ignore* meta_sheet from det? topic_expense? topic_expense? template
        |   tell_assistant
        |   tell_assistant_compound
        |   stmt_login
        |   stmt_import
        ;
```

```
parser grammar Sheet;

// Sheet or spreadsheet grammar
//
// Revised 11/22/2020
//
// MIT License

// Meta command
// e.g. "create new spreadsheet" or "create sheet"
// NOTE:  for "create", must say what they want to create
meta_sheet : create det? sheet | create det? det? sheet
           | share  det? sheet?
           | save   det? sheet?
           | print  det? sheet?
           | bye    det? sheet?
           ;

stmt_sheet_top : stmt_sheet
               | stmt_expense
               | meta ignore* meta_sheet from det? topic_expense?
                      topic_expense? template
               ;

sheet        : 'spreadsheet' | 'spread sheet' | 'sheet' ;

stmt_sheet   : add_row | add_col | set_cell ;

add_row      : 'add row' cell_phrase+ ;

add_col      : ('add column') col_name ;

cell_phrase : col_name? cell_value ;

col_name     : item ;

cell_value   : ID
             | num
             | date_exp
             ;

set_cell     : ignore? 'set'? det? col_entity to_be? cell_value
             ;

col_entity   : ID ;
```

```
parser grammar ExpenseSheet;

// Sheet or spreadsheet expense report grammar
//
// Revised 9/21/2020
//
// MIT License


stmt_expense : ignore* add det? topic_expense for_of category_expense ignore*
             | to_be det? field_expense to_be category_expense and det?
                    field_expense to cell_value
             | det? field_expense for_of? det? topic_expense? to_be? num_word*
                    currency? coin_phrase? ignore*
             ;

category_expense : 'airfare' | 'lodging' | 'hotel' | 'meals' | 'other'
                 | 'miscellaneous' | ('ground' 'transportation'?)
                 ;

topic_expense    : 'travel' | ('travel')? 'expense' | 'expenses'
                 ;

field_expense    : 'description'
                 | 'date'
                 | 'amount'
                 ;
```

## Appendix D. BNF Grammars

This appendix contains BNF grammars that mirror those presented in the ANTLR appendix. These were converted from ANTLR g4 files with a custom utility and then hand-edited to create valid BNF.

As explained in the ANTLR appendix, you may start with the RootVNS, RootShop, or RootExpense grammar at the top level. Then add in the remaining grammar definitions, similar to the `import` statement seen in the ANTLR grammar files.

```
/* RootVNS */

/* Top-level grammar for catching wake phrase */
/* and switching between assistants */

/* Revised 11/28/2020 */
/* MIT License */

<input> ::= <meta>
  | <meta> <stmt>
  | <meta_other> <stmt>
  | <meta_other>
  | <stmt>

<stmt> ::= <tell_assistant>
  | <tell_assistant_compound>
  | <stmt_login>
  | <stmt_import>
```

```
/* Switch */
/* Grammar for catching wake phrase */
/* and switching between assistants */

/* Revised 11/28/2020 */
/* MIT License */

/* Atom literals should typically be all lower case, as we match that way */

/* Note we can't use keyword "import" as an element in ANTLR as it's a keyword */
/* NOTE:  [item] can match more than one word -- see CommonParser grammar */

<tell_assistant_compound> ::= <tell_assistant> <and> <tell_assistant>

<tell_assistant> ::= <meta> <ignore>* <connect> <assistant>
  | <connect> <assistant> <service_action>?
  | <tell_assistant_at_time> <service_action>

<tell_assistant_at_time> ::= <meta> <ignore>* <connect> <assistant>
    <day_relative> <time_of_day>? (<for_of> <time_str>)?

<service_action> ::= <to>? <action> <item> <for_of> <service> <music>?
  | <to>? <action> <det>? <room> <thing> <ignore>?

<meta_other> ::= <ignore>* <connect> <det>? <assistant> <for_of> <det>?
      <skill_remote> <ignore>?
  | <ignore>* <wake>? <connect>? <det>? <assistant> "assistant"? <robot>?
     (<for_of|to>)? <det>? <skill_remote> <ignore>?
  | <ignore>* <wake>? <connect>? <assistant> <for_of>? <det>?
     <skill_remote> <ignore>?
  | <ignore>* <wake>? <connect>? <assistant> <ignore>?

<stmt_login> ::= <login> <username>

<stmt_import> ::= <imp> <item> ("as" <item>)?
  | <imp> <item> "from" <item> ("as" <item>)?
  | "from" <item> <imp> <item> ("as" <item>)?

<meta> ::= <attn>? <wake> <ignore>?

<attn> ::= "hey" | "ok"
  | "hello" | "hi"

<wake> ::= "huey" | "sigma" | "sigmoid" | "sigmund"
  | "alexander" | "alex"

<connect> ::= "ask" | "tell"
  | ("connect" <to>?)
  | <switch_to>
  | ("go" <to>?)
```

```
<switch_to> ::= "switch" ("back")? <to>?

<assistant> ::= <google>
  | "amazon"? "alexa"
  | "apple"? "siri" | "cortana"
  | <wake> | <item>

<robot> ::= "robot" | "bot" | "android" | <item>

<google> ::= "google" "assistant"?
  | "google" "home"?

<service> ::= "spotify" | "pandora"
  | "amazon"
  | "apple" "itunes"?
  | "youtube" | "yt"
  | "tidal"

<music> ::= "music" | "songs" | "playlist"

<skill_remote> ::= "forecast" | "weather" | "weather" "forecast"
  | "calendar" | "appointments"
  | "schedule" | "alarms"
  | (("daily" | "flash" | "today" | "todays")? "briefing")
  | ("sing" | "singing") (<item>)?
```

```
/* grammar RootShop; */
/* Grammar for shopping commands */
/* Revised 11/28/2020 */
/* MIT License */

<input> ::= <meta> | <meta_shop>
  | <meta> <stmt>
  | <meta_other> <stmt>
  | <meta_other>
  | <stmt>

<stmt> ::= <stmt_shop_top>
  | <tell_assistant>
  | <tell_assistant_compound>
  | <stmt_login>
  | <stmt_import>
```

```
/* ShoppingList */
/* Huey Shopping list grammar */
/* Conversational commerce example */
/* CRUD operations for a shopping list in memory */

/* Revised 11/28/2020 */
/* MIT License */

/* NOTES: */
/* All atoms should be all lower case, for easier parsing */

/* Meta command */
/* e.g. "create new shopping list" or "create shopping list" */

/* NOTE:  for "create", must say what they want to create */

/* Just an "item" or item with units, optional add additional items */
/* Cannot use del here, because delete and purge must not overlap */
/* First word before 'list' is required */

<meta_shop> ::= <create> <det>? (<store> | <place>)? <shoppingList>
  | <share> <det>? <shoppingList>?
  | <save> <det>? <shoppingList>?
  | <print> <det>? (<store> | <place>)? <shoppingList>?
  | <bye> <det>? <shoppingList>?
  | <connect>

<stmt_shop_top> ::= <stmt_shop> <and> <stmt_shop>
  | <stmt_shop>

<stmt_shop> ::= <ignore>? <add_item> | <merge_lists> | <del_item> | <send_list>
  | <purge_list> | <sort_list>

<add_item> ::= <add> <det>? <to_shop_list>? <add_qty_item> <more_items>?
  | <add_qty_item> <more_items>? <to_shop_list>?
  | <add> <qty>? <bundled_item> <to_shop_list>?
  | <add> <qty>? <item> <to_shop_list>?
  | <add> <store>? <item> <to_shop_list>?
  | <ignore>* <to>? <buy> <det>? <bundled_item> <to_shop_list>?

<buy> ::= "buy" | "purchase" | "get"

<to_shop_list> ::= <to>? <det>? (<store> | <place>)? <shoppingList>?

<bundled_item> ::= <det>? <unit>? <for_of>? <det>? <item> <more_items>?

<add_qty_item> ::= <ignore>? <add> <ignore>? <qty>? <ignore>? <det>?
  <ignore>? <item>

<more_items> ::= (<and> <add>? <qty>? <det>? <item>)*
```

```
<merge_lists> ::= <ignore>* <merge> <det>? (<store> | <place>)
  <shoppingList> <from> <det>? (<store> | <place>) <shoppingList> <ignore>*

<purge_list> ::= <purge> <all>? <from>? <det>? (<store> | <place>)?
  <shoppingList>?

<sort_list> ::= <sort> <det>? (<store> | <place>)? <shoppingList>?

<del_item> ::= <ignore>* <del> <det>? <item> <from>? <det>?
    (<store> | <place>)? <shoppingList>?
  | <ignore>* <del> <bundled_item> <from>? <det>? (<store> | <place>)?
    <shoppingList>?

<send_list> ::= <ignore>* <share> <det>? <place> <shoppingList> <to>
  <det>? <place> <email_addr>

<phrase> ::= (<ID> <WS>?)+

<store> ::= "whole foods" | "lidl" | "target" | "star market"

<shoppingList> ::= ("shopping" | "grocery" | "supermarket")? "list"
```

```
/* grammar RootExpense; */
/* Grammar that includes sheet and expense but  */
/* not other use cases (such as shopping). */
/* Revised 11/28/2020 */
/* MIT License */

<input> ::= <meta> | <meta_sheet>
  | <stmt_sheet_top>
  | <meta> <stmt>
  | <meta_other> <stmt>
  | <meta_other>
  | <stmt>

<stmt> ::= <meta> <ignore>* <meta_sheet> <from> <det>? <topic_expense>?
    <topic_expense>? <template>
  | <tell_assistant>
  | <tell_assistant_compound>
  | <stmt_login>
  | <stmt_import>
```

```
/* Sheet or spreadsheet grammar */

/* Revised 11/28/2020 */
/* MIT License */

/* Meta command */
/* e.g. "create new spreadsheet" or "create sheet" */

/* NOTE:  for "create", must say what they want to create */

<meta_sheet> ::= <create> <det>? <sheet> | <create> <det>? <det>? <sheet>
  | <share> <det>? <sheet>?
  | <save> <det>? <sheet>?
  | <print> <det>? <sheet>?
  | <bye> <det>? <sheet>?

<stmt_sheet_top> ::= <stmt_sheet>
  | <stmt_expense>
  | <meta> <ignore>* <meta_sheet> <from> <det>? <topic_expense>?
    <topic_expense>? <template>

<sheet> ::= "spreadsheet" | "spread sheet" | "sheet"
<stmt_sheet> ::= <add_row> | <add_col> | <set_cell>
<add_row> ::= "add row" <cell_phrase>+
<add_col> ::= ("add column") <col_name>
<cell_phrase> ::= <col_name>? <cell_value>
<col_name> ::= <item>
<cell_value> ::= <ID>
  | <num>
  | <date_exp>

<set_cell> ::= <ignore>? "set"? <det>? <col_entity> <to_be>? <cell_value>

<col_entity> ::= <ID>
```

```
/* ExpenseSheet */
/* Sheet or spreadsheet expense report grammar */

/* Revised 11/28/2020 */
/* MIT License */

<stmt_expense> ::= <ignore>* <add> <det>? <topic_expense> <for_of>
    <category_expense> <ignore>*
  | <to_be> <det>? <field_expense> <to_be> <category_expense> <and>
    <det>? <field_expense> <to> <cell_value>
  | <det>? <field_expense> <for_of>? <det>? <topic_expense>? <to_be>?
    <num_word>* <currency>? <coin_phrase>? <ignore>*

<category_expense> ::= "airfare" | "lodging" | "hotel" | "meals" | "other"
  | "miscellaneous" | ("ground" "transportation"?)

<topic_expense> ::= "travel"
  | ("travel")? "expense" | "expenses"

<field_expense> ::= "description"
  | "date"
  | "amount"
```

```
/* CommonParser */
/* Parser grammar imported by Root grammar and shared across use cases */

/* Revised 11/28/2020 */
/* MIT License */

/* del and purge must not overlap */

/* Avoid using [item] as a catch-all here, and instead use [ID], because of */
/* interpreting [item] elsewhere, e.g. in handler or interpreter code */

/* Date expressions */
<item> ::= <ID> <ID> <ID> <ID> <ID> <ID>
  | <ID> <ID> <ID> <ID> <ID>
  | <ID> <ID> <ID> <ID>
  | <ID> <ID> <ID>
  | <ID> <ID>
  | <ID>

<num> ::= <NUMBER>
<email_addr> ::= "email"
  | "email address"

<template> ::= "template"
<verb> ::= <add>
  | <del>
  | <sel>

<login> ::= ("login" | "log in" | "sign in") <to>?

<username> ::= <ID>
<imp> ::= "import"
<add> ::= "add"
  | "append"

<merge> ::= "merge" | "combine" | "join"

<del> ::= "delete" | "remove"

<purge> ::= "purge" | "clear everything" | "clear all" | "clear"

<sort> ::= "sort" | "arrange" | "reorder"

<sel> ::= "select"
  | "highlight"

<qty> ::= <NUMBER>

<unit> ::= "pound" | "pounds" | "ounce" | "ounces" | "oz" | "quarts"
  | "gallon" | "gallons" | "dozen" | "few" | "lot" | "some"
```

```
  | "group" | "groups" | "pint" | "pints" | "half-pint" | "half pint"
  | "half-pints" | "half pints" | "container" | "containers"
  | "basket" | "baskets" | "bushel" | "bushels"
  | "sack" | "sacks" | "bag" | "bags" | "box" | "boxes" | "tray" | "trays"
  | <ID>

<create> ::= "create"
  | "make"

<share> ::= "share"
  | "send"

<save> ::= "save"

<print> ::= "print"
  | "display"
  | "show"

<bye> ::= "bye"
  | "goodbye"
  | "good bye"
  | "exit"
  | "close"

<det> ::= "a" | "an"
  | "this"
  | "the"
  | "my"
  | "our"
  | "new"

<and> ::= "and" | "plus"

<for_of> ::= "for" | "of" | "at" | "by"

<from> ::= "from" | "with" | "using"
<to> ::= "to" | "to my" | "through" | "thru" | "-"
<to_be> ::= "to" | "to be" | "to become" | "set" | "set to"
  | "is" | "was"

<all> ::= "all" | "everything"

<ignore> ::= "please"
  | "can you" | "would you" | "could you"
  | "i" | "want" | "need"
  | "now" | "um" | "uh" | "ah"
  | <COMMA> | <COLON> | <DOT> | <BANG> | <SEMI>

<place> ::= "home" | "work" | "office"
```

```
<room> ::= "kitchen" | "front door"
  | "living room" | "den"

<action> ::= "play" | "search" | "turn on" | "turn off"

<thing> ::= "breakfast" | "dinner" | "tip" | "tips"
  | "light" | "lights"
  | "fan" | "air conditioner"
  | "lunch" | "ride" | <item>

<date_exp> ::= <month> "/" <day> "/" <year>
  | <month_name> (<day> | <day_name>) <year>
  | <month_name> (<day> | <day_name>)

<month> ::= <MONTH_NUM>
<day> ::= <DAY_NUM>
<year> ::= <YEAR_NUM>
<day_name> ::= "first" | "second" | "third" | "fourth" | "fifth"
  | "sixth" | "seventh" | "eighth" | "ninth" | "ten" | "tenth"
  | "eleven" | "eleventh" | "twelve" | "twelfth" | "thirteen" | "thirteenth"
<month_name> ::= "january" | "february" | "march" | "april" | "may" | "june"
  | "july" | "august" | "september" | "october" | "november" | "december"
<day_relative> ::= "today" | "tomorrow" | "yesterday"
  | "next week" | "last week"

<time_of_day> ::= "morning" | "afternoon" | "evening" | "night"

<time_str> ::= <time_num> <WS>? ("am" | "AM" | "pm" | "PM")

<time_num> ::= <NUMBER>
  | <NUMBER> <COLON> <NUMBER>

<num_word> ::= "one" | "two" | "three" | "four" | "five" | "six" | "seven"
  | "eight" | "nine" | "ten" | "twenty" | "fifty" | "hundred" | "thousand"
  | "million" | "billion" | "trillion" | "zillion" | "quadrillion"

<currency> ::= "dollar" | "dollars" | "buck" | "bucks"

<coin_phrase> ::= <and>? <num_word>? <num_word>? <coin>

<coin> ::= "cent" | "cents"
```

```
/* CommonLexer */
/* Common lexer grammar */

/* Revised 11/28/2020 */
/* MIT License */

/* Punctuation */
/* Character sequences used as separators, delimters, operators, etc */

/* Use NUMBER or FLOAT in parser rules, not INT */

<DOT> ::= "."
<SEMI> ::= ";"
<COLON> ::= ":"
<COMMA> ::= ","
<BANG> ::= "!"

<NUMBER> ::= <DIGIT_NO_ZERO> | <DIGIT> | <FLOAT>
  | <LEAD_ZERO_TWO_DIGITS> | <NO_LEAD_ZERO_TWO_DIGITS>
  | <DIGIT_NO_ZERO> <DIGIT>*

<INT2> ::= <DIGIT> <DIGIT>
<INT4> ::= <DIGIT> <DIGIT> <DIGIT> <DIGIT>

<FLOAT>     ::= [0-9] <DOT> [0-9]
<MONTH_NUM> ::= <DIGIT_NO_ZERO> | <LEAD_ZERO_TWO_DIGITS> | "1" <ZERO_ONE_TWO>
<DAY_NUM>   ::= <DIGIT_NO_ZERO> | <LEAD_ZERO_TWO_DIGITS> | <ONE_TWO> <DIGIT>
  | "30" | "31"

<YEAR_NUM>  ::= (<NINETEEN> | <TWENTY>)? (<DIGIT> <DIGIT>)

<DIGIT> ::= [0-9]
<DIGIT_NO_ZERO> ::= [1-9]

<LEAD_ZERO_TWO_DIGITS> ::= "0" [1-9]
<NO_LEAD_ZERO_TWO_DIGITS> ::= [1-9] [0-9]

<ZERO>          ::= "0"
<ZERO_ONE>      ::= [0-1]
<ZERO_ONE_TWO> ::= [0-2]
<ONE_TWO>       ::= [1-2]

<NINETEEN> ::= "19"
<TWENTY>   ::= "20"

<LETTER> ::= [a-z] | [A-Z]

<STRING> ::= (<WS> | <LETTER> | <DIGIT>)+

<ID> ::= <LETTER> (<LETTER> | <DIGIT>)*
```

```
<WS> ::= " "*
<LINE_COMMENT> ::= "//" <STRING>*

<COMMENT> ::= "/*" <STRING>* "*/"
```

## Appendix E. Python Sample Code

This appendix contains sample Python code to demonstrate the feasibility of working with s-expressions and Jak with Huey applications.

The first example `s-to-jak.py` will produce the following output when run with Python 3. Note that for demo purposes we only parse the `search_item` term.

```
Convert s-expr to Jak demo
==========================
Input:
(action_shop (shop_search
(search_item (qty 2) (unit tickets) (when Sunday)
(item red sox game))))

Expected output:
set(search_item, qty=2, unit="tickets", when="Sunday", item="red sox game")

Actual output:
set(search_item, qty=2, unit="tickets", when="Sunday", item="red sox game")

Success! Actual output matches expected output
```

```python
# s-to-jak.py
# Python interpreter that converts a Huey s-expr
# to a Jak statement.
#
# For demonstration purposes -- real interpreter would
# process a more complete set of inputs
#
# Tested with Python 3.8
# Revised 10/18/2020
#

# Test input
s_expr = "(action_shop (shop_search (search_item " + \
    "(qty 2) (unit tickets) (when Sunday) (item red sox game))" + \
    "))"

expect_output = \
    'set(search_item, qty=2, unit="tickets", when="Sunday", item="red sox game")'

# For parsing
match_term = "search_item"
lparen = "("
rparen = ")"
dquote = '"'
found = False
```

```
attr = cleaned = val = ''

# Hardcoded output format for demo purposes
output_prefix = "set(" + match_term
output = output_prefix

# Dictionary to hold terms (attribute/value pairs)
output_terms = {}

# Function definitions

def push_term(k, v):
    """Save a term as an attribute/value pair"""
    output_terms[k] = v

def to_str(v):
    """Output a string version of the passed value"""
    if isinstance(v, str):
        return v
    else:
        return str(v)

def wrap_str(v):
    """Wrap input with quotes if necessary"""
    if v.isnumeric():
        return str(v)
    else:
        return dquote + v.strip() + dquote

def concat(orig, add):
    """Concatenate token to end of existing string"""
    return orig + " " + to_str(add)

# Main program
print("Convert s-expr to Jak demo")
print("==========================")
print("Input:")
print(s_expr)
print("\nExpected output:")
print(expect_output)

# Parse input s-expr, split on spaces
tokens = s_expr.split(' ')

# Loop over tokens starting from left side
for token in tokens:
    if token.startswith(lparen):
        cleaned = token.strip(lparen)
        if cleaned == match_term:
            found = True
```